\PassOptionsToPackage{usenames,dvipsnames}{xcolor}
\documentclass{article} 
\usepackage{iclr2023_conference,times}

\definecolor{mydarkblue}{rgb}{0,0.08,0.45}
\usepackage[colorlinks,citecolor=mydarkblue,urlcolor=mydarkblue,linkcolor=mydarkblue,filecolor=mydarkblue]{hyperref}
\usepackage{url}

   \usepackage[compact]{titlesec}
   \titlespacing{\section}{0pt}{1ex}{0ex}
   \titlespacing{\subsection}{0pt}{1ex}{0ex}
   \titlespacing{\subsubsection}{0pt}{0.5ex}{0ex}
  
\input{preamble/packages}

\usepackage{amsmath,amsfonts,bm}

\def\eqref#1{equation~\ref{#1}}

\def\1{\bm{1}}

\def\rvu{{\mathbf{i}}}

\def\rvk{{\mathbf{k}}}

\def\rvu{{\mathbf{u}}}
\def\rvv{{\mathbf{v}}}

\def\rvx{{\mathbf{x}}}
\def\rvy{{\mathbf{y}}}
\def\rvz{{\mathbf{z}}}

\def\rmV{{\mathbf{V}}}

\DeclareMathAlphabet{\mathsfit}{\encodingdefault}{\sfdefault}{m}{sl}
\SetMathAlphabet{\mathsfit}{bold}{\encodingdefault}{\sfdefault}{bx}{n}

\def\gB{{\mathcal{B}}}

\def\gD{{\mathcal{D}}}
\def\gE{{\mathcal{E}}}
\def\gF{{\mathcal{F}}}
\def\gG{{\mathcal{G}}}

\def\gX{{\mathcal{X}}}
\def\gY{{\mathcal{Y}}}

\def\sR{{\mathbb{R}}}

\newcommand{\E}{\mathbb{E}}

\newcommand{\R}{\mathbb{R}}

\DeclareMathOperator*{\argmax}{arg\,max}
\DeclareMathOperator*{\argmin}{arg\,min}

\newacronym{DNN}{DNN}{deep neural network}

\newacronym{ISP}{\textsc{Isp}}{\textit{Introspective Self-play}}

\newacronym{KKT}{KKT}{Karush–Kuhn–Tucker}

\newacronym{ERM}{ERM}{empirical risk minimization}

\newacronym{JTT}{JTT}{Just Train Twice}

\newacronym{IRM}{IRM}{\textit{Invariant Risk Minimization}}

\newacronym{DRO}{DRO}{Distributionally Robust Optimization}

\newacronym{SSA}{SSA}{Spread Spurious Attribute}

\newacronym{ML}{ML}{machine learning}

\newacronym{MCD}{MCD}{Monte Carlo Dropout}

\newacronym{AL}{AL}{active learning}

\newacronym{DE}{DE}{Deep Ensemble}

\newacronym{GP}{GP}{Gaussian process}

\newacronym{OOD}{OOD}{out-of-domain}

\newacronym{BALD}{BALD}{Bayesian active learning by disagreement}

\newacronym{BADGE}{BADGE}{Batch Active learning by Diverse Gradient Embeddings}

\newacronym{CDF}{CDF}{cumulative distribution function}

\newacronym{GCE}{GCE}{generalized cross entropy}

\newacronym{FL}{FL}{focal loss}

\newtheorem{theorem_app}{Theorem}[section]

\newtheorem{theorem}{Theorem}

\newtheorem{proposition}{Proposition}

\newtheorem{definition}{Definition}

\makeatletter
\newcommand*{\addFileDependency}[1]{
  \typeout{(#1)}
  \@addtofilelist{#1}
  \IfFileExists{#1}{}{\typeout{No file #1.}}
}
\makeatother

\AtAppendix{\counterwithin{proposition}{section}}

\DeclareMathOperator*{\minimize}{minimize}

\algnewcommand{\LeftComment}[1]{\Statex \(\triangleright\) #1}

\algnewcommand{\Inputs}[1]{%
  \State \textbf{Inputs:} #1
}
\algnewcommand{\Output}[1]{%
  \State \textbf{Output:} #1
}
\algnewcommand{\Outputs}[1]{%
  \State \textbf{Outputs:} #1
}
\algnewcommand{\Compute}[1]{%
  \State \textbf{Compute} #1
}
\algnewcommand{\Initialize}[1]{%
  \State \textbf{Initialize:} #1
}
\algnewcommand{\OpEstimate}[1]{%
  \State \textbf{(Optional) Estimate:} #1
}
\algnewcommand{\Estimate}[1]{%
  \State \textbf{Estimate:} #1
}
\algnewcommand{\Infer}[1]{%
  \State \textbf{Infer:} #1
}

\renewcommand{\footnotesize}{\scriptsize} 

\title{
Pushing the
{\color{black}
Accuracy-Group Robustness Frontier}
with Introspective Self-play
}

\author{\textbf{Jeremiah Zhe Liu$^1$, 
Krishnamurthy Dj Dvijotham$^1$, 
Jihyeon Lee$^1$, 
Quan Yuan$^1$, 
Martin Strobel$^{2\dag}$,
} \\  
\textbf{Balaji Lakshminarayanan$^1$\thanks{
Co-sernior authors.
$^\dag$ Work done as a student researcher at Google. MS contributed to the manuscript after its initial ICLR submission.}, 
Deepak Ramachandran$^{1 *}$} \\
$^1$ Google Research $^2$ National University of Singapore
\\
{\scriptsize\texttt{\{jereliu,dvij,jihyeonlee,yquan,
martinstrobel,
balajiln,ramachandrand\}@google.com}}
}

\newcommand{\X}{\mathcal{X}}
\newcommand{\Y}{\mathcal{Y}}
\newcommand{\br}[1]{\left({#1}\right)}

\newcommand{\Rfair}{R_{\text{fair}}}
\newcommand{\Rfairhat}{{\hat{R}_{\text{fair}}}}
\newcommand{\Rfairhatan}{{\hat{R}_{\text{fair}, \alpha, n}}}

\iclrfinalcopy 
\begin{document}









\maketitle
\setcounter{page}{1}

\vspace{-2.5em}
\begin{abstract}
{\color{black}
Standard \gls{ERM} training can produce \gls{DNN} models that are accurate on average but underperform in underrepresented population subgroups, especially when there are imbalanced group distributions in the long-tailed training data. Therefore, approaches that improve the accuracy - group robustness tradeoff frontier of a \gls{DNN} model (i.e. improving worst-group accuracy without sacrificing average accuracy, or vice versa) is of crucial importance.
}
Uncertainty-based \gls{AL} can potentially improve the frontier by preferentially sampling underrepresented subgroups to create a more balanced training dataset.
However, the quality of uncertainty estimates from modern \gls{DNN}s tend to degrade in the presence of spurious correlations and dataset bias,
compromising the effectiveness of \gls{AL} for sampling tail groups.
In this work, we propose \textit{\gls{ISP}}, a simple approach to improve the uncertainty estimation of a deep neural network under dataset bias, by adding an auxiliary \textit{introspection} task requiring a model to predict the bias for each data point in addition to the label.
We show that \gls{ISP} provably improves the \textit{bias-awareness} of the model representation and the resulting uncertainty estimates.
On two real-world tabular and language tasks, \gls{ISP} serves as a simple ``plug-in" for \gls{AL} model training, consistently improving both the tail-group sampling rate and the final accuracy-fairness trade-off frontier of popular \gls{AL} methods.
\end{abstract}
\glsresetall
\vspace{-1.5em}

\section{Introduction}
\label{sec:intro}
\vspace{-0.5em}

\begin{wrapfigure}[12]{R}{0.25\textwidth}
\vspace{-2em}
\centering
\includegraphics[width=0.25\textwidth]{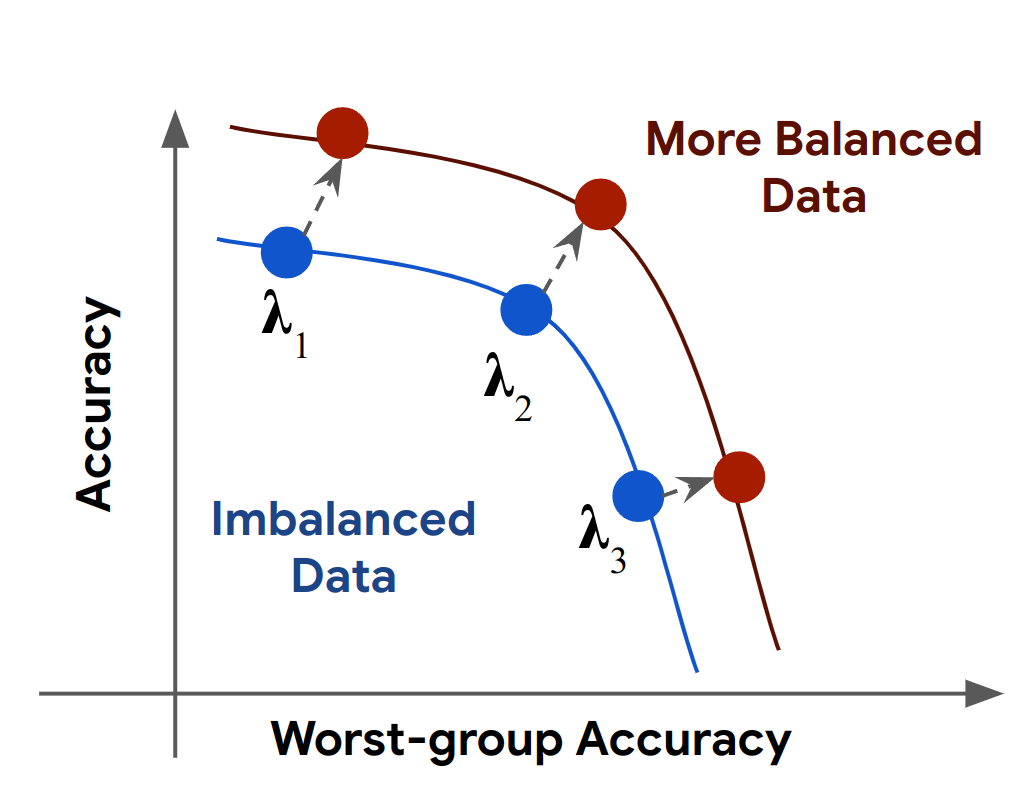}
\vspace{-2em}
\caption{Example of accuracy-fairness frontier. 
Under a more balanced training data distribution, the model can attain a better accuracy-fairness frontier ({\color{red}\textbf{Red}}) when compared to training under an imbalanced distribution ({\color{blue}\textbf{Blue}}) at every tradeoff level $\lambda$ (\Cref{eq:frontier_risk_intro}).}
\label{fig:acc_fair_frontier_intro}
\end{wrapfigure}

Modern \glsfirst{DNN} models are commonly trained on large-scale datasets \citep{deng2009imagenet, raffel2020exploring}.
These datasets often exhibit an imbalanced long-tail distribution with many small population subgroups, reflecting the nature of the physical and social processes generating the data distribution 
\citep{zhu2014capturing, feldman2020neural
}. 
This imbalance in training data distribution, i.e., \textbf{\textit{dataset bias}}, prevents \gls{DNN} models from generalizing equitably to the underrepresented population groups
\citep{hasnain2007disparities
}. 

\textbf{Accuracy-{\color{black} Group Robustness} Frontier:} In response, the existing bias mitigation literature has focused on improving training procedures under a fixed and imbalanced training dataset, striving to balance performance between model accuracy and fairness (e.g., the average-case v.s.\ worst-group performance)
\citep{agarwal2018reductions, martinez2020minimax, martinez2021blind}
. Formally, this goal corresponds to identifying an optimal model $f \in \gF$ that attains the \textit{Pareto efficiency frontier} of the accuracy-{\color{black}group robustness} trade-off (e.g., see \Cref{fig:acc_fair_frontier_intro}), so that under the same training data $D=\{y_i, \rvx_i\}_{i=1}^n$, we cannot find another model $f' \in \gF$ that outperforms $f$ in both accuracy and  {\color{black}worst-group performance}. In the literature, this \textit{accuracy-{\color{black}group robustness} frontier} is often characterized by a trade-off objective \citep{martinez2021blind}:
\begin{align}
    f_\lambda = \argmin_{f\in \gF} F_{\lambda}(f|D); 
    \qquad
    F_{\lambda}(f|D) \coloneqq R_{acc}(f|D) + \lambda R_{\color{black} robust}(f|D),
    \label{eq:frontier_risk_intro}
\end{align}
\vspace{-1.5em}

where $R_{acc}$ and $R_{\color{black} robust}$ are risk functions for a model's accuracy and {\color{black} group robustness} (modeled here-in as worst-group accuracy), and $\lambda > 0$ a trade-off parameter.
Then, $f_\lambda$ cannot be outperformed by any other $f'$ at the same trade-off level $\lambda$.
The entire frontier under a dataset $D$ can then be characterized by finding $f_\lambda$ that minimizes the {\color{black}robustness}-accuracy objective (\ref{eq:frontier_risk_intro}) at every trade-off level $\lambda$, and tracing out its $(R_{acc},  R_{\color{black} robust})$ performances
(\Cref{fig:acc_fair_frontier_intro}).

\textbf{Goal:} However, the limited size of the tail-group examples restricts the \gls{DNN} model's worst-group performance, leading to a compromised accuracy-{\color{black} group robustness} frontier 
\citep{zhao2019inherent, dutta2020there
}, and thus we ask: \textit{Under a fixed learning algorithm, can we meaningfully push the model's accuracy-{\color{black} group robustness} frontier by improving the training data distribution using active learning?} 
That is, denoting by $D_{\alpha, n}=\{(y_i, \rvx_i)\}_{i=1}^n$ a training dataset with $K$ subgroups and the group size distribution $\alpha = [\alpha_1, \dots, \alpha_K]$, we study whether a model's accuracy-{\color{black} group robustness} performance $F_\lambda$ can be improved by rebalancing the group distribution of the training data $D_{\alpha, n}$, i.e., we seek to optimize an outer problem:
\vspace{-1em}
\begin{align}
{\color{Maroon} \minimize_{\alpha \in \Delta^{|\gG|}}}
\Big[ \min_{f \in \gF} F_{\lambda}(f|D_{{\color{Maroon}\alpha},n}) \Big],
\label{eq:al_objective_intro}
\end{align}
where $\Delta^K$ is the simplex of all possible group distributions \citep{rolf2021representation}. Our key observation is that given a sampling model with \textit{well-calibrated} uncertainty (i.e., the model uncertainty is well-correlated with generalization error), \gls{AL} can preferentially acquire tail-group examples from unlabelled data \textit{without needing group annotations}, and add them to the training data to reach a more balanced data distribution \citep{branchaud2021can}. 
{\color{black}\Cref{sec:fairness_connection_app} discusses the connection between group robustness with fairness.}

\textbf{Challenges with DNN Uncertainty under Bias:} However, recent work suggests that a \gls{DNN} model's uncertainty estimate is less trustworthy under spurious correlations and distributional shift, potentially compromising the \gls{AL} performance under dataset bias.
For example, \citet{ovadia2019can} show that a \gls{DNN}'s expected calibration error increases as the testing data distribution deviates from the training data distribution, and \citet{ming2022impact} show that a \gls{DNN}'s ability in detecting out-of-distribution examples is significantly hampered by spurious patterns. Looking deeper, \citet{liu2022simple, van2020uncertainty} suggest that this failure mode in \gls{DNN} uncertainty can be caused by an issue in representation learning known as \textit{feature collapse}, where the \gls{DNN} over-focuses on correlational features that help to distinguish between output classes on the training data, but ignore the non-predictive but semantically meaningful input features that are important for uncertainty quantification (\Cref{fig:2d_example}). In this work, we show that this failure mode can be provably mitigated by a training procedure we term \textit{introspective training} (\Cref{sec:method}). 
Briefly, introspective training adds an auxiliary \textit{introspection} task to model training, asking the model to predict whether an example belongs to an underrepresented group. 
It comes with a guarantee in injecting \textit{bias-awareness} into model representation (\Cref{thm:bias_awareness}), encouraging it to learn diverse hidden features that distinguish the minority-group examples from the majority, even if these features are not correlated with the training labels. 
Hence it can serve as a simple ``plug-in" to the training procedure of any active learning method, leading to improved uncertainty quality for tail groups (\Cref{fig:2d_example}). 

\textbf{Contributions:} In summary, our contributions are:
\vspace{-0.5em}
\begin{itemize}[noitemsep,
topsep=0pt,leftmargin=0.45cm]
\item We introduce \textbf{\gls{ISP}}, a simple training approach to improve a \gls{DNN} model's uncertainty quality for underrepresented groups (\Cref{sec:method}). Using group annotations from the training data, \gls{ISP} conducts \textit{introspective training} to provably improve a \gls{DNN}'s representation and uncertainty quality for the tail groups. When group annotations are not available, \gls{ISP} can be combined with a cross-validation-based \textit{self-play} procedure that uses a noise-bias-variance decomposition of the model's generalization error \citep{domingos2000unified}.
\item \textbf{Theoretical Analysis.} 
We theoretically analyze the optimization problem in  \Cref{eq:al_objective_intro}  under a group-specific learning rate model \citep{rolf2021representation} (\Cref{sec:theory}). Our result elucidates the dependence of the group distribution $\alpha$ in the model’s best-attainable accuracy-{\color{black} group robustness} frontier $F_\lambda$. In particular, it confirms the theoretical necessity of up-sampling the underrepresented groups for obtaining the optimal accuracy-{\color{black} group robustness} frontier, and reveals that \textit{underrepresentation} is in fact caused by an interplay of the subgroup's learning difficulty and its prevalence in the population.
\item \textbf{Empirical Effectiveness.} Under two challenging real-world tasks (census income prediction and toxic comment detection), we empirically validate the effectiveness of \gls{ISP} in improving the performance of AL with a \gls{DNN} model under dataset bias (\Cref{sec:exp}). For both classic and state-of-the-art uncertainty-based \gls{AL} methods, ISP improves tail-group sampling rate, meaningfully pushing the accuracy-{\color{black} group robustness} frontier of the final model.
\end{itemize}

\Cref{sec:related_work} surveys related work.

\textbf{Notation and Problem Setup.} We consider a dataset $D$ where each labeled example $\{\rvx_i, y_i\}$ is associated with a discrete group label $g_i \in \gG=\{1, \dots, |\gG|\}$. We denote $\gD=P(y, \rvx, g)$ the joint distribution of the label, feature and groups, so that $D$ can be understood as a size-$n$ set of i.i.d.\ samples from $\gD$
We denote the prevalence of each group as $\gamma_g=E_{(y, \rvx, g) \sim \gD}(1_{G=g})$ and associate \textit{dataset bias} with the imbalance in group distribution $P(G)=[\gamma_1, \dots, \gamma_{|\gG|}]$ \citep{rolf2021representation}. In the applications we consider, there exists a subset of \textit{underrepresented} groups $\gB \subset \gG$ which are not sufficiently represented in the population distribution $\gD$ so that $\gamma_g \ll \frac{1}{|\gG|}$  for $g \in \gB$ \citep{sagawa2019distributionally, sagawa2020investigation}. We denote $L(y, \hat{y})$ as a  loss function from the Bregman divergence family, and $\gF$ the hypothesis space of predictors $f: \X \mapsto \Y$. We require the model class $\gF$ to be sufficiently expressive so it can model the Bayes-optimal predictor $\tilde{y}(\rvx)=\argmin_y' E_{y \sim P(y|\rvx)}(L(y, y'))$). We also assume $\gF$ has a 
certain degree of smoothness, so that the model $f \in \gF$ cannot arbitrarily overfit to the noisy labels in the training set.\footnote{In the case of over-parameterized models, this usually implies $\gF$ is subject to certain regularization appropriate for the model class (e.g., early stopping for SGD-trained neural networks) \citep{li2020gradient}.} 

\section{Method}
\label{sec:method}
In this section, we introduce \textit{\glsfirst{ISP}}, a simple training approach to improve model quality in representation learning and uncertainty quantification under dataset bias. 
Briefly, \gls{ISP} performs \textit{introspective training} by adding a \textit{underrepresention prediction} head to the model and training it to distinguish whether an example $(y_i, \rvx_i, g_i)$ is from the set of underrepresented groups $\gB$ (\Cref{sec:introspective_training}). When the underrepresentation label $b_i = I(g_i \in \gB)$ is not available, \gls{ISP}  
estimates it based on a cross-validation-based procedure we term \textit{cross-validated self-play} (\Cref{sec:bias_estimation}). 
As we will show, \gls{ISP} carries a guarantee for the model's representation learning and uncertainty estimation quality under dataset bias (\Cref{thm:bias_awareness}).

\subsection{Introspective Training}
\label{sec:introspective_training}

We consider models of the form $p(y|\rvx)=  \sigma\big( f_y(\rvx) \big) = \sigma\big( \beta_y^\top h(\rvx) \big)$, where $h: \gX \rightarrow \sR^D$ is a $D$-dimensional embedding function, $\beta_y \in \sR^D$ the output weights, and $\sigma(\cdot)$ the activation function. 
Given model $f_y=\beta_y^\top h$, \textit{introspective training} adds a bias head $f_b=\beta_b^\top h$ to the model, so it becomes a multi-task architecture $f=(f_y, f_b)$  with shared embedding:
\begin{align}
    p(y|\rvx)=\sigma(f_y(\rvx)), \; p(b|\rvx)=\sigma_{sigmoid}(f_b(\rvx));
    \; \mbox{where} \; 
    (f_y, f_b) = \big( \beta_y^\top h + b_y, \; \beta_b^\top h + b_b \big).
    \label{eq:introspective_predictions}
\end{align}
Given examples $D=\{\rvx_i, y_i, g_i\}_{i=1}^n$, we generate the underrepresentation labels as $b_i = I(g_i \in \gB)$ and train the model with the target and underrepresentation labels $(y_i, b_i)$ by minimizing a standard multi-task learning objective:
\begin{align}
L((y_i, b_i), \rvx_i) = 
L(y_i, f_y(\rvx_i)) + 
L_b(b_i, f_b(\rvx_i)),
\label{eq:introspective_objective}
\end{align}
where $L$ is the standard loss function for the task, and $L_b$ is the cross-entropy loss.
As a result, given training examples $\{\rvx_i\}_{i=1}^n$, \textit{introspective training} not only trains the model to predict the outcome $y_i$, but also instructs it to recognize its potential bias $b_i$ by predicting whether $\rvx_i$ is from an underrepresented group.

\begin{figure}[ht]
\captionsetup[subfigure]{justification=centering}
\centering
\begin{adjustbox}{minipage=\linewidth,scale=0.9}
\begin{subfigure}{.23\textwidth}
  \centering
  \includegraphics[width=\linewidth]{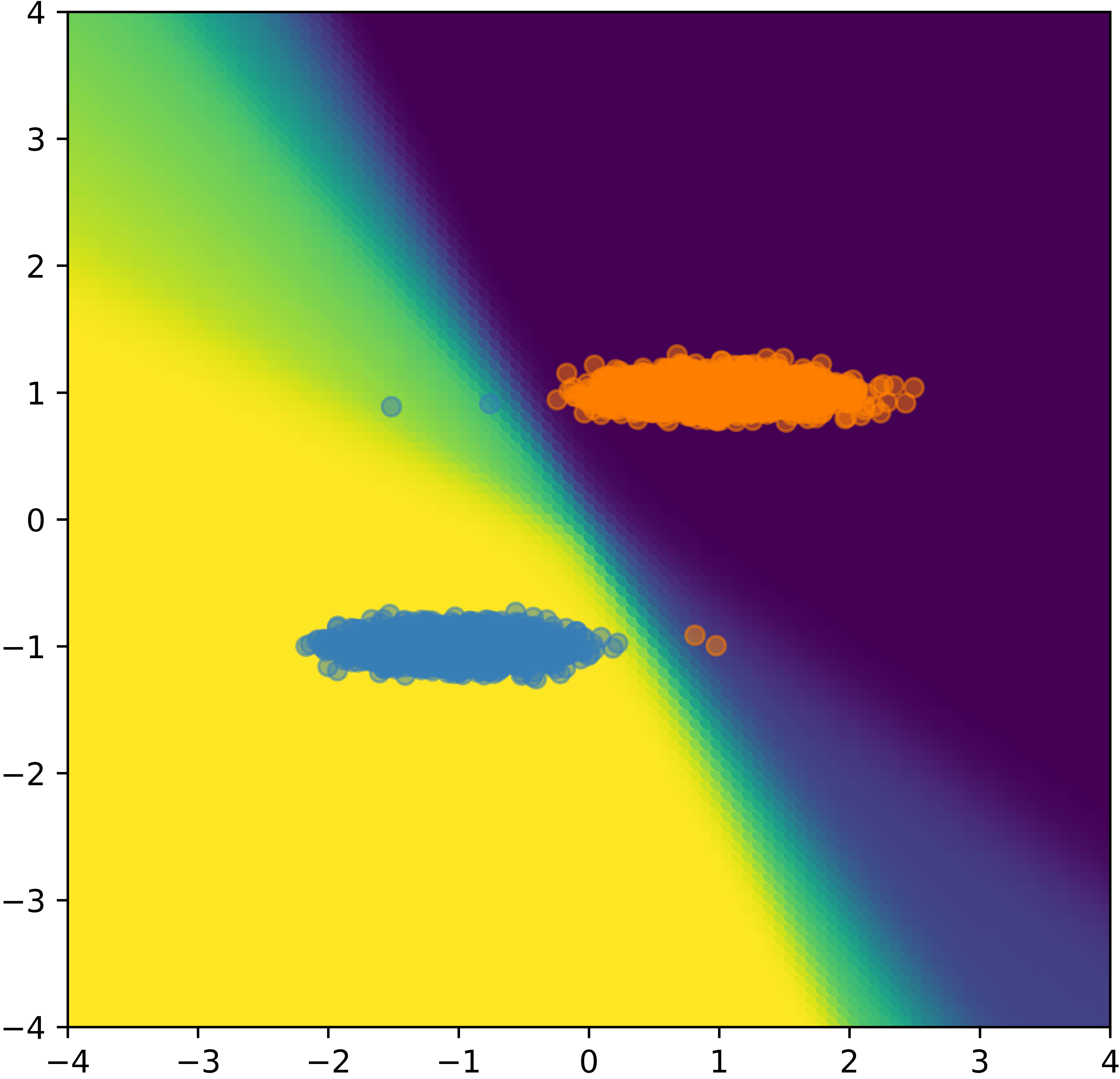}
  \caption{Predicted Probability \\ \quad Introspective Training}
  \label{fig:2d_prediction_isp}
\end{subfigure}
\begin{subfigure}{.23\textwidth}
  \centering
  \includegraphics[width=\linewidth]{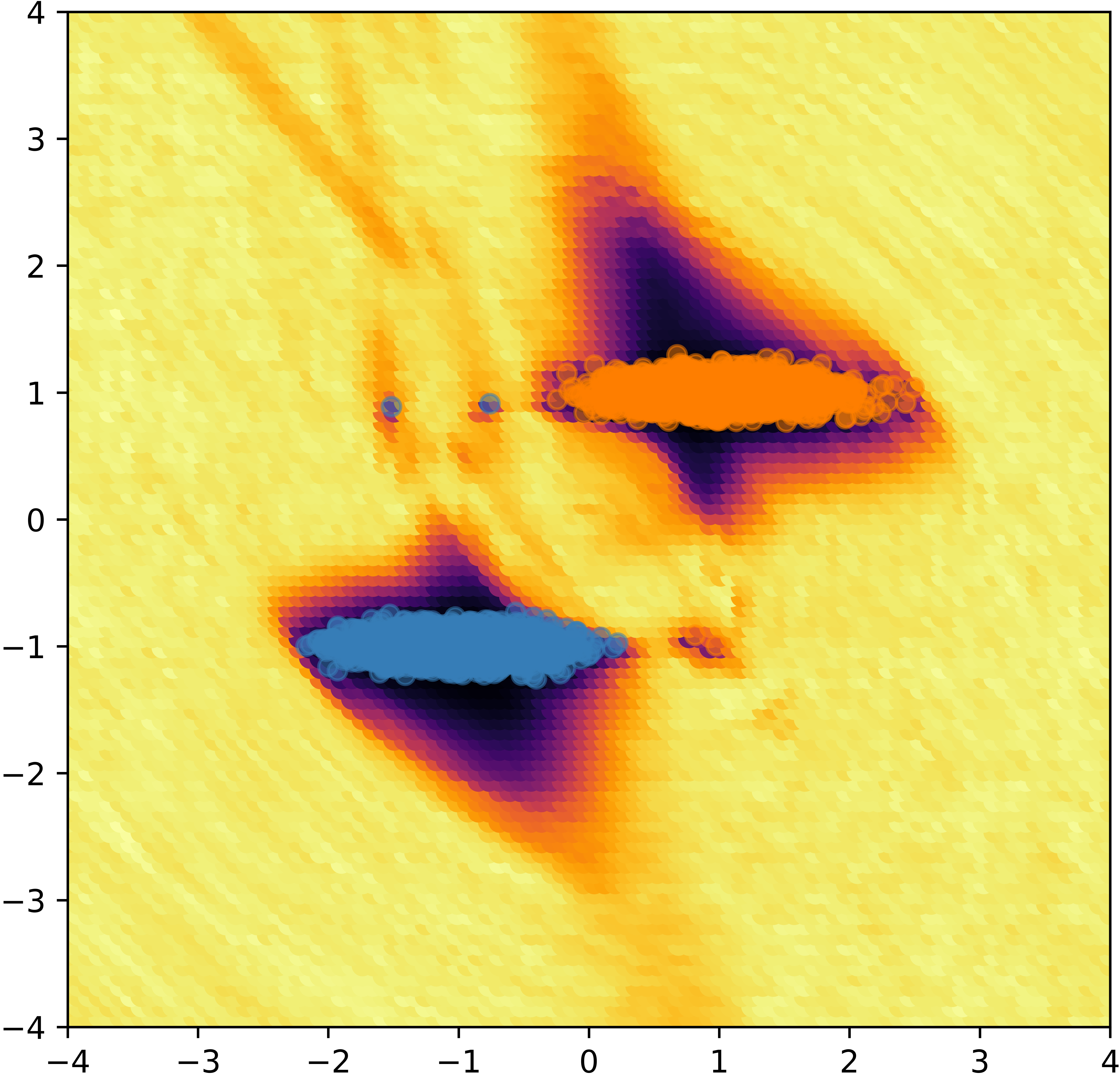}
  \caption{Predictive Variance \\ \quad Introspective Training}
  \label{fig:2d_uncertainty_isp}
\end{subfigure}
\begin{subfigure}{.23\textwidth}
  \centering
  \includegraphics[width=\linewidth]{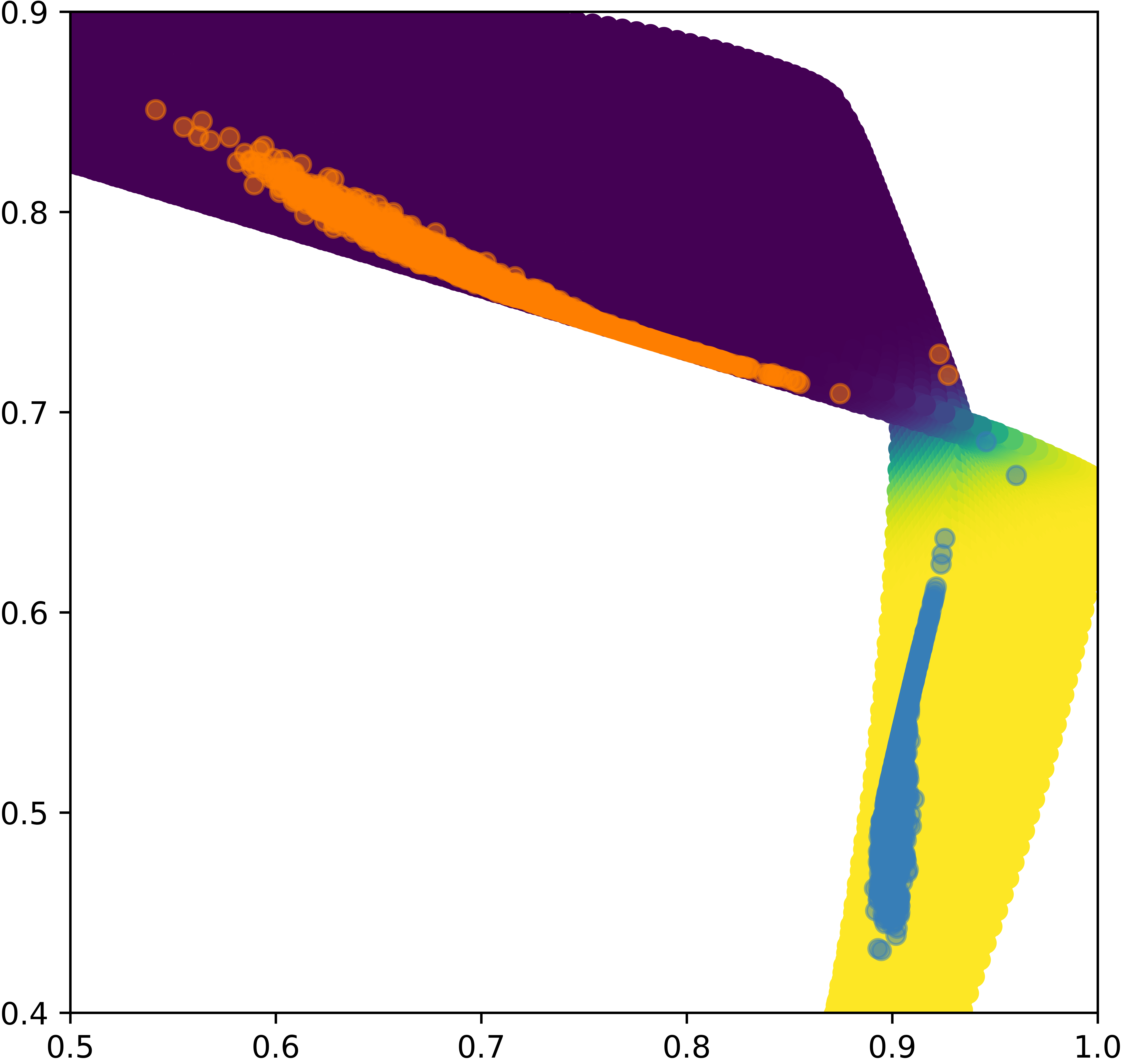}
  \caption{Representation Space \\ \quad Introspective Training \\ {\tiny Colored by Predicted Probability.}}
  \label{fig:2d_representation_isp}
\end{subfigure}
\begin{subfigure}{.23\textwidth}
  \centering
  \includegraphics[width=\linewidth]{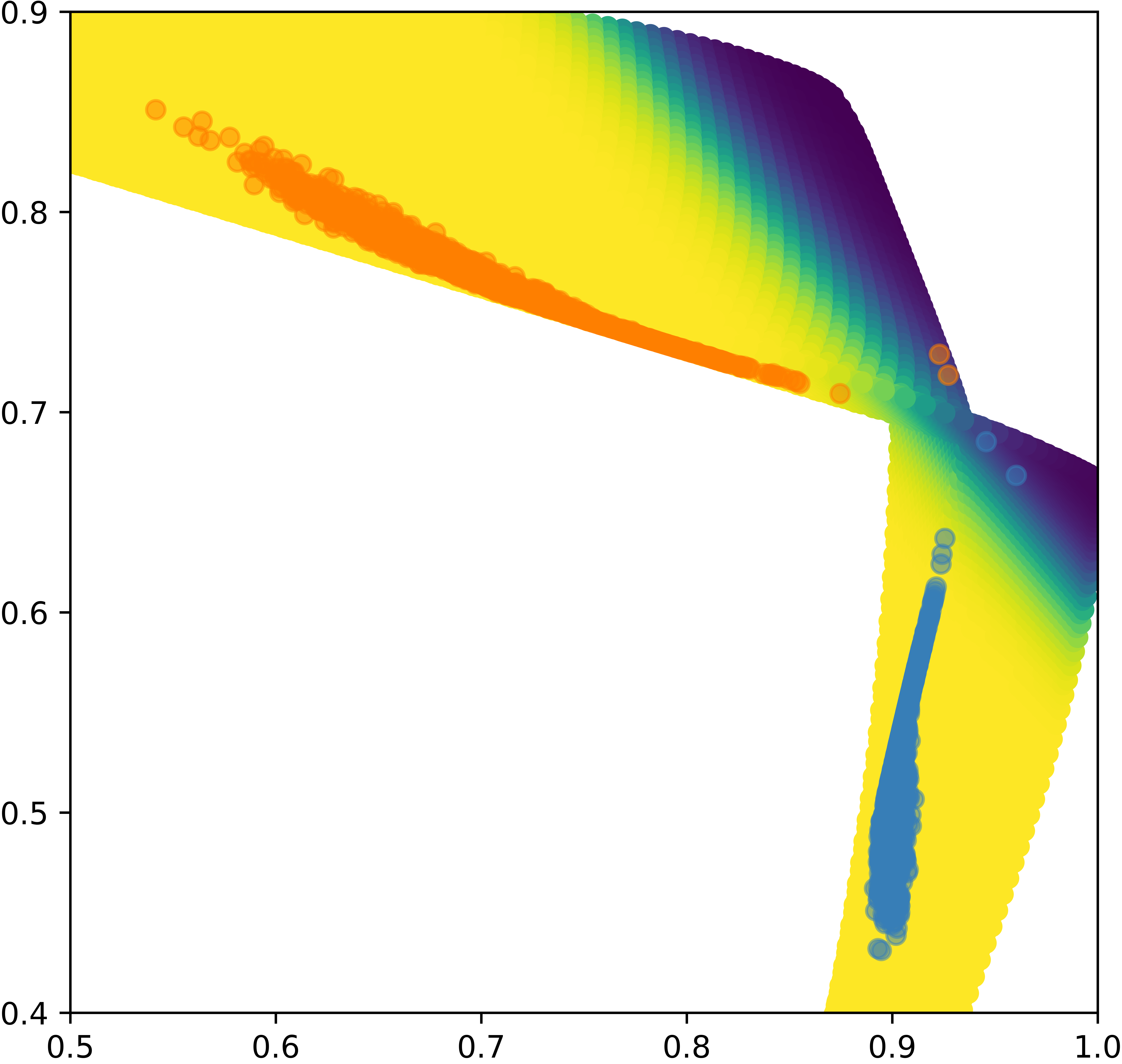}
  \caption{Representation Space \\ \quad Introspective Training \\ {\tiny Colored by Predicted Underrep.}}
  \label{fig:2d_representation_isp_bias}
\end{subfigure}

\begin{subfigure}{.23\textwidth}
  \centering
  \includegraphics[width=\linewidth]{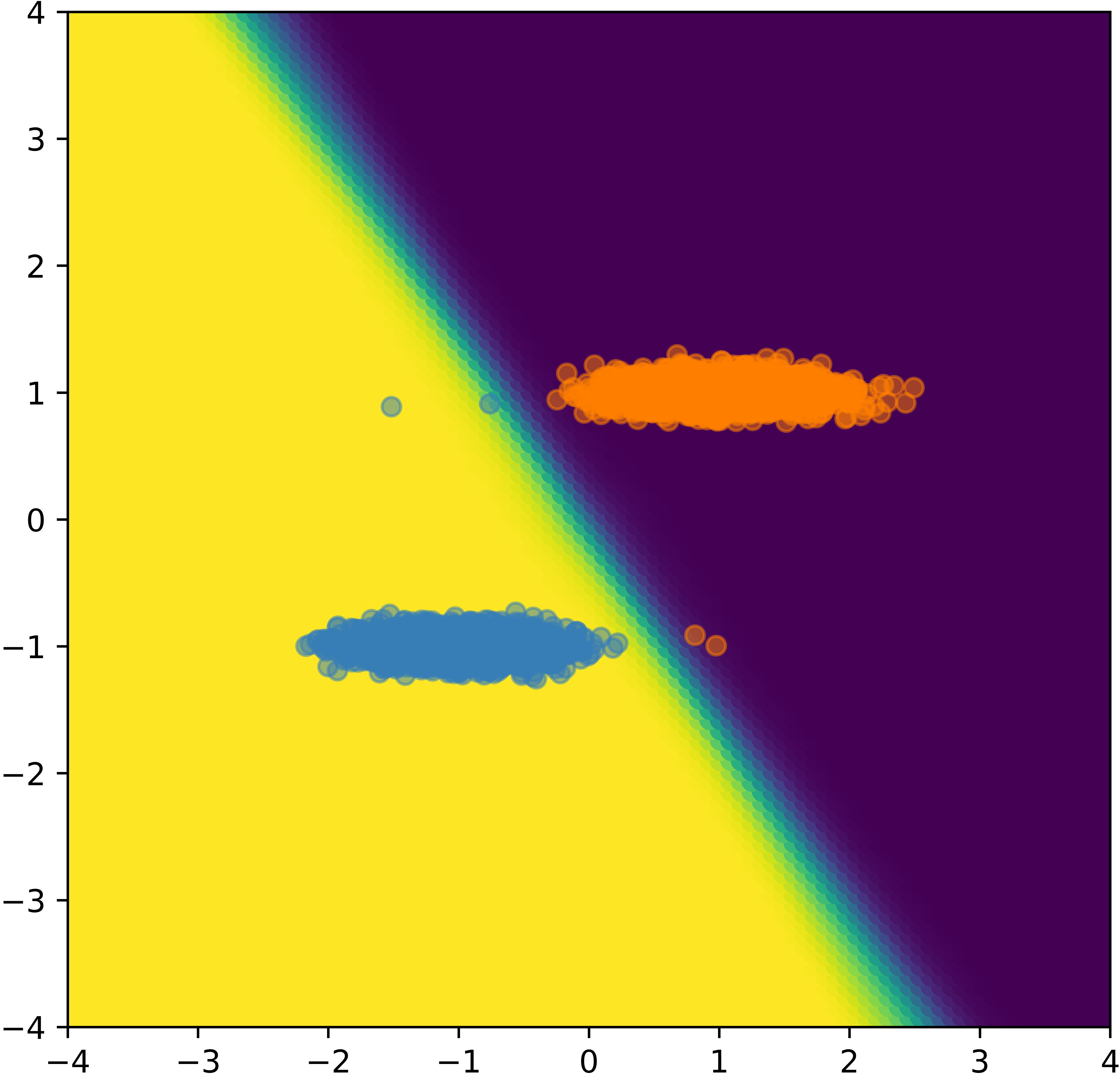}
  \caption{Predicted Probability \\ \quad ERM Training}
  \label{fig:2d_prediction_base}
\end{subfigure}
\begin{subfigure}{.23\textwidth}
  \centering
  \includegraphics[width=\linewidth]{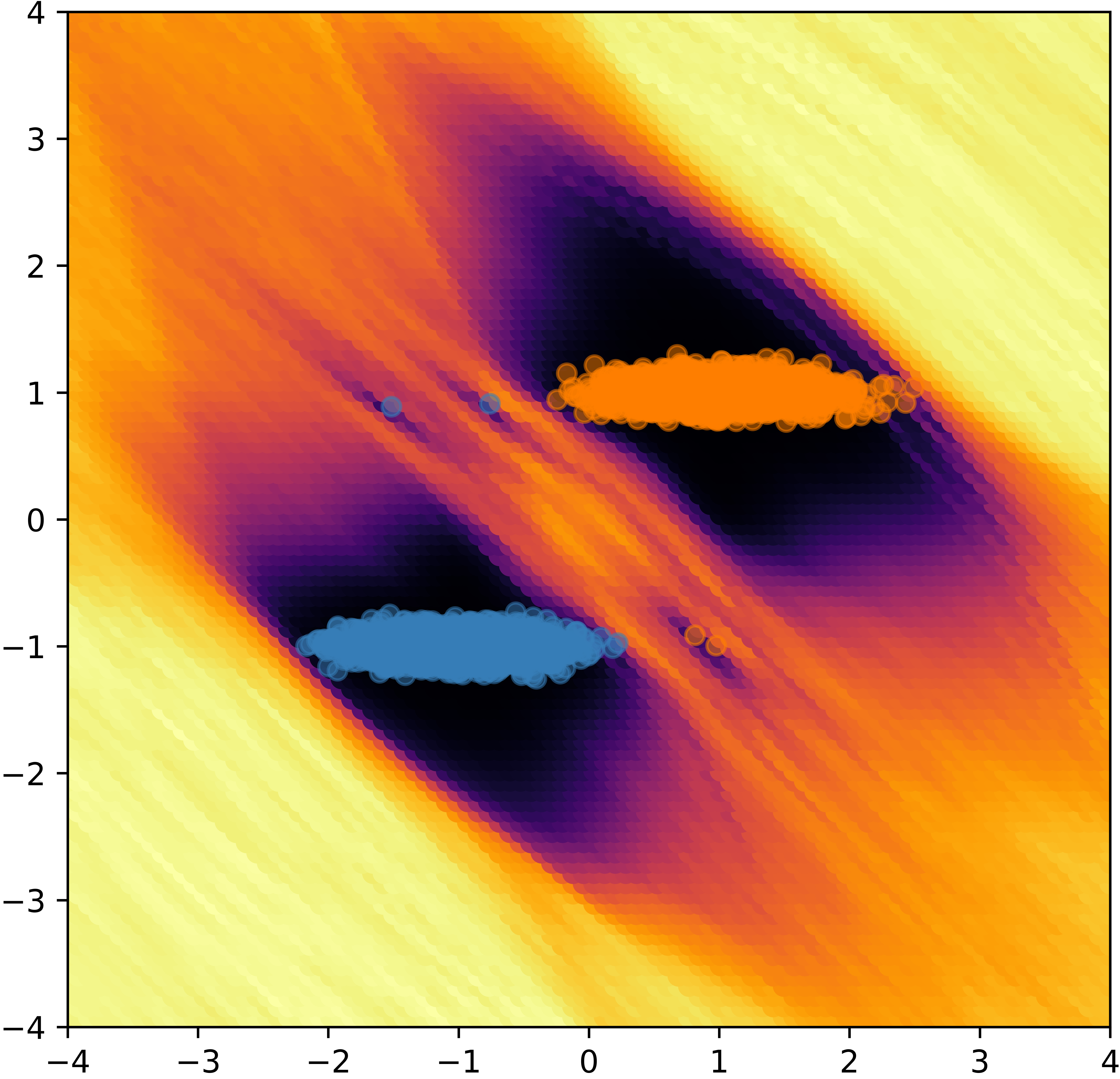}
  \caption{Predictive Variance \\ \quad ERM Training}
  \label{fig:2d_uncertainty_base}
\end{subfigure}
\begin{subfigure}{.23\textwidth}
  \centering
  \includegraphics[width=\linewidth]{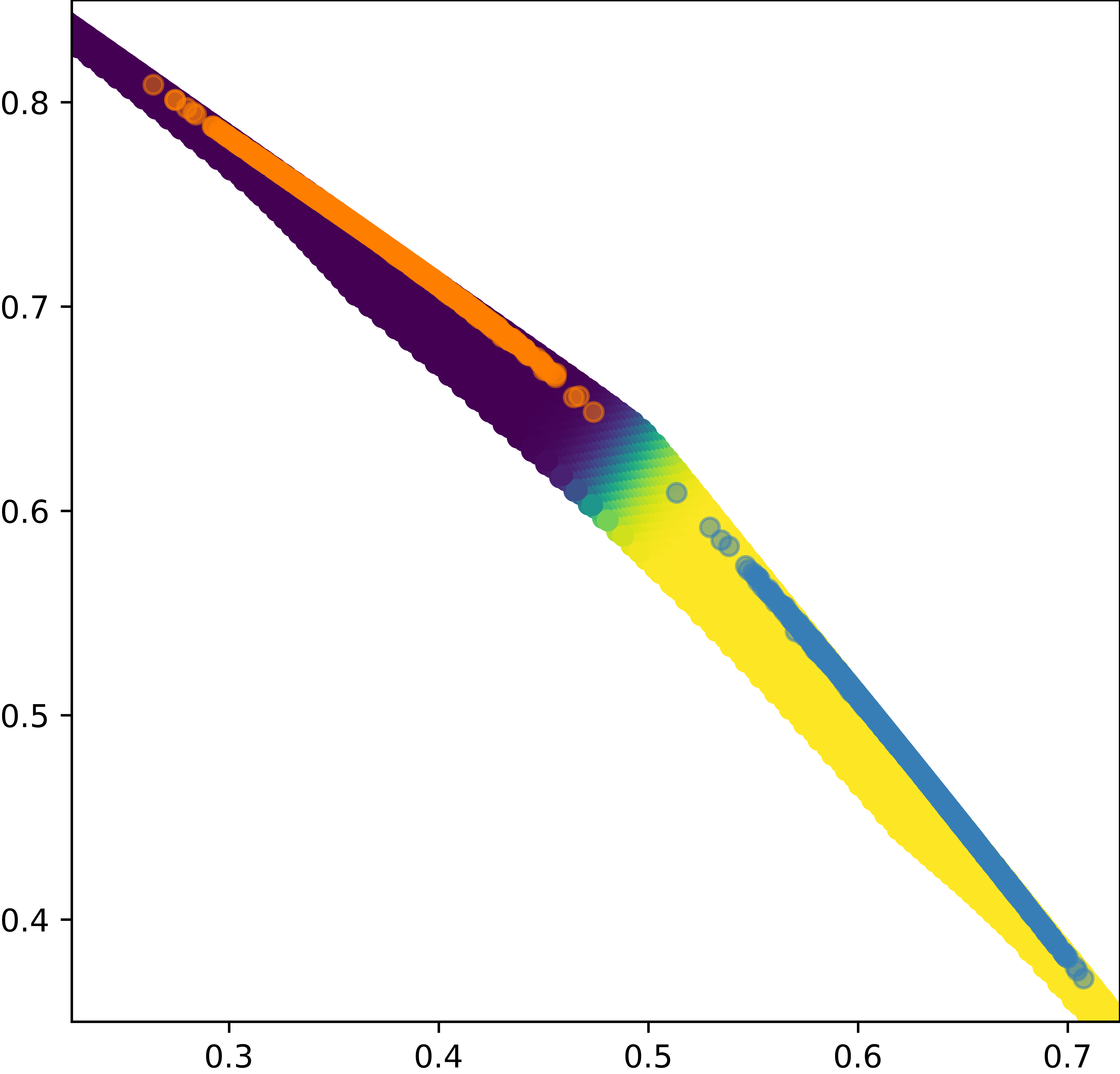}
  \caption{Representation Space \\ \quad ERM Training}
  \label{fig:2d_representation_base}
\end{subfigure}
\begin{subfigure}{.23\textwidth}
  \centering
  \includegraphics[width=\linewidth]{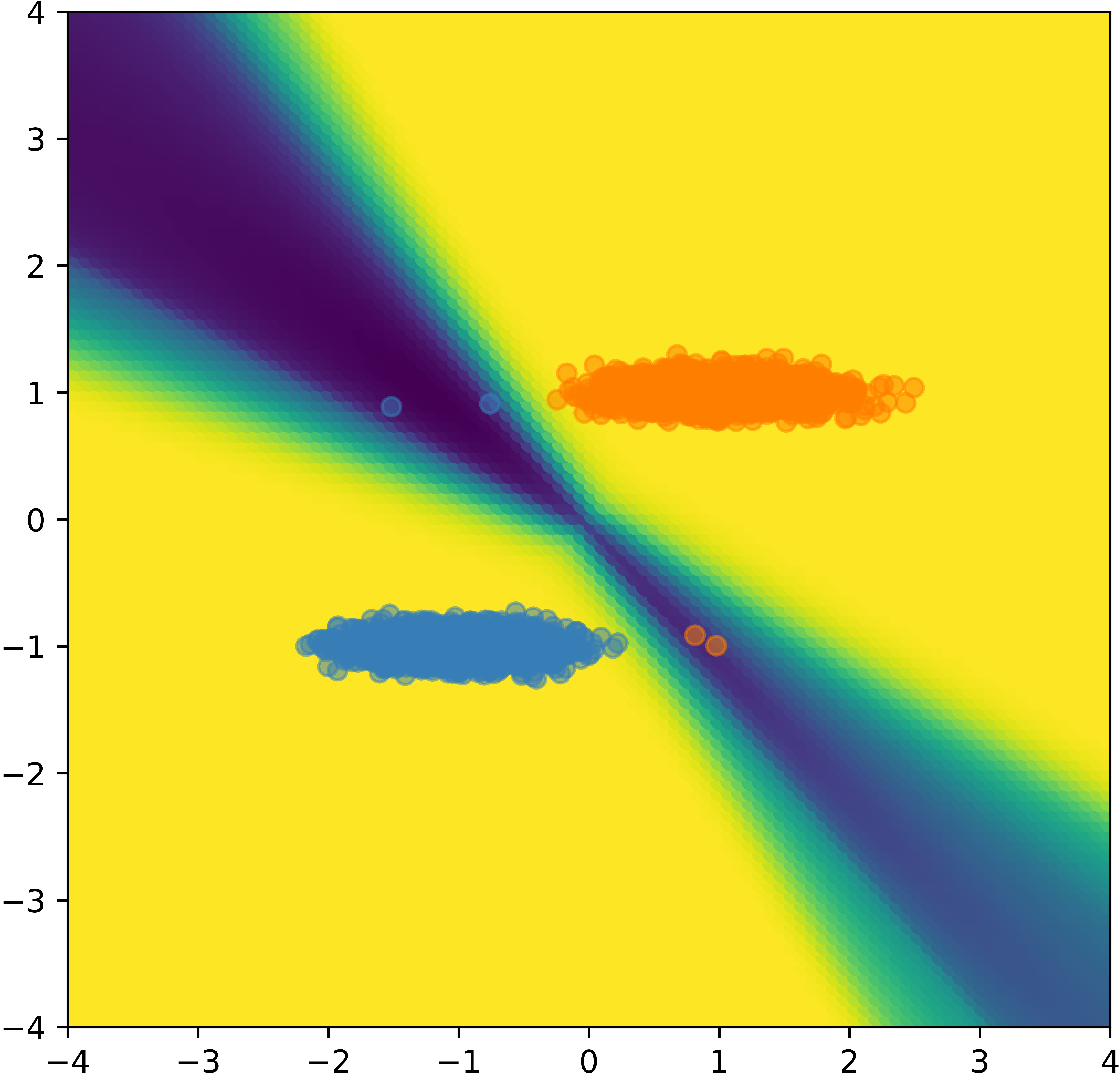}
  \caption{Predicted Underreprentation \\ \quad Introspective Training}
  \label{fig:2d_bias_isp}
\end{subfigure}

\caption{Prediction, uncertainty quantification, and representation learning behavior of introspective training v.s. ERM training in a binary classification task under severe group imbalance ($n=5000$) \citep{sagawa2020investigation}. Here, blue and orange indicates the two classes, and each class contains a minority group (the tiny clusters on the diagonal with $n<5$) and a majority group (the large clusters on the off-diagonal).
\textbf{Column 1-2} depicts the models' predictive probability and predictive uncertain surface in the data space. 
\textbf{Column 3} depicts the models' decision surface in the last-layer representation space, colored by the predictive probability of the target label. \textbf{Column 4} depicts the introspective-trained model's predicted bias probability in the representation space (\cref{fig:2d_representation_isp_bias}) and in the data space (\cref{fig:2d_bias_isp}), colored by the predictive probability of the underrepresentation.
\Cref{sec:exp_2d_app} described further detail.
}
\label{fig:2d_example}
\end{adjustbox}
\vspace{-0.5em}
\end{figure}

Despite its simplicity, introspective training has a significant impact on the model's representation learning that is particularly important for quantifying uncertainty when dataset exhibits significant bias. Figure \ref{fig:2d_example} illustrates this on a binary classification task under severe group imbalance \citep{sagawa2020investigation}, where we compare two dense ResNet ensemble models trained using the introspection objective v.s. the \gls{ERM} objective (i.e., only use $L(y_i, f_y(\rvx_i))$ in \Cref{eq:introspective_objective}), respectively.  

Comparing figures 2a and 2e, we observe that the decision boundaries for the predicted label are very similar between introspective training and \gls{ERM}. However, the predictive variance (obtained via a \gls{GP} layer \citep{liu2022simple}) exhibits sizable differences. In particular, the variance estimates for introspective training are uniformly high outside of the two clouds of underrepresented groups in the data. However, for \gls{ERM}, the model confidence is high along the decision boundary, even in the unseen regions without training data. This is due to the fact that when training with \gls{ERM}, the representation collapses in the direction that is not correlated with training label (i.e., parallel to decision boundary) and does not retrain any input information regarding the underrepresented groups in its representation (\cref{fig:2d_representation_base}). However, with introspective training, the representations indeed are morphed to reflect the differences between the underrepresented examples and the majority group (as can be seen in figures \cref{fig:2d_representation_base} vs \cref{fig:2d_representation_isp}), helping the model to better distinguish them in the representation space, and hence lead to improved uncertainty estimate in the neighborhood of underrepresented examples. \Cref{sec:exp_2d_app} contains further description.

Formally, introspective training induces the below guarantee on the model's \textit{bias-awareness} in its hidden representation and uncertainty estimates:
\begin{proposition}[Introspective Training induces Bias-awareness]
Denote $o_b(\rvx)=p(\rvx|b=1)/p(\rvx|b=0)$ the odds for $\rvx$ belongs to the underrepresented group $\gB$.
For a well-trained model $f=(f_y, f_b)$ that minimizes the introspective training objective (\ref{eq:introspective_objective}), so that $p(b=1|\rvx)=\sigma(f_b(\rvx))$, we then have:
\begin{itemize}[topsep=0pt, leftmargin=0.8cm]
\item [(I)] \textbf{(Bias-aware Hidden Representation)} The hidden representation $h(\rvx)$ is aware of the likelihood ratio of whether an example $\rvx$ belongs to the underrepresented group $b=I(g \in \gB)$, i.e. $p(\rvx|b=1)/p(\rvx|b=0)$, such that:
\vspace{-0.5em}
\begin{align}
\beta_b^\top h(\rvx) + b_b = \log \, o_b(\rvx) + \log \frac{p(b=1)}{p(b=0)}.
\label{eq:bias_aware_representation}
\end{align}
\vspace{-1.5em}

\item [(II)] \textbf{(Bias-aware Embedding Distance)} 
For two examples $(\rvx_1, \rvx_2)$, the embedding distance $||h(\rvx_1)-h(\rvx_2)||_2$ is lower bounded by (up to a scaling constant) the odds ratio of whether $\rvx_1$ belongs to the underrepresented groups versus that for $\rvx_2$:
\vspace{-0.5em}
\begin{align}
||h(\rvx_1)-h(\rvx_2)||_2 \geq \frac{1}{||\beta_b||_2} \times 
\max\Big(
\log \frac{o_b(\rvx_1)}{o_b(\rvx_2)},
\log \frac{o_b(\rvx_2)}{o_b(\rvx_1)}
\Big),
\label{eq:bias_aware_distance}
\end{align}
\vspace{-1.5em}

such that the distance between a pair of minority and majority examples $(\rvx_1, \rvx_2)$ is large due to the high values of the log odds ratio.
\end{itemize}
\label{thm:bias_awareness}
\end{proposition}
\vspace{-0.5em}

The proof is in \Cref{sec:bias_awareness_proof}. Part (I) provides a consistency guarantee for the hidden representation $h(\cdot)$'s ability in expressing the likelihood of whether an example $\rvx$ belongs to the underrepresented group $\gB$ , i.e., \textit{bias awareness}. The form of (\ref{eq:bias_aware_representation}) is similar to the representation learning guarantee in the noise contrastive learning literature, as it shares the same underlying principle of encouraging feature diversity and disentanglement via contrastive comparison between groups
\citep{gutmann2010noise, sugiyama2010density, hyvarinen2016unsupervised}
. Part (II) is a corollary of (\ref{eq:bias_aware_representation}) and provides a direct guarantee on the model's learned embedding distance. It states that under introspective training, the model cannot discard important input features that distinguishes the minority-group examples from the majority group, even if they are not predictive of the target label. In this way, the model is guarded from collapsing the representation of majority and minority examples together (i.e., making $||h(\rvx_1)-h(\rvx_2)||_2$ excessively small for two examples $(\rvx_1, \rvx_2)$  from the majority and minority group, respectively), creating difficulty for identifying underrepresented groups in the feature space with uncertainty-based active learning. Empirically, we find the benefit of introspective training extends to other uncertainty-based active learning signals as well (e.g., margin and ensemble diversity, see \Cref{sec:exp_ablation}). \Cref{sec:theory_discussion} contains further discussion.

\subsection{Estimating unknown bias via  Cross-validated Self-play}
\label{sec:bias_estimation}

So far, we introduced \textit{introspective training} in the setting where the group annotation $g_i$ is available on training data, so that the underrepresentation label $b_i=I(g_i \in \gB)$ can be directly computed. 
In this section, we consider how to estimate the underrepresentation label $b_i$ when it is absent, so that \gls{ISP} can be applied to the setting where group annotations $g_i$ is too expensive to obtain. A popular practice in the literature is to estimate dataset bias as the predictive error of a single (biased) model. That is, given a trained model $f_D$, prior work \citep{clark2019don, he2019unlearn, nam2020learning, sanh2020learning, liu2021just} estimates the underrepresentation label as the observed error $L(y_i, f_D(\rvx_i))$. To better understand this estimator for the generalization error of the underrepresented groups, we perform a noise-bias-variance decomposition (\citet{domingos2000unified}) of the model error $L(y, f_D)$, which reveals, in the expectation of the random draws of the dataset $D \sim \gD$:
\begin{align}
\underbrace{E_D[L(y, f_D(\rvx))]}_{error} = 
{\color{OliveGreen} 
\underbrace{E_D \big[ L(y, \tilde{y}(\rvx)) \big]}_{noise}
}
\quad + \quad
{\color{Maroon}
\underbrace{L(\tilde{y}(\rvx), \bar{f}(\rvx))}_{bias}
}
\quad + \quad
{\color{RoyalBlue}
\underbrace{E_{D} \big[ L(\bar{f}(\rvx), f_D(\rvx)) \big]}_{variance},
}
\label{eq:bias_estimator_naive}
\end{align}
where $\tilde{y}(\rvx)=\argmin_{y'} E_{y \sim P(y|\rvx)}[L(y, y')]$ is the (Bayes) optimal predictor and $\bar{f}(\rvx)=\argmin_f E_D[L(f, f_D(\rvx))]$ is the `ensemble' predictor of the single models $\{f_D\}_{D\sim \gD}$ trained from random data draws (see \Cref{sec:error_decomposition} for a review).
From (\ref{eq:bias_estimator_naive}), we see that for the purpose of estimating generalization error due to dataset bias, the naive estimator $\hat{b}_0 = L(y, f_D)$ based on single-model error suffers from two issues: 
(1) $\hat{b}_0$ conflates {\color{OliveGreen} \textit{noise}} (typically arising from label noise or feature ambiguity) with the dataset bias signal we wish to capture, potentially leading to compromised quality in real datasets
\citep{lahoti2020fairness, li2022more}. (2) As $\hat{b}_0$ is calculated from a single model, its estimate of the  {\color{RoyalBlue} \textit{variance}} term (an important component of generalization error
\citep{yang2020rethinking}) is often not stable. This is exacerbated when $\hat{b}_0$ is computed from the training error, since
\gls{DNN}s tend to severely underestimate the model variance on training data \citep{liu2021just}.\footnote{As an illustrative example, the generalization error of a ridge regression model under orthogonal design and group-specific noise is $E_D(L(y, f_D(\rvx_g)))={\color{OliveGreen} \sigma_g^2} + {\color{Maroon} \frac{(\lambda\theta_g)^{ 2}}{(n_g + \lambda)^2}} + {\color{RoyalBlue} \frac{\sigma^2 n_g}{(n_g + \lambda)^2}}$, where $\sigma_g$ is the noise level for group $g\in \gG$, $n_g$ is the sample size for group $g\in \gG$, and $\lambda$ is the ridge regularization parameter. See \Cref{sec:error_decomposition_ridge} for details.}


This observation motivates us to propose \textbf{\textit{cross-validated self-play}}, a simple method to estimate a  model's generalization gap. Briefly, given training data $D$ divided into $K$ splits, we train a bootstrap ensemble of $K$ models $\{f_k\}_{k=1}^K$ with \gls{ERM} training, where each $f_k$ sees a fraction of the training data (see Appendix Fig. \ref{fig:self_play_ensemble}). As a result, for each $(\rvx_i, y_i)$, there exists a collection of in-sample predictions $\{f_{in,k'}(\rvx_i)\}_{k'=1}^{K_{in}}$ trained on data splits containing $(\rvx_i, y_i)$, and a collection of out-of-sample predictions $\{f_{out,k}(\rvx_i)\}_{k=1}^{K_{out}}$ trained on data splits not containing $(\rvx_i, y_i)$. 
Then, the \textbf{\textit{self-play estimator}} of the model's generalization gap is
\footnote{In this work, we use mean squared error $L(y, f)=\sqrt{(y - \sigma_{sigmoid}(f))^2}$ for the generalization gap computation, so that $\hat{b}_i \in [0, 1]$.}
\begin{align}
\hat{b}_i 
&= 
\underbrace{\E_k[L(y_i, f_{out,k}(\rvx_i))]}_{estimated \; error} - 
{\color{OliveGreen}
\underbrace{L(y_i, \bar{f}_{in}(\rvx_i))}_{estimated \; noise}}
=
\E_k[L(\bar{f}_{in}(\rvx_i), f_{out,k}(\rvx_i))]
.
\label{eq:bias_estimator}
\end{align}
where $\bar{f}_{in}$ is the ensemble prediction based on in-sample predictors $f_{in,k'}$, the expectation $\E_k$ is taken with respect to the out-of-sample predictions, and we are estimating the Bayes optimal predictor $\tilde{y}$ using the in-domain prediction $\bar{f}_{in}$ (since the model class $\gF$ is subject to suitable regularization,  the $\bar{f}_{in}$'s do not arbitrarily overfit the noisy labels).
Note that under the well-specified setting, $\bar{f}_{in}$ converges asymptotically to the Bayes optimal predictor $\tilde{y}$ as $n \rightarrow \infty$. In practice, we can ensure the validity of $L(y_i, \bar{f}_{in}(\rvx_i))$ as an estimator of noise by applying early stopping with a stability criterion based on out-of-sample predictions  $\bar{f}_{out}(\rvx_i)$  \citep{li2020gradient, song2020prestopping}.
Compared to the standard alternatives in the literature (e.g., single-model error $L(y, f_D)$), the \textit{self-play} estimator $\hat{b}_i$ has the appealing property of controlling {\color{OliveGreen} \textit{noise}} (by using $\bar{f}_{in}$) while better estimating {\color{RoyalBlue} \textit{variance}} (by using expectations over $\bar{f}_{out, k}$), thereby constituting a more informative signal for the underrepresented groups under dataset bias, label noise and feature ambiguity. \Cref{sec:cv_ensemble} contains additional comments, {\color{black} and \Cref{sec:detection_guanrantee_app} develops a data-dependent bound for group detection performance.}

\begin{figure}[ht]
\centering
\includegraphics[trim=0.5cm 16cm 0.5cm 1cm, width=0.75\columnwidth]{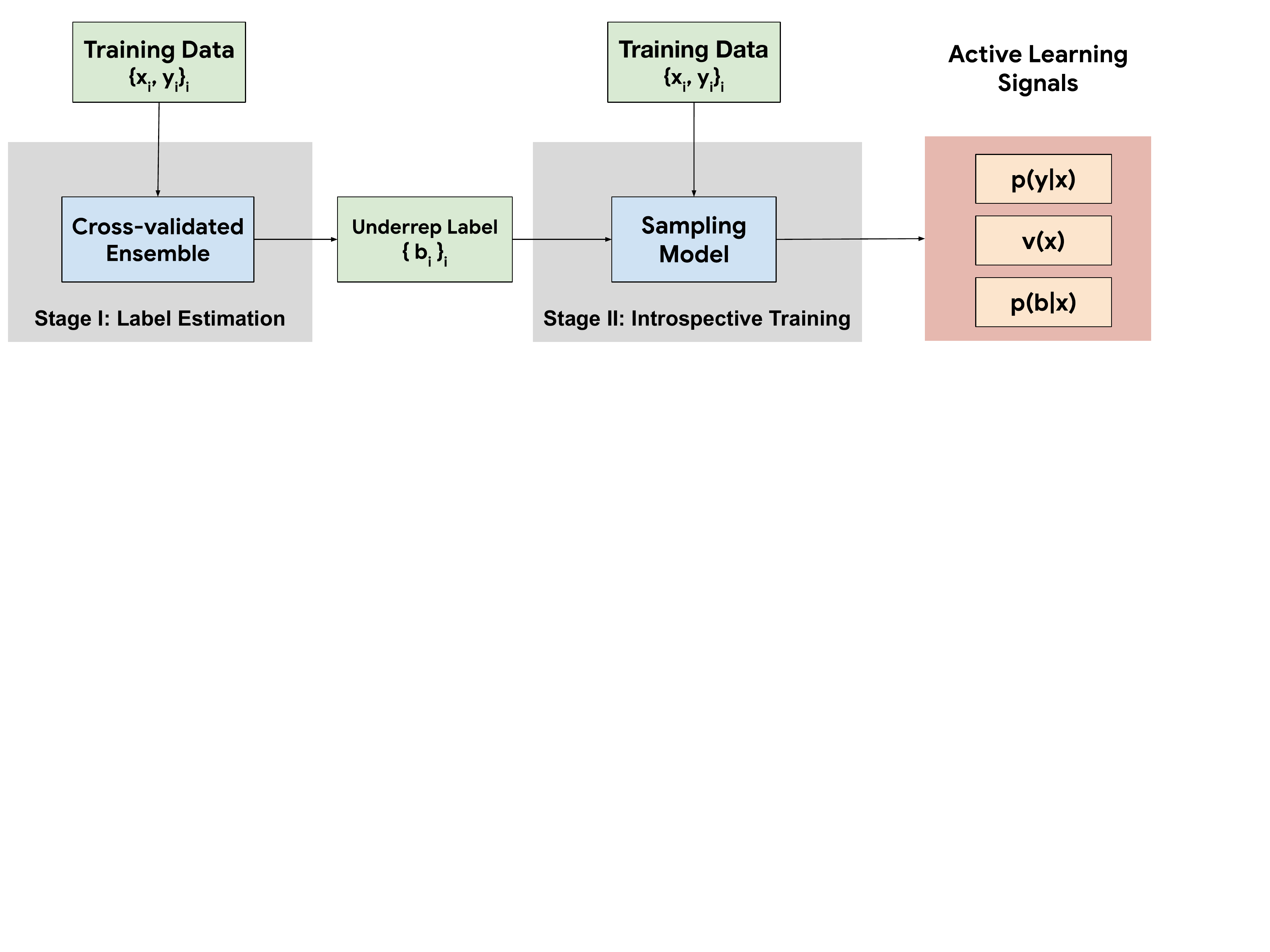}
\caption{
The two-stage \glsfirst{ISP} model: In the first stage, we first train a bootstrap ensemble of \gls{DNN}'s on cross-validation splits of the data (\Cref{fig:self_play_ensemble}). At the end of Stage I training, we can use the in-sample and out-of-sample predictions of the ensemble to estimate the underrepresentation label $b_i$ of each training data point $\rvx_i$ by computing the generalization gap (\Cref{eq:bias_estimator}). Then in Stage II, we use the underrepresentation labels $b_i$ to train the actual sampling model via introspective training (\Cref{eq:introspective_objective}), and use it to generate the active learning signals (e.g., predictive probability $p(y|x)$, predictive variance $v(x)$ and predicted underrepresentation $p(b|x)$) for the unlabelled set.
}
\vspace{-1em}
\label{fig:introspective_self_play}
\end{figure}

\textbf{Method Summary: Introspective Self-play.} Combining the \textit{self-play bias estimation} and \textit{introspective training} together, we arrive at \glsfirst{ISP}, a simple two-stage method that provably improves the representation quality and uncertainty estimates of a \gls{DNN} for underrepresented population groups. \Cref{fig:introspective_self_play} illustrates the full \gls{ISP} procedure. \gls{ISP} first (optionally) estimates underrepresentation labels using \textit{cross-validated self-play} if the group annotation is not available, and then conducts \textit{introspective training} to train the model to recognize its own bias while learning to predict the target label. For the unlabelled data to be sampled, the resulting model generates (1) predictive probability $p(y|\rvx)$, (2) uncertainty estimates $\hat{v}(\rvx)$ and (3) predicted probability for underrepresentation $p(b|\rvx)=\sigma(f_b(\rvx))$, offering a rich collection of active learning signals for downstream applications. 

\section{Theoretical Analysis: 
Improving Accuracy-{\color{black} Group Robustness} Frontier by Optimizing Training Data Distribution 
}
\label{sec:theory}

Denote by $\Delta^K$ the $K$-simplex so that $\forall \alpha=[\alpha_1, \dots, \alpha_K] \in \Delta^K$,  $\sum_k \alpha_k = 1$ and $\forall k$, $\alpha_k > 0$. Let $D_{\alpha, n}$ denote a dataset of size $n$ sampled with group allocations $\alpha \in \Delta^{|\gG|}$, i.e., $\alpha_g$ represents the fraction of the dataset sampled from group $g \in \gG$. Our goal is to find the optimal allocation that minimizes a weighted combination of the population risk and the worst-case risk over all subgroups.

\begin{definition}[Risk and Fairness Risk] Given $f \in \gF$, let $R(f|\alpha, n)\coloneqq \sum_{g\in \gG} \gamma_g R(f|G=g)$ be the risk where $\gamma \in \Delta^{|\mathcal{G}|}$ represents the true subgroup proportions in the underlying data distribution, and $\Rfair(f)$ be the fairness risk which is the worst-case risk among the groups $g\in \gG$: 
$\Rfair(f) \coloneqq \max_g R(f|G=g)$. 
\vspace{-1em}

\end{definition}


Let $\hat{R}$ and $\Rfairhat$ denote the empirical estimates of $R, \Rfair$ estimated based on the finite dataset $D_{\alpha, n}$:
\[\hat{R}_{\alpha, n, g}\br{f} = \frac{\sum_{(x, y) \in D_{\alpha, n, g}} L\br{f\br{x}, y}}{|D_{\alpha, n, g}|} , \hat{R}_{\alpha, n}\br{f}=\sum_g \gamma_g \hat{R}_{\alpha, n, g}\br{f}, \Rfairhatan\br{f} = \max_g \hat{R}_{\alpha, n, g}\br{f}\]

where $D_{\alpha, n, g}$ is the subset of $D_{\alpha, n}$ with group label $g$. Let $\hat{f}$ denote the classifier obtained as per $\gamma$-weighted empirical risk minimization (ERM)
$\hat{f}_{\alpha, n} = \argmin_{f \in \gF} \hat{R}_{\alpha, n}\br{f}$. 

\begin{definition}[Accuracy-Fairness Frontier Risk]
For non-negative weighting coefficients $\omega \in [0, 1]$, the accuracy-fairness frontier risk is defined as:
\begin{align}
F_{\omega}(\alpha, n) \coloneqq 
\omega E[R(\hat{f}_{\alpha, n})] + (1-\omega) E[\Rfair(\hat{f}_{\alpha, n})]
\label{eq:frontier_risk}
\end{align}
where the expectation is taken wrt the randomness inherent in $\hat{f}$ as it is estimated based on a random dataset $D_{\alpha, n}$ drawn iid from the underlying data distribution.  
\vspace{-1em}
\label{def:fairness_accuracy_frontier}
\end{definition}
As we sweep over $\omega$ going from $0$ to $1$ and find the best classifier/allocation for each value, we would trace an accuracy fairness frontier as in \Cref{fig:acc_fair_frontier_intro}.

\textbf{Analyzing optimal allocation between subgroups.} We theoretically analyze the optimization problem
$\alpha^* = 
 \argmin_{\alpha \in \Delta^{|\gG|}} 
\Big[ F_{\omega}(\alpha, n) \Big].$
We build on the theoretical models for group-specific scaling laws of generalization error introduced in previous work. \citep{chen2018my, rolf2021representation}. We assume that the group-specific true risk decays with the size of the dataset at a rate of the form:
\[E[R\br{\hat{f}_{\alpha, n}|G=g}]=c_g \br{\alpha_g n}^{-p} + \tau n^{-q} + \delta\]
for some $p >0, q > 0, c_g > 0, \tau, \delta > 0$. The first term represents the impact of the group representation, the second  the aggregate impact of the dataset size, and the third term a constant offset (the irreducible risk for this group). In particular, $c_g$ represents the ``difficulty`` of learning group $g$, as the larger $c_g$ is, the higher the risk is for group $g$ for any given allocation $\alpha$.

\begin{theorem}[Optimal Group-size Allocation for Accuracy-Fairness Frontier Risk (Informal)]
The optimal allocation $\alpha^\star$ is of the form $\alpha_g^\star \propto \br{\gamma_g c_g \theta_g}^{\frac{1}{p+1}}$ where
$\theta_g \geq \omega$ represents an up-sampling factor for group $g$. Let $g_1, g_2, \ldots$ be the subgroups sorted in ascending order according to the value $\gamma_g c_g^{-1/p}$, which represents the subgroup representation normalized by the difficulty of learning the subgroup. Then, it holds that there exists an integer $k > 0$ such that $\theta_{g_i} = \omega$ for $i = k, \ldots |\gG|$, $\theta_{g_i} > \omega$ for  $i= 1, \ldots, k$. Thus, the subgroups with low normalized representation are systematically up-sampled in the optimal allocation. 
\label{thm:optimal_allocation_frontier_main}
\end{theorem}

Based on the above theorem, we can the set of underrepresented groups as 
$\mathcal{B}=\{g_1, g_2, \ldots g_k\}.$, validating the idea that underrepresented groups can be formally defined. While the result above is derived under a simplified theoretical model, it validates the intuition behind ISP: indeed, ISP attempts to infer whether datapoints belong to $\mathcal{B}$ and systematically up-samples them via active learning to increase the overall representation and achieve an allocation closer to the theoretically optimal allocation defined above. We present a full theorem statement and proof in \Cref{sec:optimal_allocation_frontier_proof} and a discussion in \Cref{sec:optimal_allocation_frontier_discussion}

\section{Experiments}
\label{sec:exp}

We first demonstrate that for each task, \gls{ISP} meaningfully improves the tail-group sampling rate and the accuracy-{\color{black} group robustness} performance of state-of-the-art \gls{AL} methods (\Cref{sec:exp_results}), and then conduct detailed ablation analysis to understand how the choice of different underrepresentation labels impacts (1) the final model's accuracy-{\color{black} group robustness} performance when trained on data collected by different \gls{AL} methods; and (2) the sampling performance of different active sampling signals  (\Cref{sec:exp_ablation}).\\
\textbf{Datasets.} We consider two challenging real-world datasets: \underline{Census Income} \citep{le2022survey} that contains 32,561 training records from the 1994 U.S. census survey. The task is to predict whether an individual's income is $>$50K, and the tail groups are female or non-white individuals with high income. We also consider \underline{Toxicity Detection} \citep{borkan2019nuanced} that contains 405,130 online comments from the CivilComments platform. The goal is to predict whether a given comment is toxic, and the tail groups are demographic identities $\times$ label class (male, female, White, Black, LGBTQ, Muslim, Christian, other religion) $\times$ (toxic, non-toxic) following \citet{koh2021wilds}.

\begin{table}[ht]
\centering
\resizebox{0.8\columnwidth}{!}{%
\begin{tabular}{c|ccc|l}
\toprule
\gls{AL} Training & Group identity label & Training Mechanism & Underrepresentation & \multicolumn{1}{|c}{Available Sampling Signal}\\
Method & in train set? & & Label $b_i$ & 
\\\hline\hline
(Random) & $\checkmark$ & -  & Group Identity & Random
\\
RWT \citep{idrissi2022simple} & $\checkmark$ & Reweighting  & Group Identity & Margin / Diversity / Variance
\\
DRO \citep{sagawa2019distributionally} & $\checkmark$ & Worst-group Loss  & Group Identity & Margin / Diversity / Variance
\\\hline 
\gls{ISP}-Identity &  $\checkmark$ & {\color{Maroon}Introspection}  & Group Identity & Margin / Diversity / Variance / {\color{Maroon}Predicted Underrep.}
\\\toprule\toprule
(ERM) & $\times$ & -  & Train Error &  Margin / Diversity / Variance
\\
JTT \citep{liu2021just} & $\times$ & Reweighting & Train Error & Margin / Diversity / Variance
\\\hline
ISP - Gap & $\times$ & {\color{Maroon}Introspection}  & {\color{Maroon}Generalization Gap} & Margin / Diversity / Variance / {\color{Maroon}Predicted Underrep.}
\\\bottomrule
\end{tabular}
}
\caption{
Training methods to be compared in the experiment study.
Components proposed in this work are highlighted in {\color{Maroon}red}. The two baselines \textbf{(Random) $\&$ (ERM)} does not use underrepresentation label to train AL model, and only use it as reweighting signal for the reweighted training of the final model.
For detailed definition of the sampling signals, see \Cref{sec:exp_ml_app}.
}
\label{tab:methods_main}
\vspace{-1.2em}
\end{table}

\textbf{\gls{AL} Baselines and Method Variations.} For all tasks, we use a 10-member \gls{DNN} ensemble $f=\{f_k\}_{k=1}^{10}$ as the \gls{AL} model, and replace their last layers with a random-feature \gls{GP} layer \citep{liu2022simple} in order to compute posterior variance (see  \Cref{sec:deep_uncertainty}). We compare the impact of different training methods in two settings depending on whether the group identity label will be annotated in the labelled set (they are \textit{never} available in the unlabelled set). As shown in \Cref{tab:methods_main}, when group label is available, we compare \gls{ISP}-identity (i.e., \gls{ISP} with group identity as training label $b_i=I(g_i \in \gB)$) to a group-specific reweighting (RWT) \citep{idrissi2022simple} and a group DRO \citep{sagawa2019distributionally} baselines \citep{idrissi2022simple}
When the group label is not known, 
we consider \textbf{\gls{ISP}-Gap} using the \textit{self-play}-estimated generalization gap $\hat{b}_i=\E_k[L(\bar{f}_{in}(\rvx_i), f_{out,k}(\rvx_i))]$ as the representation label (i.e., \Cref{eq:bias_estimator}), and compare it to an ensemble of \gls{JTT} which uses the ensemble training error $\hat{b}_i=\E_k[L(y_i, f_{in,k}(\rvx_i))]$ to determine the training set. We also compare to an \gls{ERM} baseline which trains the \gls{AL} models with a routine ERM objective, but uses error for the reweighted training of the final model. We consider other method combinations in the ablation study.
\\
\textbf{Active Learning Protocol and Final Accuracy-{\color{black} Group Robustness} Evaluation.}
For active learning, we start with a randomly sampled initial dataset, and conduct active learning for 8 rounds until reaching roughly half of the training set (to ensure there's sufficient variation in the sampled data between methods). To evaluate the accuracy-{\color{black} group robustness} performance of the final model, given a dataset collected by an \gls{AL} method, we train a final model using the standard re-weighting objective $\sum_{(x,y)\not\in \hat{\gB}} L_{ce}(y, f(\rvx)) + \lambda \sum_{(x,y)\in \hat{\gB}} L_{ce}(y, f(\rvx))$ where $\hat{\gB}$ is the set of underrepresented examples identified by the underrepresentation label, i.e., $(x_i,y_i) \in \hat{\gB}$ if $\hat{b}_i > t$. We vary the thresholds $t$ and the up-weight coefficient $\lambda$ over a 2D grid to get a collection of model accuracy-{\color{black} group robustness} performances (i.e., accuracy v.s. worst-group accuracy), and use them to identify the Pareto frontier defined by this combination of data and reweighting signal. 
\Cref{sec:exp_ml_app} describes further detail.\\
\vspace{-2em}
\begin{table}[ht]
\centering
\resizebox{1.0\textwidth}{!}{%
\begin{tabular}{cccccccc}
\toprule
\multirow{2}{*}{\shortstack{\gls{AL} Training \\ Method}} & 
\multirow{2}{*}{\shortstack{Group identity label \\ in train set?}} & 
\multicolumn{3}{c}{Census Income} & 
\multicolumn{3}{c}{Toxicity Detection}
\\\cmidrule(lr){3-5} \cmidrule(lr){6-8}
&  & 
Tail Sampling Rate & Combined Acc. & Worst-group Acc. & 
Tail Sampling Rate & Combined Acc. & Worst-group Acc. 
\\\midrule
(Random)  & $\checkmark$ &  
0.475 & 0.746 & 0.659 & 
0.556 & 0.708 & 0.490 
\\
RWT & $\checkmark$ &  
0.797 & 0.772 & 0.761 & 
0.857 & 0.709 & 0.482 
\\
{\color{black} DRO} & $\checkmark$ &  
{\color{black} 0.755} & {\color{black} 0.759} & {\color{black} 0.729} & 
{\color{black} 0.841} & {\color{black} 0.710} & {\color{black} 0.506}
\\
\gls{ISP}-Identity (Ours) & $\checkmark$ &  
\cellcolor{gray!25}\textbf{0.907} & 
\cellcolor{gray!25}\textbf{0.785} & 
\cellcolor{gray!25}{\color{black}\textbf{0.774}} & 
\cellcolor{gray!25} \textbf{0.905} & 
\cellcolor{gray!25} \textbf{0.719} & 
\cellcolor{gray!25} \textbf{0.506} 
\\\toprule\toprule
ERM  & $\times$ &  
0.791 & 0.736 & 0.658 & 
0.852 & 0.735 & 0.539
\\
JTT & $\times$ &  
\cellcolor{gray!25}\textbf{0.839} & 0.752 & 0.695 & 
0.866 & 0.747 & 0.571
\\
\gls{ISP}-Gap (Ours) & $\times$ &  
\cellcolor{gray!25}\textbf{0.839} & 
\cellcolor{gray!25}\textbf{0.770} & 
\cellcolor{gray!25} {\color{black}\textbf{0.753}} &
\cellcolor{gray!25} \textbf{0.867} & 
\cellcolor{gray!25} \textbf{0.759} & 
\cellcolor{gray!25} \textbf{0.597}
\\\bottomrule
\end{tabular}
}
\caption{The tail-group sampling rate and final-model accuracy v.s. {\color{black}group robustness} performances under different AL model training methods. Here we show the best active learning signal for each task (i.e., variance for Census Income, and margin for toxicity detection).
\textbf{Tail Sampling Rate}: The ratio between num. of sampled tail group examples (in final round) v.s. the total num. of tail group in population. \textbf{Combined Acc}: The combined accuracy-{\color{black}robustness} score defined as (acc + worst-group acc)/2. It is proportional to the perimeter of the rectangle defined by a point on the accuracy-{\color{black}group robustness} curve.}
\label{tab:result_main}
\vspace{-2em}
\end{table}

\subsection{Main Results}
\label{sec:exp_results}

\begin{wrapfigure}[8]{r}{0.25\textwidth}
\vspace{-4em}
\centering
\includegraphics[width=0.25\textwidth]{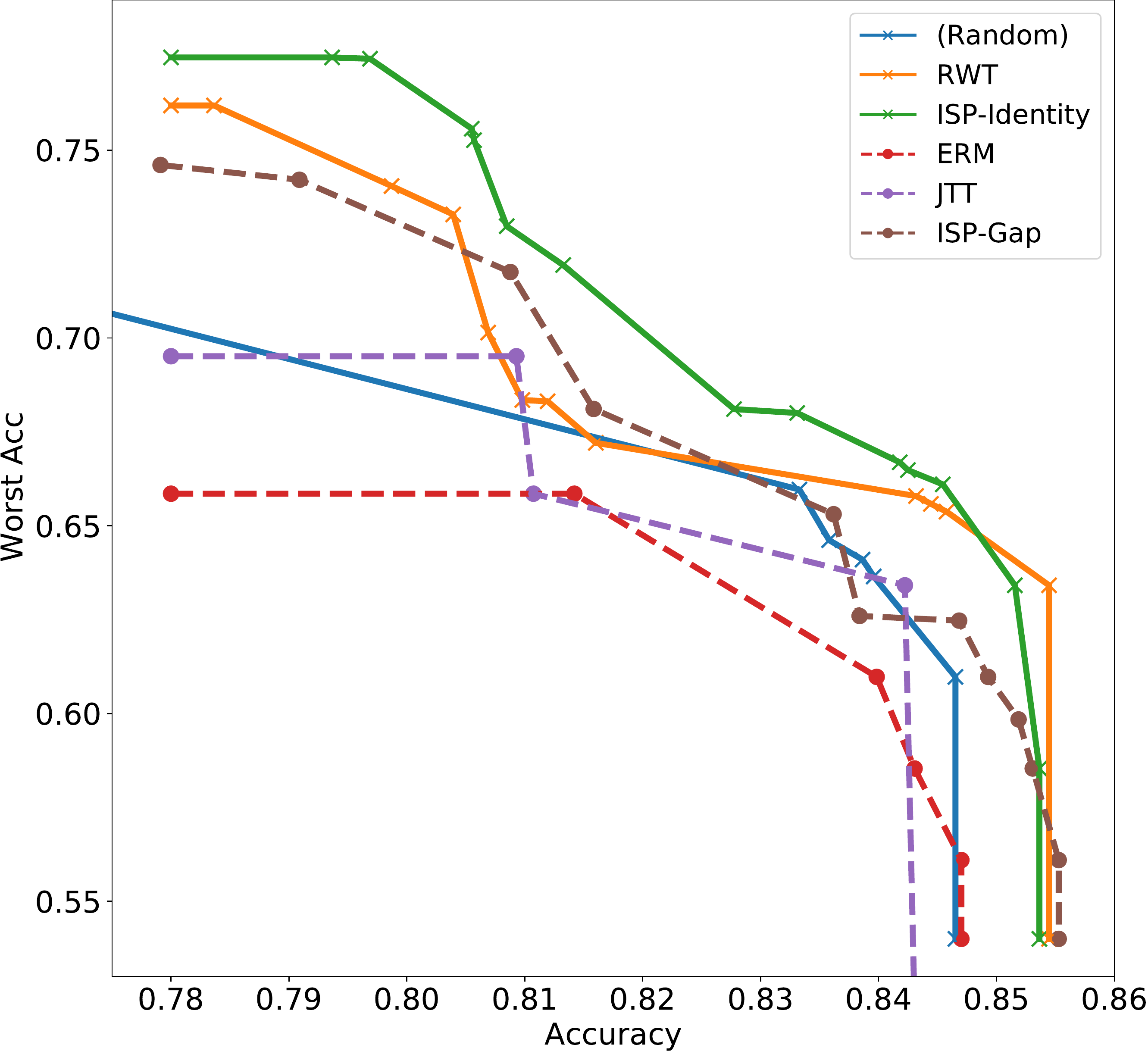}
\vspace{-2em}
\caption{Accuracy-{\color{black}Group Robustness} Frontier for Census Income.}
\label{fig:acc_fair_curves_exp_main}
\end{wrapfigure}

\Cref{tab:result_main} shows sampling performance and the final-model {\color{black} group robustness}-accuracy performance of each \gls{AL} model training method (described in \Cref{tab:methods_main}), and 
\Cref{fig:acc_fair_curves_exp_main} visualizes the full accuracy-fairness frontier of the final models (trained on the data and re-weighting signals provided by that method).
Our main conclusions are: \textbf{(1) Effectiveness of \gls{ISP} training}: Compared to non-\gls{ISP} baselines, we find \gls{ISP} consistently improves a \gls{AL} model's active learning (measured by tail-group sampling rate) and accuracy-{\color{black}group robustness} performance (measured by combined accuracy, which is defined as (accuracy + worst-group accuracy)/2). This advantage is seen in both settings where the group label is available or unavailable. In particular, in \Cref{fig:acc_fair_curves_exp_main}, the final model from \textbf{\gls{ISP}-Gap} (pink dashed line, trained on actively sampled data and using estimated underrepresentation label for final-model re-weighted training) almost dominates \textbf{Random} (blue solid line, trained on randomly sampled data and using true group label for final model training) despite not having access to true group label in the final reweighted training, highlighting the importance of the data distribution in the model's accuracy-{\color{black}group robustness} performance (i.e., \Cref{eq:al_objective_intro}). \textbf{(2) Label Quality Matters}: Comparing the variants of \gls{ISP} (Identity v.s. Gap) in \Cref{tab:result_main}, we see a clear impact of the quality of introspection signal to the performance of the \gls{AL} model. For example, for active learning performance, we see that the sampling rate \gls{ISP}-Identity is significantly better than \gls{ISP}-Error. However, for toxicity detection where the group label suffers an under-coverage issue (i.e., the group definition excludes potentially identity-mention comments where raters disagree, see \textbf{Data} section of \Cref{sec:exp_ml_app}), we see that \gls{ISP}-Error in fact strongly outperforms \gls{ISP}-Identity in accuracy-{\color{black}group robustness} performance. This validates the observation from previous literature on the failure mode of bias-mitigation methods when the available group annotation does not cover all sources of dataset bias, and speaks to the importance of high quality estimation methods that can detect underpresentation in the presence of unknown sources of bias \citep{zhou2021examining}.

\subsection{Ablation Analysis}
\label{sec:exp_ablation}
In the main results above, we have (1) used the same underrepresentation label for both the \gls{AL}-model introspective training and the final-model reweighted training, and (2) focused on the most effective active sampling signal under each task. In this section, we conduct ablations by decoupling the signal combinations along these two axes. \\
\textbf{Impact of Data Distribution and Reweighting Signal to Accuracy-{\color{black}Group Robustness} Frontier} 
First, we investigate the joint impact of data distribution and reweighting signal on the final models' accuracy-{\color{black}group robustness} performance. We train the final model under data collected by different \gls{AL} policy (Random v.s Margin v.s. Group Identity, etc), and perform reweighted training using different underrepresentation labels (Error v.s. Gap v.s. Group Identity) and compare to an \gls{ERM} baseline without reweighted training (\Cref{tab:signal_ablation_acc_fair}). As shown, holding the choice of reweighted signal constant and compare across data distributions (i.e., comparing across columns within each row), we observe that the data distribution in general has a non-trivial impact on the final model's accuracy-{\color{black}group robustness} performance. Specifically, under appropriate sampling signal, data collected by \gls{ISP}-Gap (which has no access to true group identity label) can lead to model performance that is competitive with data collected by \gls{ISP}-Identity (e.g., the third v.s. fourth columns). 
Comparing across reweighting signals within each dataset (i.e., compare across rows within each column), we see that all underrepresentation labels brings a meaningful improvement over the \gls{ERM} baseline, with Group Identity bringing the most significant improvement when it is of high quality (i.e., Census Income), and Gap bringing the most improvement when group annotation is imperfect  (i.e., Toxicity Detection).


\begin{table}[ht]
\centering
\resizebox{0.75\textwidth}{!}{%
\begin{tabular}{ccccccccc}
\toprule
\multirow{2}{*}{\shortstack{Final Model \\ Reweighting Signal}} & 
\multicolumn{4}{c}{\gls{AL} Method, Census Income} & 
\multicolumn{4}{c}{\gls{AL} Method, Toxicity Detection}
\\\cmidrule(lr){2-5} \cmidrule(lr){6-9}
&  
Random & Margin & Variance & Group Identity &
Random & Diversity & Margin & Group Identity 
\\\midrule
(ERM) & 
0.692 & 0.669 & 0.719 & 0.720 &
0.698 & 0.699 & 0.702 & 0.703
\\\midrule
Error  & 
0.706 & 0.683 & 0.750 & 0.743 &
0.758 & 0.761 & 0.744 & 0.752
\\
Gap & 
0.692 & 0.694 & 0.770 & 0.777 &
\cellcolor{gray!25} \textbf{0.776} & \cellcolor{gray!25} \textbf{0.776} & \cellcolor{gray!25} \textbf{0.758 }& \cellcolor{gray!25} \textbf{0.810}
\\\midrule
Group Identity & 
\cellcolor{gray!25} \textbf{0.746} & \cellcolor{gray!25} \textbf{0.756} & \cellcolor{gray!25} \textbf{0.778} & \cellcolor{gray!25} \textbf{0.785} &
0.711 & 0.701 & 0.705 & 0.713
\\\bottomrule
\end{tabular}
\label{tab:signal_ablation_acc_fair}
}
\caption{Impact to final-model {\color{black}group robustness}-accuracy performance (measured by combined accuracy = acc + worst-group acc)/2) of the choice of reweighting signal (rows), across dataset collected by different active learning methods (columns). \textbf{Random}: data collected via random sampling. \textbf{Margin/Variance/Diversity}: data collected using introspective-trained AL model (with Gap as underrepresentation label) using the said sampling signal. \textbf{Group Identity}: data collected by introspective-trained AL model with group identity as introspection signal, using the best sampling signal for the task (Variance for census income and Margin for toxicity detection).}
\label{tab:signal_ablation_sampling}
\end{table}
\textbf{Impact of Underrepresentation Label on Different Sampling Signals.} Finally, we evaluate the choice of introspection signal on the sampling performance of a introspective-trained AL-model, under different types of sampling signals (\Cref{tab:signal_ablation_sampling}). As this evaluation is computationally expensive (requiring multiple active learning experiments for all underrepresentation label v.s. sampling signal combinations), here we focus on the Census Income task. As shown, 
we observe the introspective training brings a consistent performance boost across different types of sampling signals (esp. when using Group Identity), highlighting the appeal of introspective training as a ``plug-in" method that meaningfully boost the performance of a wide range of active learning methods. Interestingly, we also observe the ``Predicted Underrep." (i.e., the underrepresentation prediction in $p(b|\rvx)$ in \Cref{fig:introspective_self_play}) is exceptionally effective when the group identity is available (tail sampling rate $>$ 0.95) but underperforms classic active learning signals otherwise, cautioning the proper use of $p(b|\rvx)$ as a sampling signal depending on the availability of group labels.


\begin{table}[ht]
\centering
\resizebox{0.5\textwidth}{!}{%
\begin{tabular}{ccccc}
\toprule
\multirow{2}{*}{\shortstack{Underrep. \\ Label}} & 
\multicolumn{4}{c}{\gls{AL} Method, Census Income} 
\\\cmidrule(lr){2-5} 
&  
Margin & Diversity & Variance & Predicted Underrep.
\\\midrule
Error  & 
0.780 & 0.324 & 0.771 & 0.671
\\
Gap & 
0.803 & 0.276 & 0.839 & 0.708 
\\\midrule
Group Identity & 
0.873 & 0.330 & 0.907 & 0.967 
\\\bottomrule
\end{tabular}
}
\vspace{-0.5em}
\caption{Impact to AL performance (measured by tail sampling rate) of the choice of introspection signal (rows) across different active learning methods (columns).}
\label{tab:signal_ablation_sampling}
\vspace{-1em}
\end{table}

\section{Conclusion}
In this work, we introduced \glsfirst{ISP}, a novel training approach to improve a \gls{DNN}'s representation learning and uncertainty quantification quality under dataset bias. 
\gls{ISP} uses a multi-task \textit{introspective training} approach to encourage \gls{DNN}s to learn diverse and bias-aware features for the underrepresented groups. When underrepresented group identities are not available, \gls{ISP} bootstraps them using a novel \textit{cross-validated self-play} procedure that disentangles dataset bias from irreducible noise while also generating a more stable estimate of variance. 
Theoretical analysis reveals that the optimal per-group up-sampling factors are in fact determined by an interplay of the original group rates and the group-specific scaling factors.  Models trained on data acquired by \gls{ISP} generally surpass recent competitive baselines such as RWT and JTT on the accuracy-{\color{black}group robustness} frontier. 
\paragraph{Future Directions.} Overall, our results are a concrete step to a recent but critical effort in the community to build a more holistic perspective of model performance, addressing key challenges of robustness and equity. Future work could more thoroughly investigate the relation of the noise, bias, and variance components of generalization error to underrepresentation examples, as well as the effectiveness of introspective training under different settings including training epochs, model regularization, batch size. For example, specialized training objectives such as \gls{GCE} \citep{zhang2018generalized} or \gls{FL} \citep{lin2017focal} may improve the statistical power of different components of generalization error in detecting the underrepresented groups. On the other hand, batch size may impact the quality of the learned representation under introspective training, in a manner analogous to that of the contrastive training \citep{chen2020simple}. Our framework could also be extended to other settings such as semi-supervised learning, or incorporate other kinds of introspection signals, such as those from the interpetability or differential privacy literature.

\paragraph{Ethical Statements.} 
This work proposes novel approach to encourage \gls{DNN} models to learn diverse, \textit{bias-aware} features in model representation, for the purpose of improving \gls{DNN} model's uncertainty quantification ability under dataset bias. Our method encourages \gls{DNN}s to better identify underrepresented data subgroups during data collection, and eventually achieve a more balanced performance between model accuracy and {\color{black}group robustness} by training on a more well-balanced dataset. We evaluated the method on two already publicly available dataset and uses existing metrics in the literature. No new data is collected as part of the current study.

The technique we developed in this work is simple and general-purpose, with potentially broad appeal to various downstream applications (e.g., recommendations, NLP, etc). However, two limitations highlighted by our work is that (1) when the group annotation information is imperfect, building a bias-mitigation procedure around such annotation may lead to suboptimal performance, and (2) in the presence of noisy labels, a noisy estimate of under-representation may also compromise the performance of the procedure. Therefore, practitioner should take caution in rigorously evaluate the effectiveness of the procedure in their application, taking effort to carefully evaluate the estimation result of underrepresentation labels to ensure proper application of the technique without incurring unexpected consequences.

\subsubsection*{Acknowledgments}
The authors would like to sincerely thank Ian Kivlichan at Google Jigsaw, Clara Huiyi Hu, Jie Ren, Yuyan Wang, Tania Bedrax-Weiss at Google Research
for the insightful comments and helpful discussions.





\bibliography{references}

\begin{thebibliography}{148}
\providecommand{\natexlab}[1]{#1}
\providecommand{\url}[1]{\texttt{#1}}
\expandafter\ifx\csname urlstyle\endcsname\relax
  \providecommand{\doi}[1]{doi: #1}\else
  \providecommand{\doi}{doi: \begingroup \urlstyle{rm}\Url}\fi

\bibitem[Abernethy et~al.(2022)Abernethy, Awasthi, Kleindessner, Morgenstern,
  Russell, and Zhang]{abernethy22active}
Jacob~D Abernethy, Pranjal Awasthi, Matth{\"a}us Kleindessner, Jamie
  Morgenstern, Chris Russell, and Jie Zhang.
\newblock Active sampling for min-max fairness.
\newblock In Kamalika Chaudhuri, Stefanie Jegelka, Le~Song, Csaba Szepesvari,
  Gang Niu, and Sivan Sabato (eds.), \emph{Proceedings of the 39th
  International Conference on Machine Learning}, volume 162 of
  \emph{Proceedings of Machine Learning Research}, pp.\  53--65. PMLR, 17--23
  Jul 2022.

\bibitem[Adlam \& Pennington(2020)Adlam and Pennington]{adlam2020understanding}
Ben Adlam and Jeffrey Pennington.
\newblock Understanding double descent requires a fine-grained bias-variance
  decomposition.
\newblock \emph{Advances in neural information processing systems},
  33:\penalty0 11022--11032, 2020.

\bibitem[Agarwal et~al.(2018)Agarwal, Beygelzimer, Dud{\'\i}k, Langford, and
  Wallach]{agarwal2018reductions}
Alekh Agarwal, Alina Beygelzimer, Miroslav Dud{\'\i}k, John Langford, and Hanna
  Wallach.
\newblock A reductions approach to fair classification.
\newblock In \emph{International Conference on Machine Learning}, pp.\  60--69.
  PMLR, 2018.

\bibitem[Agarwal et~al.(2022)Agarwal, Muku, Anand, and Arora]{agarwal2022does}
Sharat Agarwal, Sumanyu Muku, Saket Anand, and Chetan Arora.
\newblock Does data repair lead to fair models? curating contextually fair data
  to reduce model bias.
\newblock In \emph{Proceedings of the IEEE/CVF Winter Conference on
  Applications of Computer Vision}, pp.\  3298--3307, 2022.

\bibitem[Amini et~al.(2019)Amini, Soleimany, Schwarting, Bhatia, and
  Rus]{amini2019uncovering}
Alexander Amini, Ava~P Soleimany, Wilko Schwarting, Sangeeta~N Bhatia, and
  Daniela Rus.
\newblock Uncovering and mitigating algorithmic bias through learned latent
  structure.
\newblock In \emph{Proceedings of the 2019 AAAI/ACM Conference on AI, Ethics,
  and Society}, pp.\  289--295, 2019.

\bibitem[Anahideh et~al.(2022)Anahideh, Asudeh, and
  Thirumuruganathan]{anahideh2022fair}
Hadis Anahideh, Abolfazl Asudeh, and Saravanan Thirumuruganathan.
\newblock Fair active learning.
\newblock \emph{Expert Systems with Applications}, 199:\penalty0 116981, 2022.

\bibitem[Arjovsky(2020)]{arjovsky2020out}
Martin Arjovsky.
\newblock \emph{Out of distribution generalization in machine learning}.
\newblock PhD thesis, New York University, 2020.

\bibitem[Arjovsky et~al.(2019)Arjovsky, Bottou, Gulrajani, and
  Lopez-Paz]{arjovsky2019invariant}
Martin Arjovsky, L{\'e}on Bottou, Ishaan Gulrajani, and David Lopez-Paz.
\newblock Invariant risk minimization.
\newblock \emph{arXiv preprint arXiv:1907.02893}, 2019.

\bibitem[Ash et~al.(2019)Ash, Zhang, Krishnamurthy, Langford, and
  Agarwal]{ash2019deep}
Jordan~T Ash, Chicheng Zhang, Akshay Krishnamurthy, John Langford, and Alekh
  Agarwal.
\newblock Deep batch active learning by diverse, uncertain gradient lower
  bounds.
\newblock In \emph{International Conference on Learning Representations}, 2019.

\bibitem[Bahng et~al.(2020)Bahng, Chun, Yun, Choo, and Oh]{bahng2020learning}
Hyojin Bahng, Sanghyuk Chun, Sangdoo Yun, Jaegul Choo, and Seong~Joon Oh.
\newblock Learning de-biased representations with biased representations.
\newblock In \emph{International Conference on Machine Learning}, pp.\
  528--539. PMLR, 2020.

\bibitem[Bayle et~al.(2020)Bayle, Bayle, Janson, and Mackey]{bayle2020cross}
Pierre Bayle, Alexandre Bayle, Lucas Janson, and Lester Mackey.
\newblock Cross-validation confidence intervals for test error.
\newblock \emph{Advances in Neural Information Processing Systems},
  33:\penalty0 16339--16350, 2020.

\bibitem[Beutel et~al.(2017)Beutel, Chen, Zhao, and Chi]{beutel2017data}
Alex Beutel, Jilin Chen, Zhe Zhao, and Ed~H Chi.
\newblock Data decisions and theoretical implications when adversarially
  learning fair representations.
\newblock \emph{arXiv preprint arXiv:1707.00075}, 2017.

\bibitem[Blitzstein \& Hwang(2015)Blitzstein and
  Hwang]{blitzstein2015introduction}
Joseph~K Blitzstein and Jessica Hwang.
\newblock \emph{Introduction to probability}.
\newblock Crc Press Boca Raton, FL, 2015.

\bibitem[Blum et~al.(1999)Blum, Kalai, and Langford]{blum1999beating}
Avrim Blum, Adam Kalai, and John Langford.
\newblock Beating the hold-out: Bounds for k-fold and progressive
  cross-validation.
\newblock In \emph{Proceedings of the twelfth annual conference on
  Computational learning theory}, pp.\  203--208, 1999.

\bibitem[Borkan et~al.(2019)Borkan, Dixon, Sorensen, Thain, and
  Vasserman]{borkan2019nuanced}
Daniel Borkan, Lucas Dixon, Jeffrey Sorensen, Nithum Thain, and Lucy Vasserman.
\newblock Nuanced metrics for measuring unintended bias with real data for text
  classification.
\newblock In \emph{Companion proceedings of the 2019 world wide web
  conference}, pp.\  491--500, 2019.

\bibitem[Boyd \& Vandenberghe(2004)Boyd and Vandenberghe]{boyd2004convex}
Stephen~P Boyd and Lieven Vandenberghe.
\newblock \emph{Convex optimization}.
\newblock Cambridge university press, 2004.

\bibitem[Branchaud-Charron et~al.(2021)Branchaud-Charron, Atighehchian,
  Rodr{\'\i}guez, Abuhamad, and Lacoste]{branchaud2021can}
Fr{\'e}d{\'e}ric Branchaud-Charron, Parmida Atighehchian, Pau Rodr{\'\i}guez,
  Grace Abuhamad, and Alexandre Lacoste.
\newblock Can active learning preemptively mitigate fairness issues?
\newblock \emph{arXiv preprint arXiv:2104.06879}, 2021.

\bibitem[Byrd \& Lipton(2019)Byrd and Lipton]{byrd2019effect}
Jonathon Byrd and Zachary Lipton.
\newblock What is the effect of importance weighting in deep learning?
\newblock In \emph{International Conference on Machine Learning}, pp.\
  872--881. PMLR, 2019.

\bibitem[Cai et~al.(2021)Cai, Gao, Lee, and Lei]{cai2021theory}
Tianle Cai, Ruiqi Gao, Jason Lee, and Qi~Lei.
\newblock A theory of label propagation for subpopulation shift.
\newblock In \emph{International Conference on Machine Learning}, pp.\
  1170--1182. PMLR, 2021.

\bibitem[Cai et~al.(2022)Cai, Encarnacion, Chern, Corbett-Davies, Bogen,
  Bergman, and Goel]{cai2022adaptive}
William Cai, Ro~Encarnacion, Bobbie Chern, Sam Corbett-Davies, Miranda Bogen,
  Stevie Bergman, and Sharad Goel.
\newblock Adaptive sampling strategies to construct equitable training
  datasets.
\newblock In \emph{2022 ACM Conference on Fairness, Accountability, and
  Transparency}, FAccT '22, pp.\  1467–1478, New York, NY, USA, 2022.
  Association for Computing Machinery.
\newblock ISBN 9781450393522.
\newblock \doi{10.1145/3531146.3533203}.

\bibitem[Cao et~al.(2019)Cao, Wei, Gaidon, Arechiga, and Ma]{cao2019learning}
Kaidi Cao, Colin Wei, Adrien Gaidon, Nikos Arechiga, and Tengyu Ma.
\newblock Learning imbalanced datasets with label-distribution-aware margin
  loss.
\newblock \emph{Advances in neural information processing systems}, 32, 2019.

\bibitem[Cao et~al.(2020)Cao, Chen, Lu, Arechiga, Gaidon, and
  Ma]{cao2020heteroskedastic}
Kaidi Cao, Yining Chen, Junwei Lu, Nikos Arechiga, Adrien Gaidon, and Tengyu
  Ma.
\newblock Heteroskedastic and imbalanced deep learning with adaptive
  regularization.
\newblock In \emph{International Conference on Learning Representations}, 2020.

\bibitem[Caton \& Haas(2020)Caton and Haas]{caton2020fairness}
Simon Caton and Christian Haas.
\newblock Fairness in machine learning: A survey.
\newblock \emph{arXiv preprint arXiv:2010.04053}, 2020.

\bibitem[Chen et~al.(2018)Chen, Johansson, and Sontag]{chen2018my}
Irene Chen, Fredrik~D Johansson, and David Sontag.
\newblock Why is my classifier discriminatory?
\newblock \emph{Advances in neural information processing systems}, 31, 2018.

\bibitem[Chen et~al.(2020{\natexlab{a}})Chen, Kornblith, Norouzi, and
  Hinton]{chen2020simple}
Ting Chen, Simon Kornblith, Mohammad Norouzi, and Geoffrey Hinton.
\newblock A simple framework for contrastive learning of visual
  representations.
\newblock In \emph{International conference on machine learning}, pp.\
  1597--1607. PMLR, 2020{\natexlab{a}}.

\bibitem[Chen et~al.(2020{\natexlab{b}})Chen, Wei, Kumar, and Ma]{chen2020self}
Yining Chen, Colin Wei, Ananya Kumar, and Tengyu Ma.
\newblock Self-training avoids using spurious features under domain shift.
\newblock \emph{Advances in Neural Information Processing Systems},
  33:\penalty0 21061--21071, 2020{\natexlab{b}}.

\bibitem[Cheng et~al.(2020)Cheng, Hao, Yuan, Si, and Carin]{cheng2020fairfil}
Pengyu Cheng, Weituo Hao, Siyang Yuan, Shijing Si, and Lawrence Carin.
\newblock Fairfil: Contrastive neural debiasing method for pretrained text
  encoders.
\newblock In \emph{International Conference on Learning Representations}, 2020.

\bibitem[Cherepanova et~al.(2021)Cherepanova, Nanda, Goldblum, Dickerson, and
  Goldstein]{cherepanova2021technical}
Valeriia Cherepanova, Vedant Nanda, Micah Goldblum, John~P Dickerson, and Tom
  Goldstein.
\newblock Technical challenges for training fair neural networks.
\newblock \emph{arXiv preprint arXiv:2102.06764}, 2021.

\bibitem[Chi et~al.(2022)Chi, Shand, Yu, Chang, Zhao, and
  Tian]{chi2022conditional}
Jianfeng Chi, William Shand, Yaodong Yu, Kai-Wei Chang, Han Zhao, and Yuan
  Tian.
\newblock Conditional supervised contrastive learning for fair text
  classification.
\newblock \emph{arXiv preprint arXiv:2205.11485}, 2022.

\bibitem[Clark et~al.(2019)Clark, Yatskar, and Zettlemoyer]{clark2019don}
Christopher Clark, Mark Yatskar, and Luke Zettlemoyer.
\newblock Don't take the easy way out: Ensemble based methods for avoiding
  known dataset biases.
\newblock \emph{arXiv preprint arXiv:1909.03683}, 2019.

\bibitem[Collier et~al.(2021)Collier, Mustafa, Kokiopoulou, Jenatton, and
  Berent]{collier2021correlated}
Mark Collier, Basil Mustafa, Efi Kokiopoulou, Rodolphe Jenatton, and Jesse
  Berent.
\newblock Correlated input-dependent label noise in large-scale image
  classification.
\newblock In \emph{Proceedings of the IEEE/CVF Conference on Computer Vision
  and Pattern Recognition}, pp.\  1551--1560, 2021.

\bibitem[Creager et~al.(2021)Creager, Jacobsen, and
  Zemel]{creager2021environment}
Elliot Creager, J{\"o}rn-Henrik Jacobsen, and Richard Zemel.
\newblock Environment inference for invariant learning.
\newblock In \emph{International Conference on Machine Learning}, pp.\
  2189--2200. PMLR, 2021.

\bibitem[de~Mathelin et~al.(2021)de~Mathelin, Deheeger, Mougeot, and
  Vayatis]{de2021discrepancy}
Antoine de~Mathelin, Fran{\c{c}}ois Deheeger, Mathilde Mougeot, and Nicolas
  Vayatis.
\newblock Discrepancy-based active learning for domain adaptation.
\newblock In \emph{International Conference on Learning Representations}, 2021.

\bibitem[Deng et~al.(2009)Deng, Dong, Socher, Li, Li, and
  Fei-Fei]{deng2009imagenet}
Jia Deng, Wei Dong, Richard Socher, Li-Jia Li, Kai Li, and Li~Fei-Fei.
\newblock Imagenet: A large-scale hierarchical image database.
\newblock In \emph{2009 IEEE conference on computer vision and pattern
  recognition}, pp.\  248--255. Ieee, 2009.

\bibitem[Diana et~al.(2021)Diana, Gill, Kearns, Kenthapadi, and
  Roth]{diana2021minimax}
Emily Diana, Wesley Gill, Michael Kearns, Krishnaram Kenthapadi, and Aaron
  Roth.
\newblock Minimax group fairness: Algorithms and experiments.
\newblock In \emph{Proceedings of the 2021 AAAI/ACM Conference on AI, Ethics,
  and Society}, pp.\  66--76, 2021.

\bibitem[Domingos(2000)]{domingos2000unified}
Pedro Domingos.
\newblock A unified bias-variance decomposition and its applications.
\newblock In \emph{17th International Conference on Machine Learning}, pp.\
  231--238, 2000.

\bibitem[Du et~al.(2021)Du, Mukherjee, Wang, Tang, Awadallah, and
  Hu]{du2021fairness}
Mengnan Du, Subhabrata Mukherjee, Guanchu Wang, Ruixiang Tang, Ahmed Awadallah,
  and Xia Hu.
\newblock Fairness via representation neutralization.
\newblock \emph{Advances in Neural Information Processing Systems},
  34:\penalty0 12091--12103, 2021.

\bibitem[Dusenberry et~al.(2020)Dusenberry, Jerfel, Wen, Ma, Snoek, Heller,
  Lakshminarayanan, and Tran]{dusenberry2020efficient}
Michael Dusenberry, Ghassen Jerfel, Yeming Wen, Yian Ma, Jasper Snoek,
  Katherine Heller, Balaji Lakshminarayanan, and Dustin Tran.
\newblock Efficient and scalable bayesian neural nets with rank-1 factors.
\newblock In \emph{International conference on machine learning}, pp.\
  2782--2792. PMLR, 2020.

\bibitem[Dutta et~al.(2020)Dutta, Wei, Yueksel, Chen, Liu, and
  Varshney]{dutta2020there}
Sanghamitra Dutta, Dennis Wei, Hazar Yueksel, Pin-Yu Chen, Sijia Liu, and Kush
  Varshney.
\newblock Is there a trade-off between fairness and accuracy? a perspective
  using mismatched hypothesis testing.
\newblock In \emph{International Conference on Machine Learning}, pp.\
  2803--2813. PMLR, 2020.

\bibitem[Farquhar et~al.(2020)Farquhar, Gal, and
  Rainforth]{farquhar2020statistical}
Sebastian Farquhar, Yarin Gal, and Tom Rainforth.
\newblock On statistical bias in active learning: How and when to fix it.
\newblock In \emph{International Conference on Learning Representations}, 2020.

\bibitem[Feldman \& Zhang(2020)Feldman and Zhang]{feldman2020neural}
Vitaly Feldman and Chiyuan Zhang.
\newblock What neural networks memorize and why: Discovering the long tail via
  influence estimation.
\newblock \emph{Advances in Neural Information Processing Systems},
  33:\penalty0 2881--2891, 2020.

\bibitem[Gal \& Ghahramani(2016)Gal and Ghahramani]{gal2016dropout}
Yarin Gal and Zoubin Ghahramani.
\newblock Dropout as a bayesian approximation: Representing model uncertainty
  in deep learning.
\newblock In \emph{international conference on machine learning}, pp.\
  1050--1059. PMLR, 2016.

\bibitem[Gal et~al.(2017)Gal, Islam, and Ghahramani]{gal2017deep}
Yarin Gal, Riashat Islam, and Zoubin Ghahramani.
\newblock Deep bayesian active learning with image data.
\newblock In \emph{International Conference on Machine Learning}, pp.\
  1183--1192. PMLR, 2017.

\bibitem[Goel et~al.(2020)Goel, Gu, Li, and Re]{goel2020model}
Karan Goel, Albert Gu, Yixuan Li, and Christopher Re.
\newblock Model patching: Closing the subgroup performance gap with data
  augmentation.
\newblock In \emph{International Conference on Learning Representations}, 2020.

\bibitem[Gupta et~al.(2021)Gupta, Ferber, Dilkina, and
  Ver~Steeg]{gupta2021controllable}
Umang Gupta, Aaron~M Ferber, Bistra Dilkina, and Greg Ver~Steeg.
\newblock Controllable guarantees for fair outcomes via contrastive information
  estimation.
\newblock In \emph{Proceedings of the AAAI Conference on Artificial
  Intelligence}, volume~35, pp.\  7610--7619, 2021.

\bibitem[Gutmann \& Hyv{\"a}rinen(2010)Gutmann and
  Hyv{\"a}rinen]{gutmann2010noise}
Michael Gutmann and Aapo Hyv{\"a}rinen.
\newblock Noise-contrastive estimation: A new estimation principle for
  unnormalized statistical models.
\newblock In \emph{Proceedings of the thirteenth international conference on
  artificial intelligence and statistics}, pp.\  297--304. JMLR Workshop and
  Conference Proceedings, 2010.

\bibitem[Hamidieh et~al.(2022)Hamidieh, Zhang, and
  Ghassemi]{hamidieh2022evaluating}
Kimia Hamidieh, Haoran Zhang, and Marzyeh Ghassemi.
\newblock Evaluating and improving robustness of self-supervised
  representations to spurious correlations.
\newblock In \emph{ICML 2022: Workshop on Spurious Correlations, Invariance and
  Stability}, 2022.

\bibitem[Hasnain-Wynia et~al.(2007)Hasnain-Wynia, Baker, Nerenz, Feinglass,
  Beal, Landrum, Behal, and Weissman]{hasnain2007disparities}
Romana Hasnain-Wynia, David~W Baker, David Nerenz, Joe Feinglass, Anne~C Beal,
  Mary~Beth Landrum, Raj Behal, and Joel~S Weissman.
\newblock Disparities in health care are driven by where minority patients seek
  care: examination of the hospital quality alliance measures.
\newblock \emph{Archives of internal medicine}, 167\penalty0 (12):\penalty0
  1233--1239, 2007.

\bibitem[Havasi et~al.(2020)Havasi, Jenatton, Fort, Liu, Snoek,
  Lakshminarayanan, Dai, and Tran]{havasi2020training}
Marton Havasi, Rodolphe Jenatton, Stanislav Fort, Jeremiah~Zhe Liu, Jasper
  Snoek, Balaji Lakshminarayanan, Andrew~Mingbo Dai, and Dustin Tran.
\newblock Training independent subnetworks for robust prediction.
\newblock In \emph{International Conference on Learning Representations}, 2020.

\bibitem[He et~al.(2019)He, Zha, and Wang]{he2019unlearn}
He~He, Sheng Zha, and Haohan Wang.
\newblock Unlearn dataset bias in natural language inference by fitting the
  residual.
\newblock \emph{arXiv preprint arXiv:1908.10763}, 2019.

\bibitem[Hort et~al.(2022)Hort, Chen, Zhang, Sarro, and Harman]{hort2022bia}
Max Hort, Zhenpeng Chen, Jie~M Zhang, Federica Sarro, and Mark Harman.
\newblock Bia mitigation for machine learning classifiers: A comprehensive
  survey.
\newblock \emph{arXiv preprint arXiv:2207.07068}, 2022.

\bibitem[Houlsby et~al.(2011)Houlsby, Huszar, Ghahramani, and
  Lengyel]{houlsby2011bayesian}
Neil Houlsby, Ferenc Huszar, Zoubin Ghahramani, and Mate Lengyel.
\newblock Bayesian active learning for classification and preference learning.
\newblock \emph{arXiv preprint arXiv:1112.5745}, 2011.

\bibitem[Hyvarinen \& Morioka(2016)Hyvarinen and
  Morioka]{hyvarinen2016unsupervised}
Aapo Hyvarinen and Hiroshi Morioka.
\newblock Unsupervised feature extraction by time-contrastive learning and
  nonlinear ica.
\newblock \emph{Advances in neural information processing systems}, 29, 2016.

\bibitem[Idrissi et~al.(2022)Idrissi, Arjovsky, Pezeshki, and
  Lopez-Paz]{idrissi2022simple}
Badr~Youbi Idrissi, Martin Arjovsky, Mohammad Pezeshki, and David Lopez-Paz.
\newblock Simple data balancing achieves competitive worst-group-accuracy.
\newblock In \emph{Conference on Causal Learning and Reasoning}, pp.\
  336--351. PMLR, 2022.

\bibitem[Izmailov et~al.(2021)Izmailov, Vikram, Hoffman, and
  Wilson]{izmailov2021bayesian}
Pavel Izmailov, Sharad Vikram, Matthew~D Hoffman, and Andrew Gordon~Gordon
  Wilson.
\newblock What are bayesian neural network posteriors really like?
\newblock In \emph{International conference on machine learning}, pp.\
  4629--4640. PMLR, 2021.

\bibitem[Jung et~al.(2022)Jung, Chun, and Moon]{jung2022learning}
Sangwon Jung, Sanghyuk Chun, and Taesup Moon.
\newblock Learning fair classifiers with partially annotated group labels.
\newblock In \emph{Proceedings of the IEEE/CVF Conference on Computer Vision
  and Pattern Recognition}, pp.\  10348--10357, 2022.

\bibitem[Kang et~al.(2019)Kang, Xie, Rohrbach, Yan, Gordo, Feng, and
  Kalantidis]{kang2019decoupling}
Bingyi Kang, Saining Xie, Marcus Rohrbach, Zhicheng Yan, Albert Gordo, Jiashi
  Feng, and Yannis Kalantidis.
\newblock Decoupling representation and classifier for long-tailed recognition.
\newblock In \emph{International Conference on Learning Representations}, 2019.

\bibitem[Kim et~al.(2019)Kim, Kim, Kim, Kim, and Kim]{kim2019learning}
Byungju Kim, Hyunwoo Kim, Kyungsu Kim, Sungjin Kim, and Junmo Kim.
\newblock Learning not to learn: Training deep neural networks with biased
  data.
\newblock In \emph{Proceedings of the IEEE/CVF Conference on Computer Vision
  and Pattern Recognition}, pp.\  9012--9020, 2019.

\bibitem[Kim et~al.(2022)Kim, Hwang, Ahn, Park, and Kwak]{kim2022learning}
Nayeong Kim, Sehyun Hwang, Sungsoo Ahn, Jaesik Park, and Suha Kwak.
\newblock Learning debiased classifier with biased committee.
\newblock In \emph{ICML 2022: Workshop on Spurious Correlations, Invariance and
  Stability}, 2022.

\bibitem[Kirichenko et~al.(2022)Kirichenko, Izmailov, and
  Wilson]{kirichenko2022last}
Polina Kirichenko, Pavel Izmailov, and Andrew~Gordon Wilson.
\newblock Last layer re-training is sufficient for robustness to spurious
  correlations.
\newblock In \emph{ICML 2022: Workshop on Spurious Correlations, Invariance and
  Stability}, 2022.

\bibitem[Kirsch et~al.(2019)Kirsch, Van~Amersfoort, and
  Gal]{kirsch2019batchbald}
Andreas Kirsch, Joost Van~Amersfoort, and Yarin Gal.
\newblock Batchbald: Efficient and diverse batch acquisition for deep bayesian
  active learning.
\newblock \emph{Advances in neural information processing systems}, 32, 2019.

\bibitem[Koh et~al.(2021)Koh, Sagawa, Marklund, Xie, Zhang, Balsubramani, Hu,
  Yasunaga, Phillips, Gao, et~al.]{koh2021wilds}
Pang~Wei Koh, Shiori Sagawa, Henrik Marklund, Sang~Michael Xie, Marvin Zhang,
  Akshay Balsubramani, Weihua Hu, Michihiro Yasunaga, Richard~Lanas Phillips,
  Irena Gao, et~al.
\newblock Wilds: A benchmark of in-the-wild distribution shifts.
\newblock In \emph{International Conference on Machine Learning}, pp.\
  5637--5664. PMLR, 2021.

\bibitem[Kothawade et~al.(2022)Kothawade, Savarkar, Iyer, Ramakrishnan, and
  Iyer]{kothawade2022clinical}
Suraj Kothawade, Atharv Savarkar, Venkat Iyer, Ganesh Ramakrishnan, and Rishabh
  Iyer.
\newblock Clinical: Targeted active learning for imbalanced medical image
  classification.
\newblock In Ghada Zamzmi, Sameer Antani, Ulas Bagci, Marius~George Linguraru,
  Sivaramakrishnan Rajaraman, and Zhiyun Xue (eds.), \emph{Medical Image
  Learning with Limited and Noisy Data}, pp.\  119--129, Cham, 2022. Springer
  Nature Switzerland.
\newblock ISBN 978-3-031-16760-7.

\bibitem[Krueger et~al.(2021)Krueger, Caballero, Jacobsen, Zhang, Binas, Zhang,
  Le~Priol, and Courville]{krueger2021out}
David Krueger, Ethan Caballero, Joern-Henrik Jacobsen, Amy Zhang, Jonathan
  Binas, Dinghuai Zhang, Remi Le~Priol, and Aaron Courville.
\newblock Out-of-distribution generalization via risk extrapolation (rex).
\newblock In \emph{International Conference on Machine Learning}, pp.\
  5815--5826. PMLR, 2021.

\bibitem[Kumar et~al.(2013)Kumar, Lokshtanov, Vassilvitskii, and
  Vattani]{kumar2013near}
Ravi Kumar, Daniel Lokshtanov, Sergei Vassilvitskii, and Andrea Vattani.
\newblock Near-optimal bounds for cross-validation via loss stability.
\newblock In \emph{International Conference on Machine Learning}, pp.\  27--35.
  PMLR, 2013.

\bibitem[Lahoti et~al.(2020)Lahoti, Beutel, Chen, Lee, Prost, Thain, Wang, and
  Chi]{lahoti2020fairness}
Preethi Lahoti, Alex Beutel, Jilin Chen, Kang Lee, Flavien Prost, Nithum Thain,
  Xuezhi Wang, and Ed~Chi.
\newblock Fairness without demographics through adversarially reweighted
  learning.
\newblock \emph{Advances in neural information processing systems},
  33:\penalty0 728--740, 2020.

\bibitem[Lakshminarayanan et~al.(2017)Lakshminarayanan, Pritzel, and
  Blundell]{lakshminarayanan2017simple}
Balaji Lakshminarayanan, Alexander Pritzel, and Charles Blundell.
\newblock Simple and scalable predictive uncertainty estimation using deep
  ensembles.
\newblock \emph{Advances in neural information processing systems}, 30, 2017.

\bibitem[Le~Quy et~al.(2022)Le~Quy, Roy, Iosifidis, Zhang, and
  Ntoutsi]{le2022survey}
Tai Le~Quy, Arjun Roy, Vasileios Iosifidis, Wenbin Zhang, and Eirini Ntoutsi.
\newblock A survey on datasets for fairness-aware machine learning.
\newblock \emph{Wiley Interdisciplinary Reviews: Data Mining and Knowledge
  Discovery}, pp.\  e1452, 2022.

\bibitem[Lee et~al.(2021)Lee, Kim, Lee, Lee, and Choo]{lee2021learning}
Jungsoo Lee, Eungyeup Kim, Juyoung Lee, Jihyeon Lee, and Jaegul Choo.
\newblock Learning debiased representation via disentangled feature
  augmentation.
\newblock \emph{Advances in Neural Information Processing Systems},
  34:\penalty0 25123--25133, 2021.

\bibitem[Lee et~al.(2022)Lee, Park, Kim, Lee, Choi, and
  Choo]{lee2022biasensemble}
Jungsoo Lee, Jeonghoon Park, Daeyoung Kim, Juyoung Lee, Edward Choi, and Jaegul
  Choo.
\newblock Biasensemble: Revisiting the importance of amplifying bias for
  debiasing.
\newblock \emph{arXiv preprint arXiv:2205.14594}, 2022.

\bibitem[Levy et~al.(2020)Levy, Carmon, Duchi, and Sidford]{levy2020large}
Daniel Levy, Yair Carmon, John~C Duchi, and Aaron Sidford.
\newblock Large-scale methods for distributionally robust optimization.
\newblock \emph{Advances in Neural Information Processing Systems},
  33:\penalty0 8847--8860, 2020.

\bibitem[Li et~al.(2020)Li, Soltanolkotabi, and Oymak]{li2020gradient}
Mingchen Li, Mahdi Soltanolkotabi, and Samet Oymak.
\newblock Gradient descent with early stopping is provably robust to label
  noise for overparameterized neural networks.
\newblock In \emph{International conference on artificial intelligence and
  statistics}, pp.\  4313--4324. PMLR, 2020.

\bibitem[Li \& Vasconcelos(2019)Li and Vasconcelos]{li2019repair}
Yi~Li and Nuno Vasconcelos.
\newblock Repair: Removing representation bias by dataset resampling.
\newblock In \emph{Proceedings of the IEEE/CVF conference on computer vision
  and pattern recognition}, pp.\  9572--9581, 2019.

\bibitem[Li et~al.(2022)Li, De-Arteaga, and Saar-Tsechansky]{li2022more}
Yunyi Li, Maria De-Arteaga, and Maytal Saar-Tsechansky.
\newblock More data can lead us astray: Active data acquisition in the presence
  of label bias.
\newblock \emph{arXiv preprint arXiv:2207.07723}, 2022.

\bibitem[Lin et~al.(2017)Lin, Goyal, Girshick, He, and
  Doll{\'a}r]{lin2017focal}
Tsung-Yi Lin, Priya Goyal, Ross Girshick, Kaiming He, and Piotr Doll{\'a}r.
\newblock Focal loss for dense object detection.
\newblock In \emph{Proceedings of the IEEE international conference on computer
  vision}, pp.\  2980--2988, 2017.

\bibitem[Liu et~al.(2021)Liu, Haghgoo, Chen, Raghunathan, Koh, Sagawa, Liang,
  and Finn]{liu2021just}
Evan~Z Liu, Behzad Haghgoo, Annie~S Chen, Aditi Raghunathan, Pang~Wei Koh,
  Shiori Sagawa, Percy Liang, and Chelsea Finn.
\newblock Just train twice: Improving group robustness without training group
  information.
\newblock In \emph{International Conference on Machine Learning}, pp.\
  6781--6792. PMLR, 2021.

\bibitem[Liu et~al.(2022)Liu, Padhy, Ren, Lin, Wen, Jerfel, Nado, Snoek, Tran,
  and Lakshminarayanan]{liu2022simple}
Jeremiah~Zhe Liu, Shreyas Padhy, Jie Ren, Zi~Lin, Yeming Wen, Ghassen Jerfel,
  Zack Nado, Jasper Snoek, Dustin Tran, and Balaji Lakshminarayanan.
\newblock A simple approach to improve single-model deep uncertainty via
  distance-awareness.
\newblock \emph{arXiv preprint arXiv:2205.00403}, 2022.

\bibitem[Liu et~al.(2020)Liu, Niles-Weed, Razavian, and
  Fernandez-Granda]{liu2020early}
Sheng Liu, Jonathan Niles-Weed, Narges Razavian, and Carlos Fernandez-Granda.
\newblock Early-learning regularization prevents memorization of noisy labels.
\newblock \emph{Advances in neural information processing systems},
  33:\penalty0 20331--20342, 2020.

\bibitem[Locatello et~al.(2019{\natexlab{a}})Locatello, Abbati, Rainforth,
  Bauer, Sch{\"o}lkopf, and Bachem]{locatello2019fairness}
Francesco Locatello, Gabriele Abbati, Thomas Rainforth, Stefan Bauer, Bernhard
  Sch{\"o}lkopf, and Olivier Bachem.
\newblock On the fairness of disentangled representations.
\newblock \emph{Advances in Neural Information Processing Systems}, 32,
  2019{\natexlab{a}}.

\bibitem[Locatello et~al.(2019{\natexlab{b}})Locatello, Bauer, Lucic, Raetsch,
  Gelly, Sch{\"o}lkopf, and Bachem]{locatello2019challenging}
Francesco Locatello, Stefan Bauer, Mario Lucic, Gunnar Raetsch, Sylvain Gelly,
  Bernhard Sch{\"o}lkopf, and Olivier Bachem.
\newblock Challenging common assumptions in the unsupervised learning of
  disentangled representations.
\newblock In \emph{international conference on machine learning}, pp.\
  4114--4124. PMLR, 2019{\natexlab{b}}.

\bibitem[Lyons \& Peres(2017)Lyons and Peres]{lyons2017probability}
Russell Lyons and Yuval Peres.
\newblock \emph{Probability on trees and networks}, volume~42.
\newblock Cambridge University Press, 2017.

\bibitem[Maddox et~al.(2019)Maddox, Izmailov, Garipov, Vetrov, and
  Wilson]{maddox2019simple}
Wesley~J Maddox, Pavel Izmailov, Timur Garipov, Dmitry~P Vetrov, and
  Andrew~Gordon Wilson.
\newblock A simple baseline for bayesian uncertainty in deep learning.
\newblock \emph{Advances in Neural Information Processing Systems}, 32, 2019.

\bibitem[Madras et~al.(2018)Madras, Creager, Pitassi, and
  Zemel]{madras2018learning}
David Madras, Elliot Creager, Toniann Pitassi, and Richard Zemel.
\newblock Learning adversarially fair and transferable representations.
\newblock In \emph{International Conference on Machine Learning}, pp.\
  3384--3393. PMLR, 2018.

\bibitem[Mahmood et~al.(2021)Mahmood, Fidler, and Law]{mahmood2021low}
Rafid Mahmood, Sanja Fidler, and Marc~T Law.
\newblock Low-budget active learning via wasserstein distance: An integer
  programming approach.
\newblock In \emph{International Conference on Learning Representations}, 2021.

\bibitem[Martinez et~al.(2020)Martinez, Bertran, and
  Sapiro]{martinez2020minimax}
Natalia Martinez, Martin Bertran, and Guillermo Sapiro.
\newblock Minimax pareto fairness: A multi objective perspective.
\newblock In \emph{International Conference on Machine Learning}, pp.\
  6755--6764. PMLR, 2020.

\bibitem[Martinez et~al.(2021)Martinez, Bertran, Papadaki, Rodrigues, and
  Sapiro]{martinez2021blind}
Natalia~L Martinez, Martin~A Bertran, Afroditi Papadaki, Miguel Rodrigues, and
  Guillermo Sapiro.
\newblock Blind pareto fairness and subgroup robustness.
\newblock In \emph{International Conference on Machine Learning}, pp.\
  7492--7501. PMLR, 2021.

\bibitem[Matsushita et~al.(2018)Matsushita, Matsushita, and
  Hasebe]{matsushita2018deep}
Kayo Matsushita, Kayo Matsushita, and Hasebe.
\newblock \emph{Deep active learning}.
\newblock Springer, 2018.

\bibitem[Mehrabi et~al.(2021)Mehrabi, Morstatter, Saxena, Lerman, and
  Galstyan]{mehrabi2021survey}
Ninareh Mehrabi, Fred Morstatter, Nripsuta Saxena, Kristina Lerman, and Aram
  Galstyan.
\newblock A survey on bias and fairness in machine learning.
\newblock \emph{ACM Computing Surveys (CSUR)}, 54\penalty0 (6):\penalty0 1--35,
  2021.

\bibitem[Minderer et~al.(2021)Minderer, Djolonga, Romijnders, Hubis, Zhai,
  Houlsby, Tran, and Lucic]{minderer2021revisiting}
Matthias Minderer, Josip Djolonga, Rob Romijnders, Frances Hubis, Xiaohua Zhai,
  Neil Houlsby, Dustin Tran, and Mario Lucic.
\newblock Revisiting the calibration of modern neural networks.
\newblock \emph{Advances in Neural Information Processing Systems},
  34:\penalty0 15682--15694, 2021.

\bibitem[Ming et~al.(2022)Ming, Yin, and Li]{ming2022impact}
Yifei Ming, Hang Yin, and Yixuan Li.
\newblock On the impact of spurious correlation for out-of-distribution
  detection.
\newblock In \emph{Proceedings of the AAAI Conference on Artificial
  Intelligence}, volume~36, pp.\  10051--10059, 2022.

\bibitem[Nam et~al.(2020)Nam, Cha, Ahn, Lee, and Shin]{nam2020learning}
Junhyun Nam, Hyuntak Cha, Sungsoo Ahn, Jaeho Lee, and Jinwoo Shin.
\newblock Learning from failure: De-biasing classifier from biased classifier.
\newblock \emph{Advances in Neural Information Processing Systems},
  33:\penalty0 20673--20684, 2020.

\bibitem[Neal(2012)]{neal2012bayesian}
Radford~M Neal.
\newblock \emph{Bayesian learning for neural networks}, volume 118.
\newblock Springer Science \& Business Media, 2012.

\bibitem[Ovadia et~al.(2019)Ovadia, Fertig, Ren, Nado, Sculley, Nowozin,
  Dillon, Lakshminarayanan, and Snoek]{ovadia2019can}
Yaniv Ovadia, Emily Fertig, Jie Ren, Zachary Nado, David Sculley, Sebastian
  Nowozin, Joshua Dillon, Balaji Lakshminarayanan, and Jasper Snoek.
\newblock Can you trust your model's uncertainty? evaluating predictive
  uncertainty under dataset shift.
\newblock \emph{Advances in neural information processing systems}, 32, 2019.

\bibitem[Park et~al.(2022)Park, Lee, Lee, Hwang, Kim, and Byun]{park2022fair}
Sungho Park, Jewook Lee, Pilhyeon Lee, Sunhee Hwang, Dohyung Kim, and Hyeran
  Byun.
\newblock Fair contrastive learning for facial attribute classification.
\newblock In \emph{Proceedings of the IEEE/CVF Conference on Computer Vision
  and Pattern Recognition}, pp.\  10389--10398, 2022.

\bibitem[Petrovi{\'c} et~al.(2022)Petrovi{\'c}, Nikoli{\'c}, Radovanovi{\'c},
  Deliba{\v{s}}i{\'c}, and Jovanovi{\'c}]{petrovic2022fair}
Andrija Petrovi{\'c}, Mladen Nikoli{\'c}, Sandro Radovanovi{\'c}, Boris
  Deliba{\v{s}}i{\'c}, and Milo{\v{s}} Jovanovi{\'c}.
\newblock Fair: Fair adversarial instance re-weighting.
\newblock \emph{Neurocomputing}, 476:\penalty0 14--37, 2022.

\bibitem[Pfau(2013)]{pfau2013generalized}
David Pfau.
\newblock A generalized bias-variance decomposition for bregman divergences.
\newblock \emph{Unpublished Manuscript}, 2013.

\bibitem[Raffel et~al.(2020)Raffel, Shazeer, Roberts, Lee, Narang, Matena,
  Zhou, Li, Liu, et~al.]{raffel2020exploring}
Colin Raffel, Noam Shazeer, Adam Roberts, Katherine Lee, Sharan Narang, Michael
  Matena, Yanqi Zhou, Wei Li, Peter~J Liu, et~al.
\newblock Exploring the limits of transfer learning with a unified text-to-text
  transformer.
\newblock \emph{J. Mach. Learn. Res.}, 21\penalty0 (140):\penalty0 1--67, 2020.

\bibitem[Ragonesi et~al.(2021)Ragonesi, Volpi, Cavazza, and
  Murino]{ragonesi2021learning}
Ruggero Ragonesi, Riccardo Volpi, Jacopo Cavazza, and Vittorio Murino.
\newblock Learning unbiased representations via mutual information
  backpropagation.
\newblock In \emph{Proceedings of the IEEE/CVF Conference on Computer Vision
  and Pattern Recognition}, pp.\  2729--2738, 2021.

\bibitem[Rai et~al.(2010)Rai, Saha, Daum{\'e}~III, and
  Venkatasubramanian]{rai2010domain}
Piyush Rai, Avishek Saha, Hal Daum{\'e}~III, and Suresh Venkatasubramanian.
\newblock Domain adaptation meets active learning.
\newblock In \emph{Proceedings of the NAACL HLT 2010 Workshop on Active
  Learning for Natural Language Processing}, pp.\  27--32, 2010.

\bibitem[Rawls(2001)]{rawls2001justice}
John Rawls.
\newblock \emph{Justice as fairness: A restatement}.
\newblock Harvard University Press, 2001.

\bibitem[Rawls(2004)]{rawls2004theory}
John Rawls.
\newblock A theory of justice.
\newblock In \emph{Ethics}, pp.\  229--234. Routledge, 2004.

\bibitem[Ren et~al.(2021)Ren, Xiao, Chang, Huang, Li, Gupta, Chen, and
  Wang]{ren2021survey}
Pengzhen Ren, Yun Xiao, Xiaojun Chang, Po-Yao Huang, Zhihui Li, Brij~B Gupta,
  Xiaojiang Chen, and Xin Wang.
\newblock A survey of deep active learning.
\newblock \emph{ACM computing surveys (CSUR)}, 54\penalty0 (9):\penalty0 1--40,
  2021.

\bibitem[Rolf et~al.(2021)Rolf, Worledge, Recht, and
  Jordan]{rolf2021representation}
Esther Rolf, Theodora~T Worledge, Benjamin Recht, and Michael Jordan.
\newblock Representation matters: Assessing the importance of subgroup
  allocations in training data.
\newblock In \emph{International Conference on Machine Learning}, pp.\
  9040--9051. PMLR, 2021.

\bibitem[Sagawa et~al.(2019)Sagawa, Koh, Hashimoto, and
  Liang]{sagawa2019distributionally}
Shiori Sagawa, Pang~Wei Koh, Tatsunori~B Hashimoto, and Percy Liang.
\newblock Distributionally robust neural networks.
\newblock In \emph{International Conference on Learning Representations}, 2019.

\bibitem[Sagawa et~al.(2020)Sagawa, Raghunathan, Koh, and
  Liang]{sagawa2020investigation}
Shiori Sagawa, Aditi Raghunathan, Pang~Wei Koh, and Percy Liang.
\newblock An investigation of why overparameterization exacerbates spurious
  correlations.
\newblock In \emph{International Conference on Machine Learning}, pp.\
  8346--8356. PMLR, 2020.

\bibitem[Sanh et~al.(2020)Sanh, Wolf, Belinkov, and Rush]{sanh2020learning}
Victor Sanh, Thomas Wolf, Yonatan Belinkov, and Alexander~M Rush.
\newblock Learning from others' mistakes: Avoiding dataset biases without
  modeling them.
\newblock \emph{arXiv preprint arXiv:2012.01300}, 2020.

\bibitem[Settles(1994)]{settles1994active}
Burr Settles.
\newblock Active learning literature survey.
\newblock \emph{Machine Learning}, 15\penalty0 (2):\penalty0 201--221, 1994.

\bibitem[Sharaf et~al.(2022)Sharaf, Daume~III, and Ni]{sharaf2022promoting}
Amr Sharaf, Hal Daume~III, and Renkun Ni.
\newblock Promoting fairness in learned models by learning to active learn
  under parity constraints.
\newblock In \emph{2022 ACM Conference on Fairness, Accountability, and
  Transparency}, pp.\  2149--2156, 2022.

\bibitem[Shen et~al.(2021)Shen, Han, Cohn, Baldwin, and
  Frermann]{shen2021contrastive}
Aili Shen, Xudong Han, Trevor Cohn, Timothy Baldwin, and Lea Frermann.
\newblock Contrastive learning for fair representations.
\newblock \emph{arXiv preprint arXiv:2109.10645}, 2021.

\bibitem[Shui et~al.(2020)Shui, Zhou, Gagn{\'e}, and Wang]{shui2020deep}
Changjian Shui, Fan Zhou, Christian Gagn{\'e}, and Boyu Wang.
\newblock Deep active learning: Unified and principled method for query and
  training.
\newblock In \emph{International Conference on Artificial Intelligence and
  Statistics}, pp.\  1308--1318. PMLR, 2020.

\bibitem[Shui et~al.(2022)Shui, Chen, Li, Wang, and Gagn{\'e}]{shui2022fair}
Changjian Shui, Qi~Chen, Jiaqi Li, Boyu Wang, and Christian Gagn{\'e}.
\newblock Fair representation learning through implicit path alignment.
\newblock In Kamalika Chaudhuri, Stefanie Jegelka, Le~Song, Csaba Szepesvari,
  Gang Niu, and Sivan Sabato (eds.), \emph{Proceedings of the 39th
  International Conference on Machine Learning}, volume 162 of
  \emph{Proceedings of Machine Learning Research}, pp.\  20156--20175. PMLR,
  17--23 Jul 2022.

\bibitem[S{\l}owik \& Bottou(2022)S{\l}owik and
  Bottou]{slowik2022distributionally}
Agnieszka S{\l}owik and L{\'e}on Bottou.
\newblock On distributionally robust optimization and data rebalancing.
\newblock In \emph{International Conference on Artificial Intelligence and
  Statistics}, pp.\  1283--1297. PMLR, 2022.

\bibitem[Sohoni et~al.(2020)Sohoni, Dunnmon, Angus, Gu, and
  R{\'e}]{sohoni2020no}
Nimit Sohoni, Jared Dunnmon, Geoffrey Angus, Albert Gu, and Christopher R{\'e}.
\newblock No subclass left behind: Fine-grained robustness in coarse-grained
  classification problems.
\newblock \emph{Advances in Neural Information Processing Systems},
  33:\penalty0 19339--19352, 2020.

\bibitem[Sohoni et~al.(2021)Sohoni, Sanjabi, Ballas, Grover, Nie, Firooz, and
  R{\'e}]{sohoni2021barack}
Nimit Sohoni, Maziar Sanjabi, Nicolas Ballas, Aditya Grover, Shaoliang Nie,
  Hamed Firooz, and Christopher R{\'e}.
\newblock Barack: Partially supervised group robustness with guarantees.
\newblock \emph{arXiv preprint arXiv:2201.00072}, 2021.

\bibitem[Song et~al.(2020)Song, Kim, Park, and Lee]{song2020prestopping}
Hwanjun Song, Minseok Kim, Dongmin Park, and Jae-Gil Lee.
\newblock Prestopping: How does early stopping help generalization against
  label noise?
\newblock In \emph{ICML 2020 Workshop on Uncertainty and Robustness in Deep
  Learning}, 2020.

\bibitem[Sugiyama et~al.(2010)Sugiyama, Suzuki, and
  Kanamori]{sugiyama2010density}
Masashi Sugiyama, Taiji Suzuki, and Takafumi Kanamori.
\newblock Density ratio estimation: A comprehensive review (statistical
  experiment and its related topics).
\newblock \emph{RIMS Kokyuroku}, 1703:\penalty0 10--31, 2010.

\bibitem[Tae \& Whang(2021)Tae and Whang]{tae2021slice}
Ki~Hyun Tae and Steven~Euijong Whang.
\newblock Slice tuner: A selective data acquisition framework for accurate and
  fair machine learning models.
\newblock In \emph{Proceedings of the 2021 International Conference on
  Management of Data}, pp.\  1771--1783, 2021.

\bibitem[Tan et~al.(2020)Tan, Wang, Li, Li, Ouyang, Yin, and
  Yan]{tan2020equalization}
Jingru Tan, Changbao Wang, Buyu Li, Quanquan Li, Wanli Ouyang, Changqing Yin,
  and Junjie Yan.
\newblock Equalization loss for long-tailed object recognition.
\newblock In \emph{Proceedings of the IEEE/CVF conference on computer vision
  and pattern recognition}, pp.\  11662--11671, 2020.

\bibitem[Tartaglione et~al.(2021)Tartaglione, Barbano, and
  Grangetto]{tartaglione2021end}
Enzo Tartaglione, Carlo~Alberto Barbano, and Marco Grangetto.
\newblock End: Entangling and disentangling deep representations for bias
  correction.
\newblock In \emph{Proceedings of the IEEE/CVF conference on computer vision
  and pattern recognition}, pp.\  13508--13517, 2021.

\bibitem[Teney et~al.(2021)Teney, Abbasnejad, and van~den
  Hengel]{teney2021unshuffling}
Damien Teney, Ehsan Abbasnejad, and Anton van~den Hengel.
\newblock Unshuffling data for improved generalization in visual question
  answering.
\newblock In \emph{Proceedings of the IEEE/CVF International Conference on
  Computer Vision}, pp.\  1417--1427, 2021.

\bibitem[Tosh \& Hsu(2022)Tosh and Hsu]{tosh2022simple}
Christopher~J Tosh and Daniel Hsu.
\newblock Simple and near-optimal algorithms for hidden stratification and
  multi-group learning.
\newblock In \emph{International Conference on Machine Learning}, pp.\
  21633--21657. PMLR, 2022.

\bibitem[Tran et~al.(2022)Tran, Liu, Dusenberry, Phan, Collier, Ren, Han, Wang,
  Mariet, Hu, et~al.]{tran2022plex}
Dustin Tran, Jeremiah Liu, Michael~W Dusenberry, Du~Phan, Mark Collier, Jie
  Ren, Kehang Han, Zi~Wang, Zelda Mariet, Huiyi Hu, et~al.
\newblock Plex: Towards reliability using pretrained large model extensions.
\newblock \emph{arXiv preprint arXiv:2207.07411}, 2022.

\bibitem[Tsai et~al.(2021{\natexlab{a}})Tsai, Li, Ma, Zhao, Zhang, Morency, and
  Salakhutdinov]{tsai2021conditionala}
Yao-Hung~Hubert Tsai, Tianqin Li, Martin~Q Ma, Han Zhao, Kun Zhang,
  Louis-Philippe Morency, and Ruslan Salakhutdinov.
\newblock Conditional contrastive learning with kernel.
\newblock In \emph{International Conference on Learning Representations},
  2021{\natexlab{a}}.

\bibitem[Tsai et~al.(2021{\natexlab{b}})Tsai, Ma, Zhao, Zhang, Morency, and
  Salakhutdinov]{tsai2021conditionalb}
Yao-Hung~Hubert Tsai, Martin~Q Ma, Han Zhao, Kun Zhang, Louis-Philippe Morency,
  and Ruslan Salakhutdinov.
\newblock Conditional contrastive learning: Removing undesirable information in
  self-supervised representations.
\newblock \emph{arXiv preprint arXiv:2106.02866}, 2021{\natexlab{b}}.

\bibitem[Turc et~al.(2019)Turc, Chang, Lee, and Toutanova]{turc2019well}
Iulia Turc, Ming-Wei Chang, Kenton Lee, and Kristina Toutanova.
\newblock Well-read students learn better: On the importance of pre-training
  compact models.
\newblock \emph{arXiv preprint arXiv:1908.08962}, 2019.

\bibitem[Utama et~al.(2020{\natexlab{a}})Utama, Moosavi, and
  Gurevych]{utama2020mind}
Prasetya~Ajie Utama, Nafise~Sadat Moosavi, and Iryna Gurevych.
\newblock Mind the trade-off: Debiasing nlu models without degrading the
  in-distribution performance.
\newblock In \emph{Proceedings of the 58th Annual Meeting of the Association
  for Computational Linguistics}, pp.\  8717--8729, 2020{\natexlab{a}}.

\bibitem[Utama et~al.(2020{\natexlab{b}})Utama, Moosavi, and
  Gurevych]{utama2020towards}
Prasetya~Ajie Utama, Nafise~Sadat Moosavi, and Iryna Gurevych.
\newblock Towards debiasing nlu models from unknown biases.
\newblock In \emph{Proceedings of the 2020 Conference on Empirical Methods in
  Natural Language Processing (EMNLP)}, pp.\  7597--7610, 2020{\natexlab{b}}.

\bibitem[Van~Amersfoort et~al.(2020)Van~Amersfoort, Smith, Teh, and
  Gal]{van2020uncertainty}
Joost Van~Amersfoort, Lewis Smith, Yee~Whye Teh, and Yarin Gal.
\newblock Uncertainty estimation using a single deep deterministic neural
  network.
\newblock In \emph{International conference on machine learning}, pp.\
  9690--9700. PMLR, 2020.

\bibitem[van Amersfoort et~al.(2021)van Amersfoort, Smith, Jesson, Key, and
  Gal]{van2021feature}
Joost van Amersfoort, Lewis Smith, Andrew Jesson, Oscar Key, and Yarin Gal.
\newblock On feature collapse and deep kernel learning for single forward pass
  uncertainty.
\newblock \emph{arXiv preprint arXiv:2102.11409}, 2021.

\bibitem[Wang et~al.(2014)Wang, Huang, and Schneider]{wang2014active}
Xuezhi Wang, Tzu-Kuo Huang, and Jeff Schneider.
\newblock Active transfer learning under model shift.
\newblock In \emph{International Conference on Machine Learning}, pp.\
  1305--1313. PMLR, 2014.

\bibitem[Wenzel et~al.(2020{\natexlab{a}})Wenzel, Roth, Veeling, Swiatkowski,
  Tran, Mandt, Snoek, Salimans, Jenatton, and Nowozin]{wenzel2020good}
Florian Wenzel, Kevin Roth, Bastiaan~S Veeling, Jakub Swiatkowski, Linh Tran,
  Stephan Mandt, Jasper Snoek, Tim Salimans, Rodolphe Jenatton, and Sebastian
  Nowozin.
\newblock How good is the bayes posterior in deep neural networks really?
\newblock \emph{arXiv preprint arXiv:2002.02405}, 2020{\natexlab{a}}.

\bibitem[Wenzel et~al.(2020{\natexlab{b}})Wenzel, Snoek, Tran, and
  Jenatton]{wenzel2020hyperparameter}
Florian Wenzel, Jasper Snoek, Dustin Tran, and Rodolphe Jenatton.
\newblock Hyperparameter ensembles for robustness and uncertainty
  quantification.
\newblock \emph{Advances in Neural Information Processing Systems},
  33:\penalty0 6514--6527, 2020{\natexlab{b}}.

\bibitem[Williams \& Rasmussen(2006)Williams and
  Rasmussen]{williams2006gaussian}
Christopher~KI Williams and Carl~Edward Rasmussen.
\newblock \emph{Gaussian processes for machine learning}, volume~2.
\newblock MIT press Cambridge, MA, 2006.

\bibitem[Wilson \& Izmailov(2020)Wilson and Izmailov]{wilson2020bayesian}
Andrew~G Wilson and Pavel Izmailov.
\newblock Bayesian deep learning and a probabilistic perspective of
  generalization.
\newblock \emph{Advances in neural information processing systems},
  33:\penalty0 4697--4708, 2020.

\bibitem[Wilson et~al.(2016)Wilson, Hu, Salakhutdinov, and
  Xing]{wilson2016deep}
Andrew~Gordon Wilson, Zhiting Hu, Ruslan Salakhutdinov, and Eric~P Xing.
\newblock Deep kernel learning.
\newblock In \emph{Artificial intelligence and statistics}, pp.\  370--378.
  PMLR, 2016.

\bibitem[Xie et~al.(2022)Xie, Yuan, Li, Liu, Cheng, and Wang]{xie2022active}
Binhui Xie, Longhui Yuan, Shuang Li, Chi~Harold Liu, Xinjing Cheng, and Guoren
  Wang.
\newblock Active learning for domain adaptation: An energy-based approach.
\newblock In \emph{Proceedings of the AAAI Conference on Artificial
  Intelligence}, volume~36, pp.\  8708--8716, 2022.

\bibitem[Xie et~al.(2020)Xie, Kumar, Jones, Khani, Ma, and Liang]{xie2020n}
Sang~Michael Xie, Ananya Kumar, Robbie Jones, Fereshte Khani, Tengyu Ma, and
  Percy Liang.
\newblock In-n-out: Pre-training and self-training using auxiliary information
  for out-of-distribution robustness.
\newblock In \emph{International Conference on Learning Representations}, 2020.

\bibitem[Xu et~al.(2020)Xu, Ye, and Ruan]{xu2021understanding}
Da~Xu, Yuting Ye, and Chuanwei Ruan.
\newblock Understanding the role of importance weighting for deep learning.
\newblock In \emph{International Conference on Learning Representations}, 2020.

\bibitem[Yaghoobzadeh et~al.(2021)Yaghoobzadeh, Mehri, des Combes, Hazen, and
  Sordoni]{yaghoobzadeh2021increasing}
Yadollah Yaghoobzadeh, Soroush Mehri, Remi~Tachet des Combes, Timothy~J Hazen,
  and Alessandro Sordoni.
\newblock Increasing robustness to spurious correlations using forgettable
  examples.
\newblock In \emph{Proceedings of the 16th Conference of the European Chapter
  of the Association for Computational Linguistics: Main Volume}, pp.\
  3319--3332, 2021.

\bibitem[Yang \& Xu(2020)Yang and Xu]{yang2020rethinking}
Yuzhe Yang and Zhi Xu.
\newblock Rethinking the value of labels for improving class-imbalanced
  learning.
\newblock \emph{Advances in neural information processing systems},
  33:\penalty0 19290--19301, 2020.

\bibitem[Zhang et~al.(2020)Zhang, Menon, Veit, Bhojanapalli, Kumar, and
  Sra]{zhang2020coping}
Jingzhao Zhang, Aditya~Krishna Menon, Andreas Veit, Srinadh Bhojanapalli,
  Sanjiv Kumar, and Suvrit Sra.
\newblock Coping with label shift via distributionally robust optimisation.
\newblock In \emph{International Conference on Learning Representations}, 2020.

\bibitem[Zhang et~al.(2021)Zhang, Sohoni, Zhang, Finn, and
  R{\'e}]{zhang2021correct}
Michael Zhang, Nimit~Sharad Sohoni, Hongyang~R Zhang, Chelsea Finn, and
  Christopher R{\'e}.
\newblock Correct-n-contrast: A contrastive approach for improving robustness
  to spurious correlations.
\newblock In \emph{NeurIPS 2021 Workshop on Distribution Shifts: Connecting
  Methods and Applications}, 2021.

\bibitem[Zhang \& Sabuncu(2018)Zhang and Sabuncu]{zhang2018generalized}
Zhilu Zhang and Mert Sabuncu.
\newblock Generalized cross entropy loss for training deep neural networks with
  noisy labels.
\newblock \emph{Advances in neural information processing systems}, 31, 2018.

\bibitem[Zhao et~al.(2021)Zhao, Chen, Ju, and Xia]{zhao2021learning}
Bowen Zhao, Chen Chen, Qi~Ju, and Shutao Xia.
\newblock Learning debiased models with dynamic gradient alignment and
  bias-conflicting sample mining.
\newblock \emph{arXiv preprint arXiv:2111.13108}, 2021.

\bibitem[Zhao \& Gordon(2019)Zhao and Gordon]{zhao2019inherent}
Han Zhao and Geoff Gordon.
\newblock Inherent tradeoffs in learning fair representations.
\newblock \emph{Advances in neural information processing systems}, 32, 2019.

\bibitem[Zhou et~al.(2021)Zhou, Ma, Michel, and Neubig]{zhou2021examining}
Chunting Zhou, Xuezhe Ma, Paul Michel, and Graham Neubig.
\newblock Examining and combating spurious features under distribution shift.
\newblock In \emph{International Conference on Machine Learning}, pp.\
  12857--12867. PMLR, 2021.

\bibitem[Zhu et~al.(2021)Zhu, Zheng, Liao, Li, and Luo]{zhu2021learning}
Wei Zhu, Haitian Zheng, Haofu Liao, Weijian Li, and Jiebo Luo.
\newblock Learning bias-invariant representation by cross-sample mutual
  information minimization.
\newblock In \emph{Proceedings of the IEEE/CVF International Conference on
  Computer Vision}, pp.\  15002--15012, 2021.

\bibitem[Zhu et~al.(2014)Zhu, Anguelov, and Ramanan]{zhu2014capturing}
Xiangxin Zhu, Dragomir Anguelov, and Deva Ramanan.
\newblock Capturing long-tail distributions of object subcategories.
\newblock In \emph{Proceedings of the IEEE Conference on Computer Vision and
  Pattern Recognition}, pp.\  915--922, 2014.

\end{thebibliography}
\bibliographystyle{iclr2023_conference}

\clearpage
\appendix
\section{Additional Background}
\label{sec:background_app}

\subsection{Recap: Notation and Problem Setup.} 
\label{sec:notation_app}
\emph{\bf Dataset with subgroups:} We consider a dataset  $D$ where  each example $\{\rvx_i, y_i\}$ ($\rvx_i \in \X$ denotes the features and $y_i \in \Y$ the label) is associated with a discrete group label $g_i \in \gG=\{1, \dots, |\gG|\}$. \\
\emph{\bf Joint data distribution:} We denote $\gD=P(y, \rvx, g)$ as the joint distribution of the label, feature and groups, so that $D$ above can be understood as a size-$n$ set of i.i.d.\ samples from $\gD$. 
Notice that this formulation implies a flexible noise model $P(y|\rvx, g)$ that depends on $(\rvx, g)$. It also implies a flexible group-specific distribution $P(y, \rvx | g)$, where the joint distribution of $(y, \rvx)$ varies by group. Note however that we do assume that 
{\color{black} the group label does not have additional predictive power beyond appropriate representation of the features, i.e., there exists a representation function $h^*$ such that for $\rvz=h^*(\rvx)$, we have
$P(y|\rvz, g) = P(y|\rvz)$, i.e., the semantic, label-relevant features are invariant across subgroups \citep{arjovsky2019invariant, creager2021environment, shui2022fair}.}
{\color{black} Note however that we do assume that the group label does not have additional predictive power beyond appropriate representation of the features, i.e., there exists a representation function $h^*$ such that for $\rvz=h^*(\rvx)$, we have
$P(y|\rvz, g) = P(y|\rvz)$, i.e., the semantic, label-relevant features are invariant across subgroups \citep{arjovsky2019invariant, creager2021environment, shui2022fair}}
\\
\emph{\bf Subgroup prevelance:} We denote the prevalence of each group as $\gamma_g=E_{(y, \rvx, g) \sim \gD}(1_{G=g})$. As a result, the notion of \textit{dataset bias} is reflected as the imbalance in group distribution $P(G)=[\gamma_1, \dots, \gamma_{|\gG|}]$ \citep{rolf2021representation}. In the applications we consider, it is often feasible to identify a subset of \textit{underrepresented} groups $\gB \subset \gG$ which are not sufficiently represented in the population distribution $\gD$ and have $\gamma_g \ll \frac{1}{|\gG|}$ \citep{sagawa2019distributionally, sagawa2020investigation}. To this end, we also specify $\gD^*=P(y, \rvx |g)P^*(g)$ an  optimal distribution, where $P^*(g)$ is an ideal group distribution (i.e., uniform such that $P^*(g)=\gamma_g^*=\frac{1}{|\gG|}$) so that all groups have sufficient representation in the data.
\\
\emph{\bf Loss function:} We assume a loss function $L(y, \hat{y})$, that denotes the loss incurred when the predicted label is $\hat{y}$ while the actual label is $y$. \\
\emph{\bf Hypothesis space:} We consider learning a predictor from a hypothesis space $\gF$ of functions $f: \X \mapsto \Y$. We assume that the hypothesis space is well-specified, i.e., that it contains the Bayes-optimal predictor $\tilde{y}: \gX \rightarrow \gY$:
\[\tilde{y}\br{\rvx} = \argmin_{y' \in \Y} E_{y \sim P(y|\rvx)} ( L\br{y, y'} ).\]
We require the model class $\gF$ to come with certain degree of smoothness, so that the model $f \in \gF$ cannot arbitrarily overfit to the noisy labels during the course of training. In the case of over-parameterized models, this usually implies $\gF$ is subject to certain regularization that is appropriate for the model class (e.g., early stopping for SGD-trained neural networks) \citep{li2020gradient}.

\subsection{Disentangling model error under noise and bias}
\label{sec:error_decomposition}

Given a dataset $D \sim \gD$ and a loss function $L$ , we consider learning the prediction function 
$f_D = \argmin_{f \in \gF} L(f, y|D), \text{ where } L(y, f|D)=\sum_{\{\rvx_i, y_i\}\in D} L(y_i, f(\rvx_i))$. Following the previous work \citep{pfau2013generalized}, we denote the \textit{ensemble predictor} 
$\bar{f} = \argmin_{f\in \gF} E_{D\sim\gD}(L(f_D, f))$
over ensemble members $f_D$'s, where each $f_D$ is trained on a random draw of training dataset $D \sim \gD$, and
$\tilde{y}\br{\rvx} = \argmin_{y' \in \Y} E_{y \sim P(y|\rvx)} ( L\br{y, y'} )$ the (Bayes) optimal predictor.
For test example $\{y_i, \rvx_i\}$, we can decompose the predictive error of a trained model $f_D(\rvx)$ using a generalized bias-variance decomposition for Bregman divergence:
\begin{proposition}[Noise-Bias-Variance Decomposition under Bregman divergence \citep{domingos2000unified, pfau2013generalized}] 
Given a loss function of the Bregman divergence family, for a test example $\{y, \rvx\}$ the expected prediction loss $L(y, f_D(\rvx))$ of an empirical predictor $f_D$ can be decomposed as:
\begin{align}
E_D \big[ L(y, f_D(\rvx)) \big]
&= 
\underbrace{E_{D} \big[ L(y, \tilde{y}(\rvx)) \big]}_\text{{\color{OliveGreen} Noise}}
+ \quad
\underbrace{L(\tilde{y}(\rvx), \bar{f}(\rvx))}_\text{{\color{Maroon} Bias}}
\quad + \quad
\underbrace{E_{D} \big[ L(\bar{f}(\rvx), f_D(\rvx)) \big]}_\text{{\color{RoyalBlue} Uncertainty}}
\label{eq:error_decomposition}
\end{align}
\vspace{-1.5em}
\label{thm:error_decomposition}
\end{proposition}
Given a fixed data distribution $\gD$, the first term {\color{OliveGreen} $E_{D} \big[ L(y, f^*(\rvx)) \big]$} quantifies the {\color{OliveGreen} \textit{irreducible noise}} that is due to the stochasticity in the noisy observation $y$. 
The third term {\color{RoyalBlue} $E_{D} \big[ L(\bar{f}(\rvx), f_D(\rvx)) \big]$} quantifies the {\color{RoyalBlue} \textit{variance}} in the prediction, which can be due to variations in the finite-size data $D$, the stochasticity in the randomized learning algorithm $\gF \times D \rightarrow f_D$, or the randomness in the initialization of an  overparameterized model \citep{adlam2020understanding}.  Finally, the middle term {\color{Maroon} $L(\tilde{y}(\rvx), \bar{f}(\rvx))$} quantifies the {\color{Maroon}\textit{bias}} between  $\tilde{y}(\rvx)$ (i.e., the ``true label") and the ensemble predictor $\bar{f}$ learned from the empirical data $D \sim \gD$. It is inherent to the specification of the model class and cannot be eliminated by ensembling, e.g., it can be caused by model misspecification, missing features, or regularization. 
To make the idea concrete, consider a simple example where we fit a ridge regression model $f(\rvx_i)=\beta^\top \rvx_i$ to the Gaussian observation data $y_i=\theta^{\top} \rvx_i + \epsilon, \epsilon \sim N(0, \sigma^2)$ under an imbalanced experiment design, where we have $|\gG|$ treatment groups and $n_g$ observations in each group. Here, $\rvx_i = [1_{g_i = 1}, ..., 1_{g_i = |\gG|}]$ is a $|\gG| \times 1$ one-hot indicator of the membership of $g_i$ for each group in $\gG$, and $\theta=[\theta_1, \dots, \theta_{|\gG|}]$ is the true effect for each group. Then, under ridge regression, the noise-bias-variance decomposition for group $g$ is $E_D(L(y, f_D))={\color{OliveGreen} \sigma^2} + {\color{Maroon} \frac{(\lambda\theta_g)^{ 2}}{(n_g + \lambda)^2}} + {\color{RoyalBlue} \frac{\sigma^2 n_g}{(n_g + \lambda)^2}}$, where the regularization parameter $\lambda$  modulates a trade-off between the {\color{Maroon} \textit{bias}} and {\color{RoyalBlue} \textit{variance}} terms.

\subsection{Further Decomposition}
\label{sec:error_decomposition_further}

\paragraph{Further Uncertainty Decomposition for Probabilistic Models} As an aside, when the predictive model $f_D$ is probabilistic (e.g., the model generates a posterior predictive distribution $P(f|D)$ rather than a point estimate $f$), the {\color{RoyalBlue} \textit{variance}} in \Cref{eq:error_decomposition} is further decomposed as:
\begin{align}
E_D \big[ L(\bar{f}(\rvx), f_D(\rvx)) \big]= 
\underbrace{ E_D \big[ L(\bar{f}(\rvx), \mu_D(\rvx))  \big]}_\text{
{\color{RoyalBlue}\textit{Ensemble Diversity}}} + 
\underbrace{E_D E_{f \sim P(f|D)} \big[ L(\mu_D(\rvx), f(\rvx)) \big] }_\text{
{\color{RoyalBlue}\textit{Posterior Variance}}}
\end{align}
where $\mu_D(\rvx)=E_{f\sim P(f|D)}[f(\rvx)]$ is the posterior mean, and $v_D(\rvx) = E_{f \sim P(f|D)} \big[ L(\mu_D(\rvx), f(\rvx)) \big]$ is the posterior variance of each ensemble member. As shown, comparing to an ensemble of deterministic models, the ensemble of probabilistic models provides additional flexibility in quantifying model uncertainty via the extra term of expected posterior variance.

\paragraph{Further Bias Decomposition for Minority Groups} For the examples $\rvx$ coming from the underrepresented groups with $\gamma_g \ll \frac{1}{|\gG|}$, the bias term can be further decomposed into:
\begin{align}
L(\tilde{y}(\rvx), \bar{f}(\rvx))
&= 
\underbrace{L(\tilde{y}(\rvx), \bar{f}^*(\rvx))}_\text{{\color{Maroon} \textit{Bias, Model}}}
+
\underbrace{\gE(\bar{f}^*(\rvx), \bar{f}(\rvx))}_\text{{\color{Maroon} \textit{Excess Bias, Data}}},
\label{eq:bias_decomposition}
\end{align}
where $\bar{f}^\ast = \argmin_{f\in \gF} E_{D^\ast \sim \gD^\ast}(L(f_{D^\ast}, f))$ is the optimal ensemble predictor based on size-$n$ datasets $D^\ast$ sampled from the optimal distribution $\gD^\ast$ where all groups have equal representation. Here, $L(\tilde{y}(\rvx), \bar{f}^*(\rvx))$ is the bias inherent to the model class and cannot be eliminated by ensembling. It can be caused by model misspecification, missing features, or regularization. On the other hand, $\gE(\bar{f}^*(\rvx), \bar{f}(\rvx)) = L(\tilde{y}(\rvx), \bar{f}(\rvx)) - L(\tilde{y}(\rvx), \bar{f}^*(\rvx))$ indicates the ``excess bias" for the underrepresented groups caused by the imbalance in the group distribution $P(G)$ in the data-generation distribution $\gD$. 

To make the idea concrete, consider the ridge regression example from the previous section, where the noise-bias-variance decomposition for group $g$ is $E_D(L(y, f_D))={\color{OliveGreen} \sigma^2} + {\color{Maroon} \frac{(\lambda\theta_g)^{ 2}}{(n_g + \lambda)^2}} + {\color{RoyalBlue} \frac{\sigma^2 n_g}{(n_g + \lambda)^2}}$, with the regularization parameter $\lambda$  modulating a trade-off between the {\color{Maroon} \textit{bias}} and {\color{RoyalBlue} \textit{variance}} terms 
(\Cref{sec:error_decomposition_ridge}). Consequently, for an underrepresented group with small size $\gamma_g \ll \frac{1}{|\gG|}$, its predictive bias ${\color{Maroon} \frac{(\lambda\theta_g)^{\ast 2}}{(n_g + \lambda)^2}}$ is exacerbated due to lacking sufficient statistical information to counter the regularization bias, incuring an excessive bias of $\gE(\bar{f}^*(\rvx), \bar{f}(\rvx))\approx \frac{\lambda \theta_g}{n \gamma^\ast_g \gamma_g}(\gamma^\ast_g - \gamma_g)$ when compared to an optimal ensemble predictor $\bar{f}^*$ trained from a perfectly balanced size-$n$ datasets with $\gamma^\ast_g=1/|\gG|$.


\subsection{Modern uncertainty estimation techniques in deep learning}
\label{sec:deep_uncertainty}

For a deep classifier $p(\rvx) = \sigma(f(\rvx))$ with logit function $f(\rvx)=\beta^\top h(\rvx)$ and $h(\rvx) \in \sR^M$ the last-layer hidden embeddings, the modern deep uncertainty methods quantifies model uncertainty by enabling it to generate random samples from a predictive distribution. That is, for a model trained on data $D=\{(\rvx_i, y_i)\}_{i=1}^n$, given a test data point $\rvx_{test}$, the model can return a size-$K$ sample:
\begin{align*}
    \{f_k(\rvx_{test})\}_{k=1}^K \sim P(f|\rvx_{test}, D).
\end{align*}
For example, in Monte Carlo Dropout \citep{gal2016dropout}, the samples is generated by perturbing the dropout mask in the learned predictive function $f(\cdot) =\beta^\top h(\cdot)$'s embedding function $h(\cdot)$, while in Deep Ensemble \citep{lakshminarayanan2017simple}, the sample comes directly from the multiple parallel-trained ensemble members. Finally, in a neural Gaussian process model \citep{wilson2016deep, liu2022simple, van2021feature}, the samples are generated from a Gaussian process model using the hidden embedding function $h(\rvx)$ as the input. For example, for classification problems, the predictive variance of the Gaussian process model $v(\rvx_{test}) = Var(f|\rvx_{test}, D)$ can be expressed as (\citet{williams2006gaussian}, Chapter 3): 
\begin{align*}
    v(\rvx_{test}) = \rvk(\rvx_{test})_{1 \times n}^\top \rmV_{n \times n} \rvk(\rvx_{test})_{n \times 1};
\end{align*}
where $\rmV_{n \times n}$ is a fixed matrix computed from training data, and $\rvk(\rvx_{test})=[k(\rvx_{test}, \rvx_1), \dots,  k(\rvx_{test}, \rvx_n)]$ is a vector of kernel distances between $\rvx_{test}$ and the training examples $\{\rvx_i\}_{i=1}^n$. The kernel function $k$ is commonly defined to be a monotonic function of the hidden embedding distance, e.g., $k(\rvx_{test}, \rvx_i)=exp(-||h(\rvx_{test}) - h(\rvx_i)||_2^2)$ for the RBF kernel. As a result, the predictive uncertainty for a data points $\rvx_i$ is determined by the distance between $\rvx_{test}$ from the training data $\{\rvx_i\}_{i=1}^n$. Consequently, a \gls{DNN} model's quality in representation learning has non-trivial impact on its uncertainty performance. Although first mentioned in the context of neural Gaussian process, this connection between the quality of representation learning and the quality of uncertainty quantification also holds for state-of-the-art techniques such as Deep Ensemble, as model averaging cannot eliminate the systematic errors in representation learning and consequently the issue in uncertainty quantification (for example, see \Cref{fig:2d_uncertainty_by_model_base} and the corresponding ensemble uncertainty surface \Cref{fig:2d_uncertainty_base}).

\paragraph{Neural Gaussian Process Ensemble}
In this work, to comprehensively investigate the effect of different uncertainty techniques, we should to use a Deep Ensemble of neural Gaussian process as our canonical model. That is, we parallel train $K$ neural Gaussian process models $\{f_k\}_{k=1}^K$. Then, given a test data point $\rvx_{test}$, each ensemble member will return a predictive distribution with means $\{\mu_k(\rvx)\}_{k=1}^K$ and variances $\{v_k(\rvx)\}_{k=1}^K$. Then, we can generate model prediction as $\E_k[\mu_k(\rvx)]$, and quantify uncertainty in one of the two ways:
\begin{align*}
    \mbox{Ensemble Diversity}: \quad & Var_k(\mu_k(\rvx)); \\
    \mbox{Posterior Variance}: \quad &  \E_k(v_k(\rvx)),
\end{align*}
where $Var_k$, $\E_k$ are empirical means and variances over the ensemble members. As shown, they correspond to the two components of the total model variance under squared error introduced in \ref{sec:error_decomposition_further}. We investigate the effectiveness of these two uncertainty signals in the experiments.

{\color{black}
\subsection{Connection between group robustness and fairness}
\label{sec:fairness_connection_app}

The notion of group robustness (e.g., worst-group accuracy) we considered in this work corresponds to the notion of \textit{minimax fairness} or \textit{Rawlsian max-min fairness} 
\citep{lahoti2020fairness, martinez2020minimax, martinez2021blind, diana2021minimax}. The philosophical foundation of this notion has been well-established \citep{rawls2001justice, rawls2004theory}. There have been notable works developed for both analyzing accuracy-fairness tradeoff under this notion (e.g., \citet{martinez2020minimax, martinez2021blind}), and for improving fairness performance without explicit annotation (i.e., ARL, \citep{lahoti2020fairness}). It is also identified as the original objective of the well-known \gls{IRM} work for invariant learning (\citet{arjovsky2019invariant}, Section 2).

To this end, under a sampling model with well-calibrated uncertainty, the active learning procedure as discussed in introduction is expected to carry benefit for model fairness as well. Specifically, in the context of fairness-aware learning, this means a sampling model with calibrated uncertainty preferentially samples the under-represented protected group, leading to a sampled training set with more balanced representation among population subgroups, and consequently reducing the between-group generalization gap in the trained model and promoting its fairness properties.
}

\newpage
\section{Method Summary}

\subsection{Algorithm}
\label{sec:alg}

\begin{algorithm}[ht]
\captionsetup{font=small}
\caption{\glsfirst{ISP}}
\label{alg:introspective_self_play}
\small
\begin{minipage}{1\linewidth}
\begin{algorithmic}
\Inputs{Training data $D_{train}=\{y_i, \rvx_i\}_{i=1}^n$; (Optional) Group annotation $G_{train}=\{g_i\}_{i=1}^n$;\\
$\qquad\quad\;$ Unlabelled data $D_{pool}=\{\rvx_j\}_{j=1}^{n'}$.}
\Output{Predicted probability $\{p(y|\rvx_j)\}_{j=1}^{n'}$; Bias probability $\{p(b|\rvx_j)\}_{j=1}^{n'}$; Predictive variance $\{v(\rvx_j)\}_{j=1}^{n'}$.}\\

\LeftComment{\underline{Stage I: Label Generation}}
\If{$G_{train} \neq \emptyset$}
    \State{$B_{train} = \{b_i = I(g_i \in \gB)\}$;}
    \Comment{Make underrepresentation label using group annotation $g_i$.}
\Else 
    \State{$\hat{B}_{train} = SelfPlayBiasEstimation(D_{train})$.}
    \Comment{Estimate underrepresentation label using \Cref{alg:bias_estimator}}
\EndIf
\State $ $
\LeftComment{\underline{Stage II: Introspective Training}}

\State{Train $\hat{f}$ on $D_{train}$ with multi-task introspective objective $L((y_i, b_i), \rvx_i)$.}
\Comment{\Cref{eq:introspective_objective}}

\State{Evaluate $\hat{f}$ on $\rvx_j \in D_{pool}$ to generate sampling signals $\{p(y|\rvx_j), p(b|\rvx_j), v(\rvx_j)\}_{j=1}^{n'}$.} \Comment{\Cref{eq:introspective_predictions}}

\end{algorithmic}
\end{minipage}
\end{algorithm}

\begin{algorithm}[ht]
\captionsetup{font=small}
\caption{Underrepresentation Label Estimation via \textit{Cross-validated Self-play}}
\label{alg:bias_estimator}
\small
\begin{minipage}{1\linewidth}
\begin{algorithmic}
\Inputs{Training data $D_{train}=\{y_i, \rvx_i\}_{i=1}^n$.}
\Output{Estimate underrepresentation labels $\hat{B}_{train}$.}\\

\State Train $K$-fold cross-validated ensemble $\{f_k\}_{k=1}^K$ with $D_{train}$.
\State Compute in-sample and out-of-sample ensemble predictions $\{f_{in,k}(\rvx_i)\}_{k=1}^{K_{in}}, \{f_{out,k}(\rvx_i)\}_{k=1}^{K_{out}}$ for all $\rvx_i \in D_{train}$.
\State Estimate underrepresentation labels as $\hat{B}_{train} = \{b_i=\E_k[L(\bar{f}_{in}(\rvx_i), {f}_{out, k}(\rvx_i))]\}_{i=1}^n$. \Comment{(\Cref{eq:bias_estimator})}

\end{algorithmic}
\end{minipage}
\end{algorithm}

\subsection{Estimating Generalization Gap using Cross-validated Ensemble}
\label{sec:cv_ensemble}

\paragraph{Practical Comments.} Note that due to its cross validation nature, the \textit{self-play} bias estimator $\hat{b}_i$ estimates the generalization error of a weaker model (i.e., trained on a smaller data size $n_{cv} < n$). This is in fact consistent with the practice in the previous debasing literature, where the main model is trained on the error signals from weaker and more biased models \citep{clark2019don, he2019unlearn, nam2020learning}. 

Further, in the context of SGD-trained neural networks, it is important to properly estimate the $\bar{f}_{in}(\rvx_i)$ so it does not overfit to the training label, via early stopping \citep{li2020gradient, liu2020early}. This is easy to do in the context of cross validation: during training, we collect the estimated bias $\hat{b}_{i,t}$ across the training epochs $t=1, \dots, T$, and select the optimal stopping point $t$ as the first time the out-of-sample error $\mathbb{E}\big[ L(y_i, f_{out,k}(\rvx_i))\big]$ stablizes. In practice, we specify the early-stopping criteria as when the running average (within a window $T'=5$) of the cross validation error first stablizes below a threshold $\epsilon$. This is to prevent the situation where the errors for some hard-to-learn examples keep oscillating throughout training and never stabilize.

{\color{black}
\paragraph{Is perfect group identification necessary?}

In fact, it is not necessary to perfectly identify all the subgroup examples to improve the final model's tradeoff frontier. 
For example, when there exists hard-to-learn majority-group examples and also easy-to-learn minority-group examples (also known as "benign bias" examples in the debiasing literature \citep{nam2020learning}), the active sampling budget is better spent in sampling some of the hard-to-learn majority examples over the easy-to-learn minority examples, so that the final tradeoff frontier is meaningfully improved both in the directions of subgroup performance and of overall accuracy. To this end, including some hard-to-learn majority examples (detected by the loss-based procedure) into introspective training help us to achieve that.
This is exactly the case in the toxicity detection experiments (\Cref{sec:exp}). As shown in \Cref{tab:result_main}, comparing between ISP-Identity v.s. ISP-Gap (trained on true group label v.s. cross-validation estimated label), the ISP-Gap attains a significantly stronger accuracy-fairness performance while sampling less minority examples. This is likely caused by the cross-validation estimator's ability in capturing challenging, non-identity-related examples instead of the easy-to-learn and identity-related examples, which resulted in less sampling redundancy in the minority group (i.e., the identity-related comments), and led to improved tradeoff frontier for the final model.

}

\subsection{Hyperparameters and Computational Complexity}
\label{sec:hyper_app}

\paragraph{Hyper-parameters} The full \gls{ISP} procedure contains 3 hyper-parameters: The (optional) \textit{cross-validated self-play} in Stage I contains all three hyper-parameters: (1) the number of ensemble models $K$ and (2) the number of examples $n_{cv}$ to train each model. Both are standard to the bootstrap ensemble procedure, and we set them to $K=10$ and $n_{cv}=n / K$ in this work to ensure the total computation complexity is comparable to training a single model on the full dataset. (3) the early stopping criteria $\epsilon$ for noise estimation (as discussed in the previous section \ref{sec:cv_ensemble}), we set it heuristically to $\epsilon=0.1$ in this work after visual inspection of the validation learning curves. The \textit{introspective training} in Stage II does not contain additional hyperparameter other than the standard supervised learning parameters (e.g., learning rate and training epochs). We set these parameters based on a standard supervised learning hyperparameter sweep based on the full data.

\paragraph{Computation Complexity} When the group annotation is available, the computation complexity of the \gls{ISP} procedure (i.e., Stage II only) should be equivalent to the standard \gls{ERM} procedure. On the other hand, the computational complexity of the full \gls{ISP} procedure (Stage I + II) should be comparable to that of a standard two-stage debiasing method that trains multiple single models on the full dataset \citep{utama2020towards, liu2021just, yaghoobzadeh2021increasing, nam2020learning, creager2021environment, kim2022learning}.

\begin{figure}[ht]
    \centering
    \includegraphics[width=0.25\linewidth]{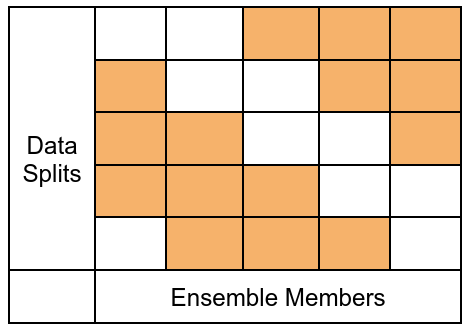}
    \caption{
    An example of 5-fold cross-validated ensemble. Each ensemble member received $60\%$ of the data (highlighted in orange), and each data split receives 3 in-sample $f_{in,k}$ and 2 out-of-sample predictions $f_{out,k}$.
    }
    \label{fig:self_play_ensemble}
\end{figure}

{\color{black}
More specifically, 
The complexity of the cross-validated ensemble approach (i.e., the cross-validated self-play) is identical to that of a single model training on full data (in the big O sense). This is because for cross-validated ensembles, each ensemble member model is trained on different subsets of the data. Therefore, if the complexity for a single model is $T$, then the complexity of the cross-validated ensemble is $K*p*T$, where $p < 1$ is the percentage of sub-sampled training examples for each ensemble member, and K is the ensemble size, notice that $K*p*T=O(T)$ since $p$ and $K$ are both constants that don't scale with data size. 

In practice, we find setting $p=1/K$ (i.e., train K model on K non-overlap data splits) works rather well and has the benefit of improving the cross-validation error estimate by reducing cross-split correlation \citep{blum1999beating}. This leads the complexity of cross-validated ensembles to be equal to $T$, the complexity of training a single model on the full data. Furthermore, since ensemble training is embarrassingly parallel, the complexity can also be massively reduced by leveraging multi-core machines to perform parallel training.}

\section{Additional Theory Discussion}
\label{sec:theory_discussion}

\subsection{\Cref{thm:bias_awareness}: when will it break down?} 

It is important to consider the situations where the guarantee in \Cref{thm:bias_awareness} can break down. First note that since \Cref{thm:bias_awareness} is a consistency result, it expects reasonable number of training examples in the neighborhood of $\rvx$ to ensure the convergence of $p(\rvx|b)$ (\Cref{eq:bias_aware_representation}). In practice, the convergence is not guaranteed for the tail groups that have extraordinarily low example count in the dataset, since they can be statistically indistinguishable from the label noise under a finite-size dataset. To this end, \Cref{fig:2d_example} provides us an empirical view of the model's bias estimation performance for the low-prevalence groups. As shown, when the tail-group examples are well-separated from the majority groups (i.e., the bias learning task is easy), a very small amount of examples is sufficient to lead to reasonable underrepresentation prediction performance (\Cref{fig:2d_bias_isp}, only 2 examples per tail group). This suggests that the actual number of examples required to attain reasonable performance depends on the difficulty of the bias learning task. For example, for a challenging tail group that shares much surface-form similarity with the majority groups (e.g., in \Cref{thm:bias_awareness}, the tail examples are extremely close or even nested within the majority groups), more examples are expected to obtain a reasonable performance in underrepresentation prediction. Also note that the learning quality of the model representation $h(\cdot)$ depends on the quality of the underrepresentation label $b$. In particular, the noise in the underrepresentation labels can negatively impact introspective training by diminishing the log likelihood ratio $|log p(\rvx|b=1) - log p(\rvx|b=0)|$ in (\ref{eq:bias_aware_representation}), leading the $\beta_b^\top h(\rvx)$ to converge to a target of lower magnitude, hampering its ability in \textit{bias-awareness}. This issue becomes relevant when the underrepresentation labels need to be estimated from data, where the vulnerability of fairness procedures to label noise has been well-noted in the literature (e.g., \citet{lahoti2020fairness}), highlighting the importance of label noise control when estimating the membership of the underrepresented groups. For example, we note that some of the now popular method (e.g., \textit{\gls{JTT}} \citep{liu2021just}) estimates the membership of the underpresented groups using generalization error, which may be susceptible to label noise (c.f., the noise-bias-variance decomposition in \Cref{sec:error_decomposition}). We develop bias estimation procedures that controls for label noise in \Cref{sec:bias_estimation}.

\subsection{On Underrepresented group $\gB$ in \Cref{thm:optimal_allocation_frontier_main}}
\label{sec:optimal_allocation_frontier_discussion}
In the main paper (section \ref{sec:theory}), we characterized the optimal allocation under assumptions on decay rates of group-specific risk. We give some further analysis and interpretation of this result here. To simplify things, we assume that there are only $\gG={1, 2}$ groups so that the optimal allocation is characterized by a single number $\alpha \in [0, 1]$ representing the allocation to group 1 (the other group allocation would be $1-\alpha$). We further assume that $\gamma_1=\gamma_2=\frac{1}{2}$ and that $c_2=tc_1$ for some $t > 0$, From the proof of theorem \Cref{thm:optimal_allocation_frontier_main} from section \ref{sec:optimal_allocation_frontier_proof}, we know that
\[\gamma_1\theta_1+\gamma_2\theta_2=1 \implies \theta_2=2-\theta_1\]
so that we can set $\theta_1=\theta, \theta_2=2-\theta$. 
Then, we have that 
\[\alpha^* = \frac{\br{\theta}^{\frac{1}{p+1}}}{\br{\theta}^{\frac{1}{p+1}}+\br{t(2-\theta)}^{\frac{1}{p+1}}}\]

Further, from the proof of the theorem, we can infer that the optimal $\theta$ is 
\[
\begin{cases}
\omega & \text{ if } \omega < \frac{2}{1+t^{\frac{1}{p}}} \\
2-\omega &\text{ if } \omega > \frac{2}{1+t^{\frac{1}{p}}} \\
\frac{2}{1+t^{\frac{1}{p}}} & \text{ if } \omega = \frac{2}{1+t^{\frac{1}{p}}}
\end{cases}
\]
so that the optimal allocation is
\[\begin{cases}
\frac{1}{1+ \br{t\br{\frac{2-\omega}{\omega}}}^{\frac{1}{p+1}}} & \text{ if } \omega < \frac{2}{1+t^{\frac{1}{p}}} \quad (\mathcal{B}=\{2\}) \\
\frac{1}{1+ \br{t\br{\frac{\omega}{2-\omega}}}^{\frac{1}{p+1}}}& \text{ if } \omega > \frac{2}{1+t^{\frac{1}{p}}} \quad (\mathcal{B}=\{1\})\\
\frac{1}{1+t} &  \text{ if } \omega = \frac{2}{1+t^{\frac{1}{p}}} \quad (\mathcal{B}=\{1, 2\})
\end{cases}
\]
This shows that the optimal allocation to group 1 decays as a function of $t$, the difficulty of learning group 2 relative to group 1. Further, it shows that this decay is more stark for smaller values of $\omega$ that weight the fairness risk stronger than the population level risk. 

Further, group $1$ belongs to the underrepresented group $\mathcal{B}$ if $\omega \geq \frac{2}{1+t^{\frac{1}{p}}}$, showing that the chances that group $1$ belongs to the underrepresented group increase as $\omega$ increases (there is greater emphasis on the fairness risk) or as $t$ decreases (group 1 becomes harder to learn relative to group 2).

\section{Related Work}
\label{sec:related_work}

\subsection{Supervised and semi-supervised learning under dataset bias}
\label{sec:related_debiasing_supervised}

In recent years, there has been significant interest in studying robust generalization for long-tail population subgroups under dataset bias. The literature is vast and encompasses topics including fairness, debiasing, long-tail recognition, spurious correlation, distributional (i.e., domain or subpopulation) shift, etc. In the following, we focus on notable and recent work that is highly relevant to the \gls{ISP} approach, and refer to works such as \citet{caton2020fairness,mehrabi2021survey,hort2022bia} for an exhaustive survey.

Majority of the fairness and debiasing work focuses on the supervised learning setting, where the model only have access to a fixed and imbalanced dataset. Among them, the earlier work operated under the assumption that the source of dataset bias is completely known, and the group annotation is available for every training example. Then these group information is use to train a robust model by modify components of the training pipeline (e.g., training objective, regularization method, or composition of training data). For example, \citet{levy2020large, sagawa2019distributionally, zhang2020coping} proposes minimizing the worst-group loss via \glsfirst{DRO}, which is shown to be equivalent to ERM training on a well-curated training set in some settings \citep{slowik2022distributionally}; 
{\color{black} \gls{IRM} \citep{arjovsky2019invariant} proposes 
learning a classifier that is simultaneously optimal for all groups by learning a invariant representation.}
\citet{teney2021unshuffling, idrissi2022simple, byrd2019effect, xu2021understanding} studies the effect of group-weighted loss in model's fairness-accuracy performance, and REx \citep{krueger2021out} minimizes a combination of group-balanced and worst-case loss. Further, the recent literature has also seen sophisticated neural-network loss that modifies gradient for the tail-group examples. For example, LDAM \citep{cao2019learning} proposes to modify group-specific logits by an offset factor that is associated with group size, and equalization loss \citep{tan2020equalization} uses a instance-specific mask to suppress the ``discouraging gradients" from majority groups to the rare groups. 
On the regularization front, the examples include 
{\color{black} IRM$\_$v1  \citep{arjovsky2019invariant} that proposed to approximate the original \gls{IRM} objective through gradient penalty.
} \textit{Heteroskedastic Adaptive Regularization} (HAR) \citep{cao2020heteroskedastic} imposes Lipschitz regularization in the neighborhood of tail-group examples. There also exists a large collection of work imposing other types of fairness constraints. 
Finally, the third class of methods modifies the composition of the training data by enriching the number of obsevations in the tail groups, this includes \citet{sagawa2020investigation, idrissi2022simple} that study the impact of resampling to the worst-group performance, and \citet{goel2020model} that generates synthetic examples for the minority groups.
In the setting where the group information is available, our work proposes a novel approach (introspective training) that has both a theoretical guarantee and is empirically competitive than reweighted training.

On the other hand, there exist a separate stream of work that allows for partial group annotation, i.e., the types of bias underlying a dataset is still completely known, but the group annotation is only available for a subset of the data. Most work along this direction employs semi-supervised learning techniques (e.g, confidence-threshold-based pseudo labeling), with examples include \gls{SSA} \citep{nam2020learning}, BARACK \citep{sohoni2021barack} and Fair-PG \citep{jung2022learning}. This setting can be considered as a special case of \gls{ISP} where we use group information as the underrepresentation label to train the $p(b|\rvx)$ predictor. However, our goal is distinct that we study the efficacy of this signal as an active learning policy, and also investigate its extension in the case where the label information is completely unobserved in the experiments \Cref{sec:exp_ablation}.


\subsection{Estimating dataset bias for model debiasing} 
\label{sec:related_bias_estimation}
In the situation where the source of dataset bias is not known and the group annotation is unavailable, several techniques has been proposed to estimate proxy bias labels for the downstream debiasing procedures. These methods roughly fall into three camps: clustering, adversarial search, and using the generalization error from a biased model. 

For clustering, GEORGE \citep{sohoni2020no} and CNC \citep{zhang2021correct} proposed estimating group memberships of examples based on clustering the last hidden-layer output. For adversarial search, REPAIR \citep{li2019repair}, ARL \citep{lahoti2020fairness}, EIIL \citep{creager2021environment}, BPF \citep{martinez2021blind}, FAIR \citep{petrovic2022fair}, Prepend \citep{tosh2022simple} 
infers the group assignments by finding the worst-case group assignments that maximize certain objective function. {\color{black} For example, \textit{Environment Inference for Invariant Learning} (EIIL) \citep{creager2021environment} infer the group membership by maximizing the gradient-based regularizer from IRM$\_$v1.}

Estimating bias label using the error from a biased model is by far the most popular technique. These include \textit{forgettable examples} \citep{yaghoobzadeh2021increasing}, \textit{Product of Experts} (PoE) \citep{clark2019don, sanh2020learning}, DRiFt \citep{he2019unlearn} and \textit{Confidence Regularization} (CR) \citep{utama2020towards, utama2020mind} that uses errors from a separate class of weak models that is different from the main model; \textit{Neutralization for Fairness} (RNF) \citep{du2021fairness} and \textit{Learning from Failure} (LfF) that trains a bias-amplified model of the same architecture using generalized cross entropy (GCE); and\textit{ \glsfirst{JTT}} that directly uses the error from a standard model trained from cross entropy loss. 

Notably, there also exists several work that estimates bias label using ensemble techniques, this includes \textit{Gradient Alignment} (GA) \citep{zhao2021learning} that identifies the tail-group (i.e., bias-conflicting)  examples based on the agreement between two sets of epoch ensembles, \textit{Bias-conflicting Detection} (BCD) \citep{lee2022biasensemble} that uses the testing error of a biased deep ensemble trained with GCE, and \textit{Learning with Biased Committee} (LWBC) uses the testing error of a bootstrap ensemble.

To this end, our work proposes a novel \textit{self-play estimator} (\Cref{eq:bias_estimator}) that uses bootstrap ensembles to estimate the \textit{generalization gap} due to dataset bias. \textit{self-play estimator} has the appealing property of better controlling for label noise while more stably estimating model variance, addressing two weaknesses of the naive predictive error estimator used in the previous works.


\subsection{Representation learning under dataset bias}

Originated from the fairness literature, 
{\color{black} the earlier work in debiased representation learning has focused on identifying a representation that improves a model's fairness properties, as measured by notions such as demographic parity (DP), equalized odds (EO), or sufficiency
\citep{arjovsky2019invariant, arjovsky2020out, creager2021environment, shui2022fair}. This is commonly achieved by striving to learn a representation that is invariant with respect to the group information.
Such methods are commonly characterized as \textit{in-processing methods} in the existing survey of the fairness literature
\citep{caton2020fairness, hort2022bia}, and was categorized into classes of approaches
}
including adversarial training \citep{beutel2017data, kim2019learning, ragonesi2021learning, zhu2021learning, madras2018learning}, regularization \citep{bahng2020learning, tartaglione2021end, arjovsky2019invariant}, contrastive learning \citep{shen2021contrastive, park2022fair, cheng2020fairfil} and its conditional variants \citep{gupta2021controllable, tsai2021conditionala, tsai2021conditionalb, chi2022conditional}, 
{\color{black} or explicit solutions to a bi-level optimization problem \citep{shui2022fair}. (Please see \citet{caton2020fairness, hort2022bia} for a complete survey).}
However, some later works questions the necessity and the sufficiency of such approaches. For example, some work shows that careful training of the output head along is sufficient to yield improved performance in fairness and bias mitigation \citep{kang2019decoupling, du2021fairness, kirichenko2022last}, and \citet{cherepanova2021technical} shows that models with fair feature representations do not necessarily yield fair model behavior.

At the meantime, a separate stream of work explores the opposite direction of encouraging the model to learn diverse hidden features. For example, \citet{locatello2019challenging, locatello2019fairness} establish a connection between the notion of feature disentanglement and fairness criteria, showing that feature disentanglement techniques may be
a useful property to encourage model fairness when sensitive variables are not observed. However, such techniques often involves specialized models (e.g., VAE) which restricts the broad applicability of such approaches. Some other work explores feature augmentation techniques to learn both invariant and spurious attributes, and use them to debias the output head \citep{lee2021learning}. Finally, a promising line of research has been focusing on using self-supervised learning to help the model avoid using spurious features in model predictions \citep{chen2020self, xie2020n, cai2021theory, hamidieh2022evaluating}. Our work follows this latter line of work by proposing novel techniques to encourage model to learn diverse features that is \textit{bias-aware}, but with a distinct purpose of better uncertainty quantification.

\subsection{Active learning under dataset bias}

In recent years, the role of training data in ensuring the model's fairness and bias-mitigation performance has been increasing noticed. Notably, \citep{chen2018my} presented some of the earlier theoretical and empirical evidence that increasing training set size along is already effective in mitigating model unfairness. Correspondingly, under the assumption that the \textit{group information in the unlabelled set is fully known}, there has been several works that studies group-based sampling strategies and their impact on model behavior. For example, \citet{rai2010domain, wang2014active} shows group-based active sampling strategy improves model performance under domain and distributional shifts, and \citet{abernethy22active} proves a guarantee for a worst-group active sampling strategy's ability in helping the SGD-trained model to convergence to a solution that attains min-max fairness.
A second line of research focuses on designing better active learning objectives that incorporates fairness constraints, e.g., \textit{Fair Active Learning (FAL)} \citep{anahideh2022fair} and \textit{PANDA} \citep{sharaf2022promoting}. \citet{agarwal2022does} introduce a data repair algorithm using the coefficient of variation to curate fair and contextually balanced data for a protected class(es).
Furthermore, there exists few active learning works formulating the objective of their method as optimizing a fairness-aware objective. For example, \textit{Slice Tuner} \citep{tae2021slice} proposes adaptive sampling strategy based on per-group learning curve to minimize fairness tradeoff, performs numeric optimization. \citet{cai2022adaptive} which formalized the fairness learning problem as an min-max optimization objective, however their did not conduct further theoretical analysis of their objective, but instead proposed a per-group sampling algorithm based predicted model error using linear regression. 
{\color{black}
Finally, a recent line of active learning work has been well-developed to formulate active learning from the perspective of distributional matching between labeled and unlabeled datasets. \citep{shui2020deep, de2021discrepancy, mahmood2021low, xie2022active}. These approaches are well-equipped to hand distributional shift, but often comes with the cost of assuming knowledge of the test feature distribution at the training time \citep{shui2020deep, de2021discrepancy, mahmood2021low, xie2022active}.} 
In comparison, our work conducts theoretical analysis of the optimization problem \Cref{sec:theory}, and our proposed method (\gls{ISP}) does not require a priori knowledge (e.g., group information) from the unlabelled set.

On the other hand, there exists active re-sampling methods that do not require the knowledge of group information in the unlabelled set. For example, \citet{amini2019uncovering} learns the data distribution using a VAE model under additional supervision of class / attribute labels, and then perform IPW sampling with respect to learned model. REPAIR \citep{li2019repair} that estimates dataset bias using prediction error of a weak model, and then re-train model via e.g., sample re-weighting based on the estimated bias. The bias estimation method used in this work is analogous to that of the \gls{JTT}, which we compare with in our work. A work close to our direction is \citet{branchaud2021can}, which shows \gls{DNN} uncertainty (i.e., BatchBALD with Monte Carlo Dropout \citep{kirsch2019batchbald}) helps the model to achieve fairness objectives in active learning on a synthetic vision problem. Our empirical result confirms the finding of \citet{branchaud2021can} on realistic datasets, and we further propose techniques to improve the vanilla \gls{DNN} uncertainty estimators for more effective active learning under dataset bias.

As an aside, a recent work \citet{farquhar2020statistical} studies the statistical bias in the estimation of active learning objectives due to the non-i.i.d. nature of active sampling. This is separate from the issue of dataset bias (i.e., imbalance in data group distribution) which we focus on in this work.

\subsection{Uncertainty estimation with \gls{DNN}s} 

In recent years, the probabilistic \gls{ML} literature has seen a plethora of work that study enabling calibrated predictive uncertainty in \glspl{DNN}s. Given a model $f$, the probabilistic \gls{DNN} model aims to learn a predictive distribution for the model function $f$, such that given training data $D=\{(y_i, \rvx_i)\}_{i=1}^n$ and a testing point $\rvx_{test}$, the model outputs a predictive distribution $f(\rvx_{test}) \sim P(f|\rvx_{test}, D)$ rather than a simple point prediction. To this end, the key challenge is to learn a predictive distribution (implicitly or explicitly) during the SGD-based training process of \gls{DNN}, generating calibrated predictive uncertainty without significantly impacting the accuracy or latency when compared to a deterministic \gls{DNN}.

To this end, the classic works focus on the study of Bayesian neural networks (BNNs) \citep{neal2012bayesian}, which took a full Bayesian approach by \textit{explicitly} placing priors to the hidden weights of the neural network, and performance MCMC or variance inference during learning. Although theoretically sound, BNN are delicate to apply in practice, with its performance highly dependent on prior choice and inference algorithm, and are observed to lead to suboptimal predictive accuracy or even poor uncertainty performance (e.g., under distributional distribution shift) \citep{wenzel2020good, izmailov2021bayesian}. Although there exists ongoing works that actively advancing the BNN practice (e.g., \cite{dusenberry2020efficient}). On the other hand, some recent work studies computationally more approaches that \textit{implicitly} learn a predictive distribution as part of deterministic SGD training. Notable examples include Monte Carlo Dropout \citep{gal2016dropout} which generates predictive distribution by enabling the random Dropout mask during inference, and ensemble approaches such as Deep Ensemble \citep{lakshminarayanan2017simple} and their later variants \citep{maddox2019simple,wenzel2020hyperparameter,havasi2020training} that trains multiple randomly-initialized networks to learn the  modes of the posterior distribution of the neural network weights \citep{wilson2020bayesian}. Although generally regarded as the state-of-the-art in deep uncertainty quantification, these methods are still computationally expensive, requiring multiple \gls{DNN} forward passes at the inference time.

At the meantime, a more recent line of research avoids probabilistic inference for the hidden weights altogether, focusing on learning a scalable probabilistic model (e.g., Gaussian process) to replace the last dense layer of the neural network \citep{van2020uncertainty,van2021feature,liu2022simple,collier2021correlated}. A key important observation in this line of work is the role of hidden representation quality in a model's ability in obtaining high-quality predictive uncertainty. In particular, \citet{liu2022simple, van2020uncertainty} suggests that this failure mode in \gls{DNN} uncertainty can be caused by an issue in representation learning known as \textit{feature collapse}, where the \gls{DNN} over-focuses on correlational features that help to distinguish between output classes on the training data, but ignore the non-predictive but semantically meaningful input features that are important for uncertainty quantification. \citep{ming2022impact} also observed that \gls{DNN} exhibits particular modes of failure in \gls{OOD} detection in the presence of dataset bias. Later, \citet{tran2022plex, minderer2021revisiting} suggests that this issue can be partially mitigated by large-scale pre-traininig with large \gls{DNN}s, where larger pre-trained \gls{DNN}'s tend to exhibit stronger uncertainty performance even under spurious correlation and subpopulational shift. In this work, we confirm this observation in the setting of dataset bias in \Cref{fig:2d_example}), and propose simple procedures to mitigate this failure mode in representation learning without needing any change to the \gls{DNN} model, and illustrates improvement even on top of large-scale pre-trained \gls{DNN}s (BERT).



\textbf{Deep uncertainty methods in active learning.} Active learning with \gls{DNN}s is an active field with numerous theoretical and applied works, we refer to \citet{matsushita2018deep, ren2021survey} for comprehensive survey, and only mention here few notable methods that involves \gls{DNN} uncertainty estimation techniques. Under a classification model, the most classic approach to uncertainty-based active learning is to use the predictive distribution's entropy, confidence or margin as the acquisition policy \citep{settles1994active}. Notice that in the binary classification setting, these three acquisition policy are rank-equivalent since they are monotonic to the distance between $max[p(\rvx), 1-p(\rvx)]$ and the null probability value of 0.5. On the other hand, \textit{\gls{BADGE}} \citep{ash2019deep} proposes to blend diversity-based acquisition policy into uncertainty-based active learning, by applying k-means++ algorithm to the gradient embedding of the class-specific logits (which quantifies uncertainty). As a result, \textit{\gls{BADGE}} may also suffer from the pathology in model representation under dataset bias, which this work is attempt to address. 

Finally, \cite{houlsby2011bayesian} has proposed a information-theoretic policy \gls{BALD}, which measures the mutual information between data points and model parameters and is adopted in the deep uncertainty literature \citep{gal2017deep, kirsch2019batchbald, kothawade2022clinical}. However, stable estimation of mutual information can be delicate in practice, and we leave the investigation of these advanced acquisition policies under dataset bias for future work.




\section{Experiment Details and further discussion}
\label{sec:exp_app}

\subsection{2D Classification}
\label{sec:exp_2d_app}

\begin{figure}[ht]
    \centering
    \includegraphics[
    width=\linewidth]{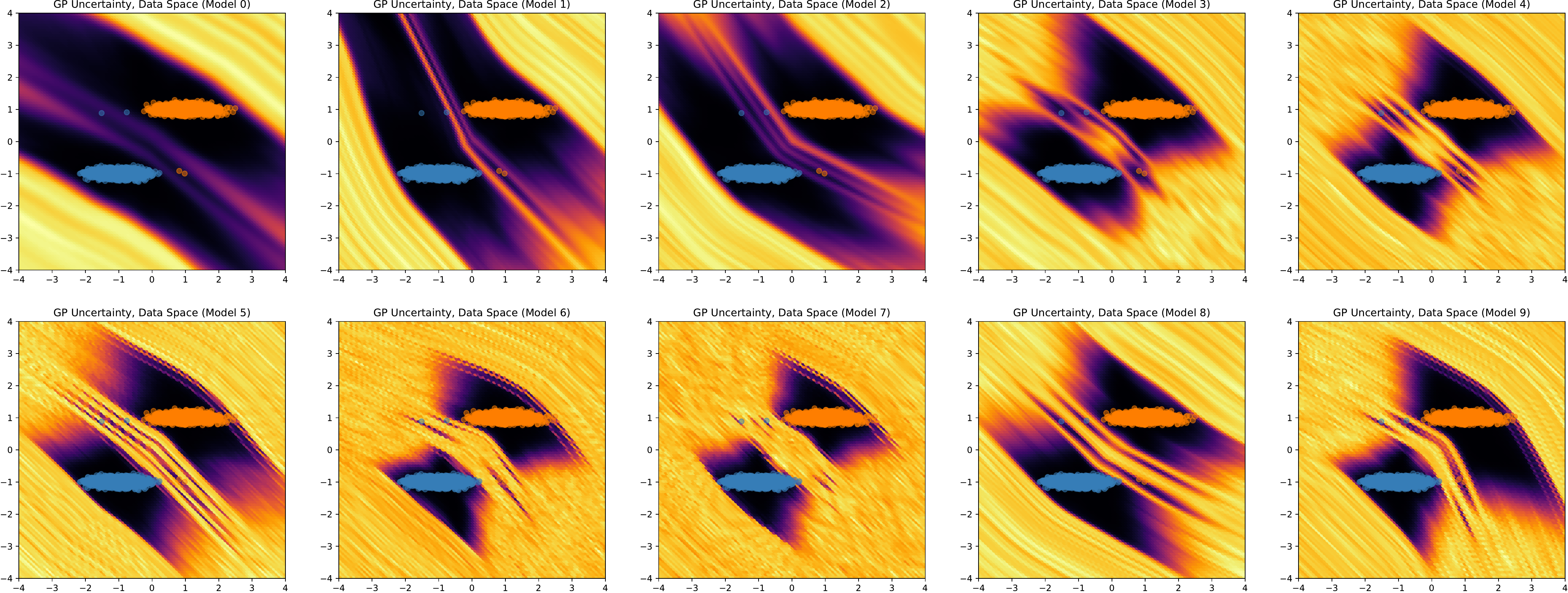}
    \caption{Uncertainty surface of individual ensemble members, ERM training}
    \label{fig:2d_uncertainty_by_model_base}
\end{figure}

\begin{figure}[ht]
    \centering
    \includegraphics[
    width=\linewidth]{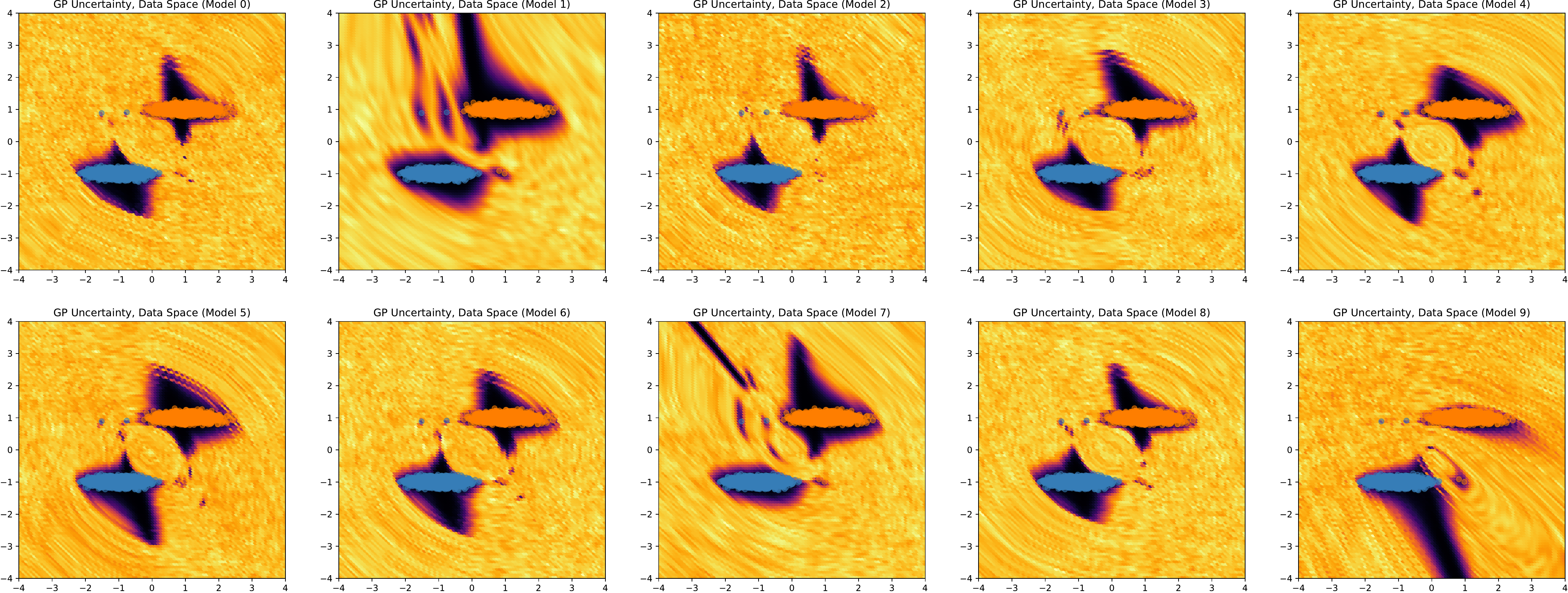}
    \caption{Uncertainty surface of individual ensemble members, introspective training.}
    \label{fig:2d_uncertainty_by_model_isp}
\end{figure}

We train a 10-member neural Gaussian process ensemble (as introduced in \Cref{sec:deep_uncertainty}), where each ensemble member is based on a 6-layer Dense residual network with 512 hidden units and pre-activation dropout mask (rate = 0.1). The model is trained using Adam optimizer ( learning rate = 0.1) under cross entropy loss, and with a batch size 512 for 100 epochs. After training, each ensemble member returns a tuple of predicted label probability, predicted under-representation probability and predictive uncertainty $\{(p_k(y|\rvx), p_k(b|\rvx), v_k(\rvx))\}_{k=1}^{10}$, and we compute the ensemble's predicted probability surface as $\E_k[p(y|\rvx)]$, predicted underrepresentation surface as $\E_k[p_k(b|\rvx)]$, and the predictive uncertainty surface as $\E_k[v_k(y|\rvx)]$, where $\E_k$ is the empirical average over the ensemble member predictions. The predictive uncertainty surface of individual members is shown in Figures \ref{fig:2d_uncertainty_by_model_base}-\ref{fig:2d_uncertainty_by_model_isp}.

As shown, compared to the \gls{ERM}-trained model, the introspective-trained model generates similar label prediction decision $I(p(y|\rvx) > 0.5)$ (Figures \ref{fig:2d_prediction_isp} v.s. \ref{fig:2d_prediction_base}), but with much improved uncertainty surface (Figures \ref{fig:2d_uncertainty_isp} v.s. \ref{fig:2d_uncertainty_base}). Specifically, we compute predictive variance using the standard Gaussian process variance formula $v(\rvx_{test})=\rvk(\rvx_{test})^\top \rmV \rvk(\rvx_{test})$, where $\rvk(\rvx_{test}) = [k(\rvx_{test}, \rvx_1), \dots, k(\rvx_{test}, \rvx_n)]_{n \times 1}$ is a vector of kernel distances based on the embedding distances $||h(\rvx_{test})-h(\rvx_i)||_2$ from the training data (\Cref{sec:deep_uncertainty}). As shown, the model uncertainty under \gls{ERM} model are not sufficiently sensitive to directions in the data space that are irrelevant for making prediction decisions on the training data (i.e., the directions that are parallel to the decision boundary) (\Cref{fig:2d_uncertainty_base}). As a result, it did not learn sufficiently diverse hidden features, leading to a significantly warped representation space that is extremely stretched out in the direction that is orthogonal to the decision boundary, and extremely compressed otherwise (\Cref{fig:2d_representation_base}). Consequently, the model cannot strongly distinguish the minority examples from the majority examples in the representation space, and can become overconfident even in unseen regions that was never covered by training data. This can be undesirable for uncertainty quantification under data bias,  especially for the purpose of identifying underrepresented minority examples, where the distinguishing features between the minority and the majority examples are not predictive for the target label (e.g., the image background). This issue is further exacerbated in the single models (see \Cref{fig:2d_uncertainty_by_model_base}). 
In comparison, the uncertainty surface from an introspective-trained model does not suffer from this failure case. As shown in \Cref{fig:2d_uncertainty_isp}, the model is less inclined to become overconfident in unseen regions, especially in the neighborhood of the minority examples. 
Correspondingly in the representation space, the model learned more diverse features and is able to better distinguish the minority examples from the majority examples (\Cref{fig:2d_uncertainty_isp}). 
To understand how introspective training induces such improvement in model behavior, Figures (\ref{fig:2d_representation_base}) and (\ref{fig:2d_bias_isp}) visualize the model's underrepresentation prediction $p(b|\rvx)$ in the representation space and the data space, respectively. As shown, due to the need of predicting the underrepresented examples (i.e., ``introspection") during training, the model has to learn hidden features that distinguishes the minority examples from the majority examples in its representation space, to a degree that they can be separated by a linear decision boundary in the last layer (\Cref{fig:2d_bias_isp}). Consequently, the model naturally learns a more disentangled representation space through simple multi-task training, and is able to provide predicted bias probabilities $p(b|\rvx)$ (\Cref{fig:2d_bias_isp}) in addition to high-quality predictive uncertainty (\Cref{fig:2d_uncertainty_isp}) for the downstream active learning applications.

\subsection{Tabular and Language Experiments}
\label{sec:exp_ml_app}

\paragraph{Data.} For tabular data, we use the U.S. Census Income data \texttt{adult} from the official UCI repository\footnote{\url{https://archive.ics.uci.edu/ml/datasets/adult}}. For the language task, we use the \texttt{CivilCommentsIdentity} from the TensorFlow Dataset repository\footnote{\url{https://www.tensorflow.org/datasets/catalog/civil_comments}}. For Census Income, we define the underrepresented groups as the union of (Female, High Income) and (Black, High Income); for Toxicity Detection, we define the underrepresented groups as the identity $\times$ label combination (male, female, white, black, LGBTQ, christian, muslim, other religion) $\times$ (toxic, non-toxic) (16 groups in total) as in \citep{koh2021wilds}. 
For CivilComments, the identity annotation is a value between $(0, 1)$ (it is the average rating among multiple raters), and we include an example into the underrepresented group only if the rating $>$ 0.99 (i.e. all raters agree about the identity) following \citep{koh2021wilds}. However, we do note that this leads to a under coverage of the group membership, as many comments with plausible identity mentions are not included into the group identity labels.

\paragraph{Model.} For tabular experiments, we use a 2-layer Dense ResNet model with 128 hidden units and pre-activation dropout rate = 0.1, using a random-feature Gaussian process with hidden dimension 256 as the output layer \citep{liu2022simple} (In the preliminary experiments, we tried larger models with update to 6-layers and 1024 hidden units, and did not observe significant improvement). For language experiments, we used BERT$_{\mbox{small}}$ mode initialized from the official pre-trained checkpoint released at BERT GitHub page\citep{turc2019well}\footnote{\url{https://github.com/google-research/bert}}. In each active learning round, we train the Dense ResNet model with Adam optimizer with learning rate 0.1, batch size 256 and maximum epoch 200; and train the BERT model with AdamW optimizer (learning rate 1e-5) for 6 epochs with batch size 16.

\begin{figure}[ht]
    \centering
    \includegraphics[
    width=0.8\linewidth]{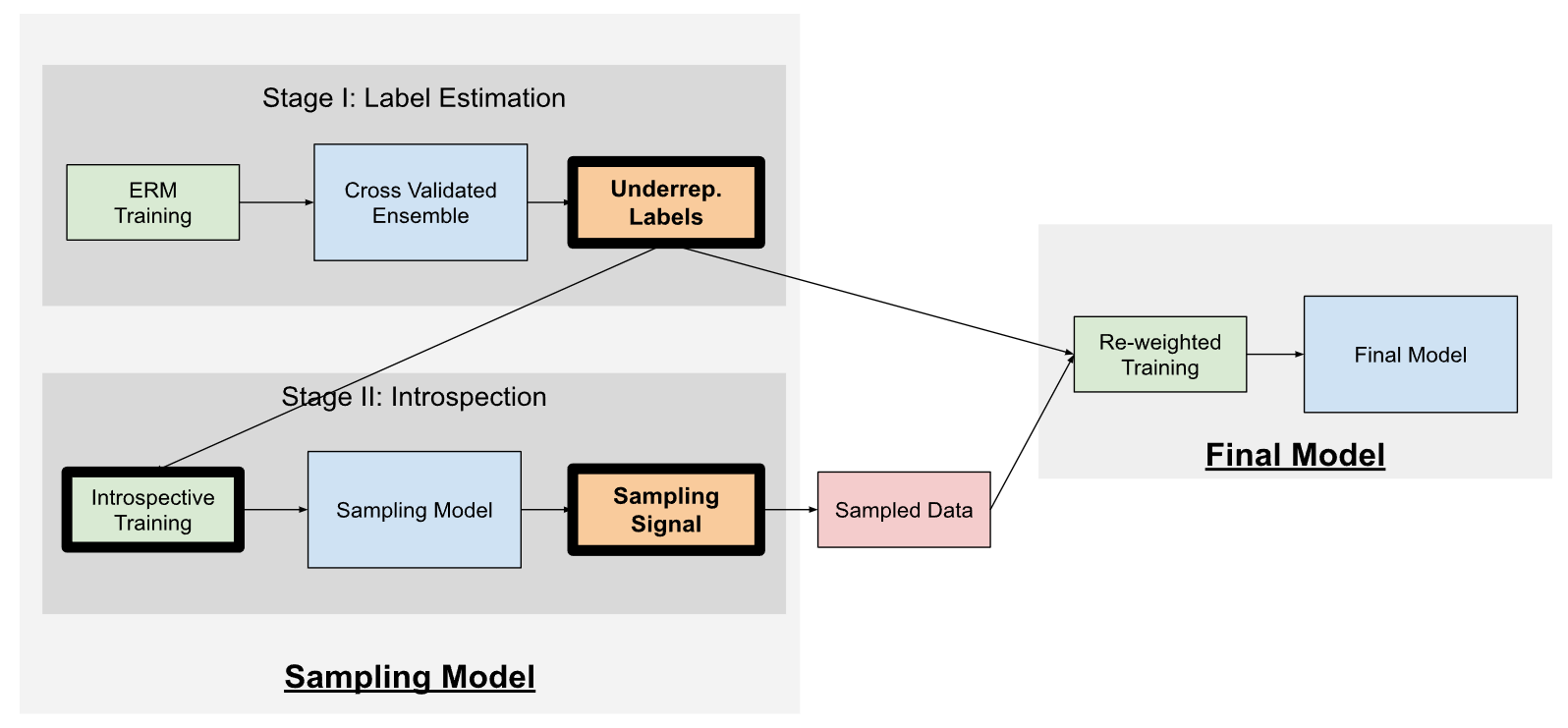}
    \caption{Experiment Protocol. Boxes with thick outlines (Underrepresentation Label, Introspective Training Method, Sampling Signal) indicates the experiment components where the methods differ.}
    \label{fig:exp_protocol}
\end{figure}

\paragraph{Active Learning Protocol.} 

\Cref{fig:exp_protocol} visualizes the experiment protocol. As shown, in each stage, we first (optionally) trains a cross validated ensemble to estimate the under-representation labels, where we split the data into 10 cross-validation splits, and train ensemble members on 1 split and predict the rest of the 9 splits. We then use the ensemble's in-sample and out-of-sample predictions to compute the underrepresentation label $\hat{b}_i$ (\Cref{eq:bias_estimator}), and conduct introspective training (\cref{eq:introspective_objective}) to generate the final active sampling signals for 8 rounds to generate the final sampled data (red box). At the end of round 8, we estimate the underrepresentation label for the final sampled data, and send it to the final model for reweighted training to generate the full accuracy-fairness frontier. The sampling model is always a 10-member ensemble of neural Gaussian process (introduced in \Cref{sec:deep_uncertainty}), and the final model is always a single \gls{DNN} with architecture identical to the sampling model (i.e., 2-layer Dense ResNet for census income and BERT$_{small}$ for toxicity detection).

For both tasks, we randomly sample as small subset as the initial labelled dataset (2,500 out of 32,561 total training examples for census income, and 50,000 out of total 405,130 examples for toxicity detection), and use the rest of the training set as the unlabelled set for active learning. For each sampling round, the \gls{AL} model acquires 1,500 examples for census income, and 15,000 examples for the toxicity detection, so the total sample reaches roughly half of the total training set size after 8 rounds.

In the final model training, we use the standard re-weighting objective \citep{liu2021just}: 
$$\sum_{(x,y)\not\in \hat{\gB}} L_{ce}(y, f(\rvx)) + \lambda \sum_{(x,y)\in \hat{\gB}} L_{ce}(y, f(\rvx))$$ 
where $\hat{\gB}$ is the set of underrepresented examples identified by the underrepresentation label, i.e., $(x_i,y_i) \in \hat{\gB}$ if $\hat{b}_i > t$. We vary the thresholds $t$ and the up-weight coefficient $\lambda$ over a 2D grid ($t \in \{0.05, 0.1, 0.15, ..., 1.0\}$ and $log(\lambda) \in \{0., 0.5, 1, 1.5, \dots, 10.\}$) to get a collection of model accuracy-fairness performances (i.e., accuracy v.s. worst-group accuracy), and use them to identify the Pareto frontier defined by this combination of data and reweighting signal. 

\paragraph{Active Learning Signals.}

In this work, we consider four types of active sampling signals. Recall that the sampling model (neural Gaussian process ensemble) is a K-member ensemble that generates three predictive quantities: (1) label probability $\{p_k(y|\rvx)\}_{k=1}^{10}$, (2) underrepresentation probability $\{p_k(b|\rvx)\}_{k=1}^{10}$ and (3) predictive variance $\{v(\rvx)\}_{k=1}^{10}$ (recall that $\E_k$ and $Var_k$ are the empirical mean and variance).
\begin{itemize}
    \item \textbf{Margin}: The gap between the highest class probability and the second highest class probability for the output label. In the binary prediction context, this is equivalent to $2 * |p(y|\rvx) - 0.5|$, i.e., the gap between the mean predicted probability and the null value of 0.5. We use the mean predictive label probability of the ensemble, which leads to: 
    $$Margin(\rvx)=2 * |\E_k(p_k(y|\rvx)) - 0.5|.$$
    \item \textbf{Predicted Underrepresentation}: The mean predictive underrepresentation probability of the ensemble, which leads to: 
    $$Underrep(\rvx)= \E_k(p_k(b|\rvx)).$$
    \item \textbf{Diversity}: i.e., Ensemble Diversity (introduced in \Cref{sec:deep_uncertainty}). The variance of label predictions:
    $$Diversity(\rvx)= Var_k(p_k(y|\rvx)).$$
    \item \textbf{Variance}: i.e., Predictive Variance (introduced in \Cref{sec:deep_uncertainty}). The mean of predictive variances:
    $$Variance(\rvx)= \E_k(v_k(\rvx)).$$
\end{itemize}

\section{Noise-bias-variance decomposition in Ridge Regression}
\label{sec:error_decomposition_ridge}

Consider fitting a ridge regression model $f(\rvx_i)=\beta^\top \rvx_i$ to the Gaussian observation data $y_i=\theta^{\top} \rvx_i + \epsilon, \epsilon \sim N(0, \sigma^2)$ under an imbalanced experiment design, where we have $|\gG|$ treatment groups and $n_g$ observations in each group. . Here, $\rvx_i = [1_{g_i = 1}, ..., 1_{g_i = |\gG|}]$ is a $|\gG| \times 1$ one-hot indicator of the membership of $g_i$ for each group in $\gG$, and $\theta=[\theta_1, \dots, \theta_{|\gG|}]$ is the true effect for each group. Then, under ridge regression, the noise-bias-variance decomposition for group $g$ is $E_D(L(y, f_D))={\color{OliveGreen} \sigma^2} + {\color{Maroon} \frac{(\lambda\theta_g)^{ 2}}{(n_g + \lambda)^2}} + {\color{RoyalBlue} \frac{\sigma^2 n_g}{(n_g + \lambda)^2}}$, where the regularization parameter $\lambda$  modulates a trade-off between the {\color{Maroon} \textit{bias}} and {\color{RoyalBlue} \textit{variance}} terms. In \Cref{sec:error_decomposition_ridge_orthogonal_heterogeneous}, we also treat the case of group-specific noise $\epsilon_i \stackrel{indep}{\sim} N(0, \sigma_g^2)$ .

\subsection{Error Decomposition in a General Setting}
\label{sec:error_decomposition_ridge_general}

We first derive the decomposition in a general setting with data $\{y_i, \phi_i\}_{i=1}^n$, where $\phi_i$ is the $d \times 1$ (fixed) features that follows a distribution $P(\phi)$. We consider a well-specified scenario where the data generation mechanism as:
$$ y_i = \tilde{y}_i + \epsilon, \quad \mbox{where} \quad \tilde{y}_i = \theta^\top \phi_i, \;\; \epsilon \stackrel{iid}{\sim} N(0, \sigma^2),$$
and $\theta_{d \times 1}=[\theta_1, \dots, \theta_{|\gG|}]$ is the true coefficient. Under ridge regression, we fit a linear model $f(\rvx_i)=\phi_i^\top\beta$ to the data by minimizing the following squared loss objective:
$$||\rvy_{n \times 1} - \Phi_{n \times d}\beta_{d \times 1} ||_2^2 + \lambda ||\beta||_2^2,$$
which gives rise to the following solution:
\begin{align}
    \hat{\beta} = (\Phi^\top\Phi + \lambda I_d)^{-1}\Phi^\top \rvy.
    \label{eq:ridge_coef}
\end{align}
notice $\hat{\beta}$ is a random variable that depends on the data $\Phi_{n \times d} = [\phi_1^\top, \dots, \phi_n^\top] \stackrel{iid}{\sim} P(\phi)$. Notice that under squared loss, the ensemble predictors $\bar{f} = argmin_{f} E_\Phi[(f - \hat{\beta}^\top \rvx_i)^2]$ is simply the mean of individual predictors, i.e, $\bar{f} = E_\Phi(\hat{\beta}^\top \rvx_i)=\bar{\beta}^\top \rvx_i$, where $\bar{\beta}=E_\Phi(\hat{\beta})$.

Consequently, given a new observation $\{y, \phi\}$, the noise-bias-variance decomposition of $\hat{\beta}$ under squared loss is:
\begin{align}
    E[(y - \Phi\hat{\beta})^2] 
    &= 
    E_y[(y - \tilde{y})^2] + (\tilde{y} -  \bar{\beta}^\top \phi_i)^2 + 
    E_{\Phi}[\bar{\beta}^\top \phi_i  - \hat{\beta}^\top \phi_i ]^2 
    \nonumber    \\
    &=
    \underbrace{\sigma^2\vphantom{\phi_i^\top[\theta -  \bar{\beta}]}}_{Noise} + 
    \underbrace{\phi_i^\top[\theta -  \bar{\beta}][\theta -  \bar{\beta}]^\top\phi_i}_{Bias} + 
    \underbrace{\phi_i^\top Var(\hat{\beta}) \phi_i}_{variance}.
    \label{eq:error_decomposition_ridge_full}
\end{align}
As shown, to obtain a closed-form expression of the decomposition, we need to first derive the expressions of $Bias(\hat{\beta})=[\theta - \bar{\beta}]$ and $Var(\hat{\beta})$. Under the expression of the ridge predictor (\ref{eq:ridge_coef}), we have:
\begin{align*}
    Bias(\hat{\beta}) 
    &= [\theta-\bar{\beta}] \\
    &= \theta - E[(\Phi^\top \Phi + \lambda I_d)^{-1}\Phi^\top\Phi\theta]\\
    &= E[I - (\Phi^\top \Phi + \lambda I_d)^{-1}\Phi^\top\Phi]\theta\\
    &= \lambda * E[(\Phi^\top \Phi + \lambda I_d)^{-1}]\theta;\\
    Var(\hat{\beta})
    &= E(Var(\hat{\beta}|\Phi)) + Var(E(\hat{\beta}|\Phi)),
\end{align*}
with
\begin{align*}
    E(Var(\hat{\beta}|\Phi)) 
    &= E[(\Phi^\top\Phi + \lambda I_d)^{-1}\Phi^\top Var(\rvy) \Phi(\Phi^\top\Phi + \lambda I_d)^{-1}]
    \\
    &= \sigma^2 E[(\Phi^\top\Phi + \lambda I_d)^{-1}\Phi^\top\Phi(\Phi^\top\Phi + \lambda I_d)^{-1}] \\
    &= \sigma^2 * E[(\Phi^\top\Phi + \lambda I_d)^{-1} - \lambda (\Phi^\top\Phi + \lambda I_d)^{-2}]. \\
    Var(E(\hat{\beta}|\Phi)) 
    &= E[S \theta\theta^\top S^\top] - E[S]\theta\theta^\top E[S^\top]
\end{align*}
where $S=(\Phi^\top\Phi + \lambda I_d)^{-1}\Phi^\top\Phi$.

As shown, the above expression depends on the random-matrix moments 
$E[(\Phi^\top \Phi + \lambda I_d)^{-1}]$, 
$E[(\Phi^\top\Phi + \lambda I_d)^{-2}]$, 
$E[(\Phi^\top\Phi + \lambda I_d)^{-1}\Phi^\top\Phi]$ and 
$E[S \theta\theta^\top S^\top]$.

\subsection{Error Decomposition under Orthogonal Design}
\label{sec:error_decomposition_ridge_orthogonal}

The above moments are in general difficult to solve due to the involvement of matrix inverse and product within the expectation. However, a closed-form expression is possible under an orthogonal design where $\phi_i = [1_{g_i = 1}, \dots, 1_{g_i = |\gG|}]$ is the one-hot vector of treatment group memberships. Then, denote $Diag[z_g]$ the diagonal matrix with diagonal elements $z_g$ and $[z_{g g'}]_{g g'}$ the full matrix whose $(g, g')$ element is $z_{g g'}$, we have:
\begin{align*}
    \Phi^\top\Phi &= diag[n_g] \\
     E[(\Phi^\top \Phi + \lambda I_d)^{-1}] &= diag[\frac{1}{n_g + \lambda}], \\
     E[(\Phi^\top \Phi + \lambda I_d)^{-2}] &= diag[\frac{1}{(n_g + \lambda)^2}], \\
     E[(\Phi^\top\Phi + \lambda I_d)^{-1}\Phi^\top\Phi] &= diag[\frac{n_g}{n_g + \lambda}],
\end{align*}
and
\begin{align*}
E[(\Phi^\top\Phi + \lambda I_d)^{-1}\Phi^\top\Phi \theta\theta^\top \Phi^\top\Phi(\Phi^\top\Phi + \lambda I_d)^{-1}] &=
[\frac{n_g n'_g}{(n_g + \lambda)(n'_g + \lambda)} \theta_g \theta_{g'}]_{g g'}.
\end{align*}
We are now ready to derive the full decomposition (\ref{eq:error_decomposition_ridge_full}), without loss of generality, we assume $\phi_i$ belongs to group $g$. Then:
\begin{align*}
    \phi_i^\top Bias(\hat{\beta})
    &= \lambda * \phi_i^\top E[(\Phi^\top \Phi + \lambda I_d)^{-1}]\theta 
    = \frac{\lambda}{n_g + \lambda}\theta_g;\\
    \phi_i^\top Var(\hat{\beta}) \phi_i 
    &= \sigma^2 * \phi_i^\top E[(\Phi^\top\Phi + \lambda I_d)^{-1} - \lambda (\Phi^\top\Phi + \lambda I_d)^{-2}] \phi_i; \\
    &= \frac{\sigma^2}{n_g + \lambda} - \frac{\lambda \sigma^2}{(n_g + \lambda)^2}
    = \frac{\sigma^2 n_g}{(n_g + \lambda)^2}.
\end{align*}
Consequently, we have the noise-bias-variance decomposition in (\ref{eq:error_decomposition_ridge_full}) as:
\begin{alignat*}{2}
\mbox{\textit{Noise}:} \quad & 
\sigma^2;
\\
\mbox{\textit{Bias}:} \quad & 
||\phi_i^\top Bias(\hat{\beta})||_2^2 &&= \frac{(\lambda\theta_g)^2}{(n_g + \lambda)^2};
\\
\mbox{\textit{Uncertainty}:} \quad & 
\phi_i^\top Var(\hat{\beta}) \phi_i &&= \frac{\sigma^2 n_g}{(n_g + \lambda)^2}.
\end{alignat*}

\subsection{Error Decomposition under Orthogonal Design and Heterogeneous Noise}
\label{sec:error_decomposition_ridge_orthogonal_heterogeneous}

We now consider the case where $y_i \sim N(\theta^\top\phi_i, \sigma_g^2)$ follows a normal distribution with group-specific noise. Using the same decomposition as in \ref{sec:error_decomposition_ridge_general}, we see that:
\begin{align}
    E[(y - \Phi\hat{\beta})^2] 
    &= 
    E_y[(y - \tilde{y})^2] + (\tilde{y} -  \bar{\beta}^\top \phi_i)^2 + 
    E_{\Phi}[\bar{\beta}^\top \phi_i  - \hat{\beta}^\top \phi_i ]^2 
    \nonumber    \\
    &=
    \underbrace{\sigma_g^2\vphantom{\phi_i^\top[\theta -  \bar{\beta}]}}_{Noise} + 
    \underbrace{\phi_i^\top[\theta -  \bar{\beta}][\theta -  \bar{\beta}]^\top\phi_i}_{Bias} + 
    \underbrace{\phi_i^\top Var(\hat{\beta}) \phi_i}_{variance}.
    \label{eq:error_decomposition_ridge_full}
\end{align}
As shown, the nature of the bias and variance decomposition in fact does not change, and the noise component is now the group-specific variance $\sigma_g^2$. Therefore, by following the same derivation as in \Cref{sec:error_decomposition_ridge_orthogonal}, we have:
\begin{alignat*}{2}
\mbox{\textit{Noise}:} \quad & 
\sigma_g^2;
\\
\mbox{\textit{Bias}:} \quad & 
||\phi_i^\top Bias(\hat{\beta})||_2^2 &&= \frac{(\lambda\theta_g)^2}{(n_g + \lambda)^2};
\\
\mbox{\textit{Uncertainty}:} \quad & 
\phi_i^\top Var(\hat{\beta}) \phi_i &&= \frac{\sigma^2 n_g}{(n_g + \lambda)^2}.
\end{alignat*}

\section{Proof of \Cref{thm:bias_awareness}}
\label{sec:bias_awareness_proof}

Through introspective training, there is a guarantee on a model's bias-awareness based on its hidden representation and uncertainty estimates. At convergence, a well-trained model $f = (f_y, f_b)$ should satisfy the property that $p(b=1 | x) = \sigma(f_b(\rvx))$. 

\textit{(I) \textbf{(Bias-aware Hidden Representation)}}  We denote the odds for $\rvx$ belonging to the underrepresented group $\gB$ as $o_b(\rvx) = p(\rvx|b=1)/p(\rvx|b=0)$. Using Bayes' theorem, we derive the following:

\begin{align}
\nonumber
p(b|\rvx) &= \sigma(\beta^T h(\rvx) + \beta_0) 
\\\nonumber
log \, \frac{p(b=1|\rvx)}{p(b=0|\rvx)} &= \beta^T h(\rvx) + \beta_0 
\\\nonumber
log \, \frac{p(\rvx|b=1)p(b=1)}{p(\rvx|b=0)p(b=0)} &= \beta^T h(\rvx) + \beta_0 
\\\nonumber
\beta^T h(\rvx) + \beta_0 &= log \, P(\rvx|b=1) - log P(\rvx|b=0) + log\frac{p(b=1)}{p(b=0)} 
\\
\beta^T h(\rvx) + \beta_0 &= log \, o_b(\rvx) + log\frac{p(b=1)}{p(b=0)} 
\label{eq:prop1_part1}
\end{align}

Hence, the hidden representation is aware of the likelihood ratio of whether an example $\rvx$ belongs to the underrepresented group, and the last-layer bias $\beta_0$ corresponds to the marginal likelihood ratio of the prevalence of the underrepresented groups $p(b=1)/p(b=0)$ .

\textit{(II) \textbf{(Bias-aware Embedding Distance)}} 
Next, we examine the embedding distance between two examples $(\rvx_1, \rvx_2)$, i.e., $||h(\rvx_1) - h(\rvx_2)||_2$.

The Cauchy-Schwarz inequality states that for two vectors $\rvu$ and $\rvv$ of the Euclidean space, $|\langle \rvu, \rvv \rangle| \leq ||\rvu|| \, ||\rvv||$. Hence, the distance between two embeddings can be expressed as $ \beta^T [h(\rvx_1) - h(\rvx_2)] \leq ||\beta||_2 \, ||h(\rvx_1) - h(\rvx_2)||_2$. Using this property and \Cref{eq:prop1_part1}, we derive the following:
\begin{align}
\nonumber
\beta^T h(\rvx_1) - \beta^T h(\rvx_2) &= log \, o_b(\rvx_1) - log \, o_b(\rvx_2) \\
\nonumber
\beta^T [h(\rvx_1) - h(\rvx_2)] &= log \, o_b(\rvx_1) - log \, o_b(\rvx_2) \\
\nonumber
log \, o_b(\rvx_1) - log \, o_b(\rvx_2) & \leq ||\beta||_2 \, ||h(\rvx_1) - h(\rvx_2)||_2 \\
\nonumber
\frac{1}{||\beta||_2} [log \, o_b(\rvx_1) - log \, o_b(\rvx_2)] & \leq ||h(\rvx_1) - h(\rvx_2)||_2 \\
\frac{1}{||\beta||_2} log \frac{o_b(\rvx_1)}{o_b(\rvx_2)} & \leq ||h(\rvx_1) - h(\rvx_2)||_2
\end{align}

Since the above inequality is invariant to the relative position of $(\rvx_1, \rvx_2)$, we also have: $\frac{1}{||\beta||_2} log \frac{o_b(\rvx_2)}{o_b(\rvx_1)} \leq ||h(\rvx_1) - h(\rvx_2)||_2$, which implies:
$$||h(\rvx_1) - h(\rvx_2)||_2 \geq \frac{1}{||\beta||_2} * max(log \frac{o_b(\rvx_1)}{o_b(\rvx_2)}, log \frac{o_b(\rvx_2)}{o_b(\rvx_1)}).$$

As shown, the distance between the hidden embeddings $h(\rvx_1)$, $h(\rvx_2)$ is lower-bounded by the log-odds ratio that a given example is in the underrepresented group. With this guarantee on the model's learned embedding distance, we expect the hidden features to be more diverse than when trained on the main task alone, since it needs to sufficient features to distinguish the underrepresented-group examples from those of the majority in the hidden space.



{\color{black}
\section{Performance Guarantee for Loss-based Tail-group Detection}
\label{sec:detection_guanrantee_app}

In this section, we derive a lower bound for the group detection performance based on the self-play estimator introduced in \Cref{sec:bias_estimation}. We derive the result in a general setting without further assumptions on the data distribution. The goal here is to provide a broadly applicable, and mathematically rigorous account of how various aspects of the data distribution and the model behavior impacts the performance of the subgroup detection procedure, rather than deriving the tightest possible performance guarantee for a specific  family of model class or data distribution. Our proof technique develops a novel Cantelli-type lower bound on the tail probabilities of the form $P(F(l)>q)$ (where $F$ is a \gls{CDF} and $l$ a random variable), and also a novel tight upper bound on $Var(F(l))$, which may be of independent interest.

Let's first establish some notations. Recall $P(\rvx, y, g)$ is the data-generating distribution (\Cref{sec:notation_app}). For the majority group $g=0$ and minority-group $g=1$, we denote $P_0(.)=P(.|g=0)$ and $P_1(.)=P(.|g=1)$ the data distribution for the majority and minority group, and recall $\gamma_0=P(G=0)$ and $\gamma_1=1-\gamma_0$ the prevalence of these two groups. Furthermore:

\textbf{Group-specific error and its distribution}. Given a loss function $L(.,.)$ of Bergman divergence family and a fixed, unfair model $f_y$ that violates the sufficiency criteria \citep{arjovsky2019invariant, shui2022fair}, denote $l_0=L(\tilde{y}_0, f_y(\rvx_0))$ and $l_1=L(\tilde{y}_1, f_y(\rvx_1))$ the cross-validation  generalization gaps for the majority and the minority examples $(\tilde{y}_0, x_0) \sim P_0$ and $(\tilde{y}_1, x_1) \sim P_1$, where $\tilde{y}_g$ represents the true label without label noise. Notice that $(l_0, l_1)$ are random variables due to the randomness in $(P_0, P_1)$. To this end, also denote $(\mu_0, \mu_1), (\sigma_0^2, \sigma_1^2)$ the means and variances of the group-specific losses $(l_0, l_1)$. Due to the $f_y$'s violation of the sufficiency criteria (i.e., $E_0(y|f_y(\rvx)=t) \neq E_1(y|f_y(\rvx)=t)$), we expect the model $f_y$'s cross-validation loss is systematically worse for the minority groups, i.e., there's a systematic difference in group-specific loss $\mu_1 - \mu_0 = d >0$. Finally, denote $F(l_i)=P(l < l_i)$ the \gls{CDF} for the distribution of the loss $l=L(y, f_y(\rvx)), (y, \rvx) \sim P$, and $(F_0, F_1)$ the \gls{CDF} for the distributions of $l_0, l_1$, respectively. 

\textbf{Rank-based Estimator}. We identify the minority group examples using a rank-based estimator. Specifically, recall $F(l_i)=P(l < l_i)$ is the \gls{CDF} of the population loss, then the rank-based estimator for subgroup detection is:
\begin{align}
\hat{I}(g_i=1) = I(F(l_i) > q).
\label{eq:rank_estimator}
\end{align}
That is, we include a training example $\rvx_i$ into the introspective training only if the population quantile of its generalization error is higher than $q$, which is a user-specific threshold controlling the precision and recall of the estimator's identification performance. 

Then, the below theorem describes how the estimator performance $P(g_i=1|F(l_i)>q)$ is related to the user-specified threshold $q$, and the characteristics of data distribution (i.e., the group prevalence $\gamma_0, \gamma_1$) as well as the classifier performance (in terms of the group-specific loss distribution $F_0, F_1$).

\begin{theorem_app}[Tail-group detection performance of rank-based estimator]
For $(l_0, l_1)$ a pair of random variables for minority- and majority-group examples. Denote $d=E(l_1 - l_0)>0$ and $\sigma^2=Var(l_1-l_0)$ the mean and variance of the between-group generalization gap. 
Then, given a user-specified threshold $q \in (0, \frac{d^2}{d^2 + \sigma^2})$, the performance of the rank-based estimator $P(g=1|F(l)>q)$ is bounded by:
\begin{align}
P(g=1|F(l)>q) \geq (1-\gamma_0)^2 + \gamma_0 * \frac{1-\gamma_0}{1-q} * \frac{z^2}{z^2 + 1}
\quad \mbox{where} \quad 
z=\frac{E[F_0(l_1)] - q}{\sqrt{Var[F_0(l_1)]}}.
\label{eq:rank_estimator_lower_bound_app}
\end{align}
Here, $F_0(l_1)=P(l_0 < l_1)$ is the majority-group loss \gls{CDF} $F_0$ evaluated at the minority-group loss $l_1=L(\tilde{y}_1, f_y(\rvx_1))$ where $(\tilde{y}_1, \rvx_1) \sim P_1$.
\label{thm:group_detect_guanrantee_app}
\end{theorem_app}

Proof is at \Cref{sec:group_detection_proof}.  Notice here due to the randomness in $l_1$, $F_0(l_1)$ is a random variable that follows a continuous distribution and is bounded within $F_0 \in [0, 1]$. Therefore $F_0(l_1)$ has valid moments $E[F_0(l_1)]$ and $Var[F_0(l_1)]$. We see that the performance bound (\ref{eq:rank_estimator_lower_bound_app}) is intuitively sensible: a lower majority group prevalence $\gamma_0$, a higher rank threshold $q$, and a higher  likelihood for loss dominance $F_0(l_1)=P(l_0 < l_1)$ all contributes a stronger identification performance. Here, we see that $z=\frac{E[F_0(l_1)] - q}{\sqrt{Var[F_0(l_1)]}}$ is a distribution dependent quantity that governs the difficulty of identifying the minority group.

Clearly, a larger magnitude of $z$ leads to a stronger guarantee in detection performance in Theorem \ref{thm:group_detect_guanrantee_app}. For the interested readers, the below result provides a lower bound for the magnitude of $z$ in terms of the characteristics of model performance (i.e., the moments of the group-specific loss distributions (e.g., $d=\mu_1 - \mu_2$ and $(\sigma^2_1, \sigma^2_2)$), which can be used to obtain a more precise understanding of $P(g=1|F(l)>q)$ in (\Cref{eq:rank_estimator_lower_bound_app}) in practice:

\begin{theorem_app}[Lower bound on the standardized likelihood of loss dominance $z$.]
Consider the standardized likelihood of loss dominance $z=\frac{E[F_0(l_1)] - q}{\sqrt{Var[F_0(l_1)]}}$, where the likelihood of loss dominance $F_0(l_1)=P(l_0 < l_1)$ is a random variable in terms of $l_1=L(\tilde{y}_1, f_y(\rvx_1)), (\tilde{y}_1, \rvx_1)\sim P_1$. We have:
\begin{align}
E[F_0(l_1)] &\geq \frac{d^2}{\sigma^2 + d^2} \quad \mbox{where} \quad 
d=\mu_1 - \mu_0 > 0, \; 
\sigma^2 = Var(l_1 - l_2)\leq \sigma_0^2 + \sigma_1^2,
\nonumber
\\
Var[F_0(l_1)] &\leq \frac{F_0(\mu_0)^2}{4} * \frac{\sigma_1^2}{\sigma_1^2 + d^2},
\label{eq:variance_lower_bound_app}
\end{align}
where $F_0(\mu_0)=P(l_0 \leq \mu_0)$ is the probability of the majority-group loss $l_0$ smaller than its mean $\mu_0 =E(l_0)$. Therefore, $z$ can be bounded by:
\begin{align}
z=\frac{E[F_0(l_1)] - q}{\sqrt{Var[F_0(l_1)]}} 
\geq 
\frac{2}{F_0(\mu_0)}
* \sqrt{\frac{\sigma_1^2 + d^2}{\sigma_1^2}} * (\frac{d^2}{(\sigma_1^2 + \sigma_2^2) + d^2} - q).
\label{eq:standardized_likelihood_lower_bound_app}
\end{align}
\label{thm:standardized_likelihood_guanrantee_app}
\end{theorem_app}
The proof is in \Cref{sec:group_detection_proof}. It relies on a novel upper bound of \gls{CDF} variance (i.e., \Cref{eq:variance_lower_bound_app}), which is important for guaranteeing a high magnitude of $z$ and consequently a tighter bound for group detection performance in \Cref{eq:rank_estimator_lower_bound_app}.

The combination of Theorems \ref{thm:group_detect_guanrantee_app}-\ref{thm:standardized_likelihood_guanrantee_app} provides us an opportunity to quantitatively understand of the group detection performance $P(g=1|F(l)>q)$ in terms of the estimator configuration (i.e., the user-specified threshold $q$), data distribution (i.e., the majority-group prevalence $\gamma_0$), and the degree of unfairness of the vanilla model $f_y$ (in terms of the distributions of the group-specific losses). For example, consider a setting with majority-group prevalence $\gamma_0=0.85$, expected  between-group loss gap $d=1$, group-specific variances $\sigma_1=\sigma_2=0.15$ and  $P(l_0 < \mu_0)=0.5$, using a percentile threshold $q=0.9$, the rank-based estimator $I(F(l)>q)$ has group detection performance $P(g=1|F(l)>q)>0.9$. 

It is also worth commenting that, following the previous discussion, we see that the success of group identification relies on valid estimation of model's generalization error (with respect to the true label). To this end, the K-fold cross-validated ensemble procedure we used in this work is known to produce unbiased and low-variance estimates for the expected generalization loss \citep{blum1999beating, kumar2013near}. 

Finally, we highlight that the above results are broadly applicable and derived under weak, nearly assumption-free conditions. A even tighter performance bounds can be obtained by making further assumptions on the family of data distributions or model class, which is outside the scope of the current work. Another interesting direction is to incorporate the finite-sample estimation error of K-fold cross validation into the analysis \citep{blum1999beating, kumar2013near, bayle2020cross}, although this necessitates a careful treatment of model's generalization behavior (i.e., loss stability) in the subgroup setting, which we will pursue in the future work.
}

\section{Proof of \Cref{thm:optimal_allocation_frontier_main}}
\label{sec:optimal_allocation_frontier_proof}
\begin{proof}
\newcommand{\ragg}{r_{\text{agg}}}
\newcommand{\rpinv}{r^\prime_{\text{inv}}}
Let $r\br{\alpha}=\br{\alpha n}^{-p}$ and $\ragg\br{n} = \tau n^{-q} + \delta$. As discussed in section \ref{sec:theory}, we assume that the group-specific risk decays as 
\[E[R\br{\hat{f}_{\alpha, n}|G=g}] = c_g r\br{\alpha_g} + \ragg\br{n}\]
where $\hat{f}_{\alpha, n}$ is the minimizer of the $\gamma$-weighted empirical risk on a dataset of size $n$ with allocation $\alpha$ and the expectation is with respect to the randomness of the dataset used to train $\hat{f}$. 

Under the above assumption, the optimization problem for $\alpha$ looks like
\[\min_{\alpha \in \Delta^{|\gG|}} \omega \br{\sum_g \gamma_g \br{c_g r\br{\alpha_g} + \ragg\br{n}}} + (1-\omega) \max_g \br{c_g r\br{\alpha_g} + \ragg\br{n}}\]
The term $\ragg\br{n}$ is a constant that does not impact the optimal solution. Hence, we drop it and focus on the problem
\[\min_{\alpha \in \Delta^{|\gG|}} \omega \br{\sum_g \gamma_g \br{c_g r\br{\alpha_g}}} + (1-\omega) \max_g \br{c_g r\br{\alpha_g}}\]
We begin by noting that the objective is strictly convex and the domain of optimization is bounded. Hence, there exists a unique global optimum.

Turning the constraint $\sum_{g \in \gG} \alpha_g = 1$ into a Lagrangian, we obtain
\[\min_{\alpha \geq 0} \omega \br{\sum_g \gamma_g \br{c_g r\br{\alpha_g}}} + (1-\omega) \max_g \br{c_g r\br{\alpha_g}} + \lambda\br{\sum_{g \in \gG} \alpha_g - 1}\]
where $\lambda \in \R$ is a Lagrange multiplier. 

At the optimum, we know that the Lagrangian should have $0$ within its subdifferential wrt $\alpha$  \citep{boyd2004convex}. The subdifferential is given by
\[\left\{\begin{pmatrix}\br{\omega \gamma_1 + (1-\omega)\mu_1} c_1 r^\prime\br{\alpha_1} + \lambda \\ \br{\omega\gamma_2 + (1-\omega)\mu_2} c_2 r^\prime\br{\alpha_2} + \lambda\\ \vdots \\ \br{\omega\gamma_{|\gG|} + (1-\omega)\mu_{|\gG|}} c_{|\gG|} r^\prime\br{\alpha_{|\gG|}} + \lambda \end{pmatrix}  \text{ where } \mu \in \Delta^{|\gG|} \text{ is such that } \mu_g > 0 \iff g \in \argmax_{g^\prime \in \gG} c_{g^\prime} r\br{\alpha_{g^\prime}}\right\}\]
where $r^\prime$ denotes the derivative of $r$. Since $0$ belongs to the subdifferential at the optimum, there must exist $\mu$ satisfying the constraints above such that
\[\alpha^\star_g = \rpinv\br{-\frac{\lambda}{c_g\br{\omega \gamma_g + (1-\omega) \mu_g}}} \quad \forall g \in \gG\]
where $\rpinv$ is the inverse of $r^\prime$. Since $r^\prime$ is a homogeneuous function of its argument, so is its inverse, and $\lambda$ can be eliminated to enforce the constraint $\sum_g \alpha_g=1$, so that the optimal solution $\alpha^\star_g$ is
\[\alpha^\star_g = \frac{\rpinv\br{-\br{c_g\br{\omega \gamma_g + (1-\omega) \mu_g}}^{-1}}}{\sum_{g^\prime \in \gG} \rpinv\br{-\br{c_{g^\prime}\br{\omega \gamma_{g^\prime} + (1-\omega) \mu_{g^\prime}}}}}\]
Let $s$ denote the denominator and $\theta_g = \omega + (1-\omega) \frac{\mu_g}{\gamma_g}$ so that
\[\alpha^\star_g = \frac{\rpinv\br{-\br{c_g\gamma_g \theta_g}^{-1}}}{s}\]
and $\theta$ must satisfy:
\[\sum_{g \in \gG} \theta_g \gamma_g = 1, \theta_g \geq \omega \quad \forall g \in \gG, \theta_g > \omega\iff g \in \argmax_{g^\prime \in \gG} c_{g^\prime} r\br{\alpha_{g^\prime}^\star}\]
Using the fact that $r\br{t}=\br{nt}^{-p}$, we have that 
\[\argmax_{g^\prime} c_{g^\prime} r\br{\alpha_{g^\prime}^\star} = \argmin_{g^\prime} c_{g^\prime}^{-\frac{1}{p}} \alpha_{g^\prime}^\star = \argmin_{g^\prime} c_{g^\prime}^{\frac{1}{p+1}-\frac{1}{p}} \br{\gamma_g \theta_g}^{\frac{1}{p+1}} = \argmin_{g^\prime} c_{g^\prime}^{-\frac{1}{p}} \gamma_g \theta_g\]
If we sort groups in ascending order according to $c_g^{-\frac{1}{p}} \gamma_g$ to obtain $g_1, g_2, \ldots$, a value of $\theta$ satisfying the conditions above can be computed as follows: \\
Initialize $\theta_g = \omega  \quad \forall g \in \gG$ \\
Set $k=1$ \\
Until $\sum_{g} \theta_g \gamma_g = 1$ or $k=|\gG|$ repeat: \\
\[
\begin{array}{ll}
\text{a)} &  l = \min(k+1, |\gG|) \\
\text{b)} & \text{ Set } \theta_{g_j}=\theta_{g_j} t \text{ (for } j \leq k \text{) for the largest $t \geq 1$ such that (i) } \sum_{g} \theta_g \gamma_g = 1 \text{ or (ii) } c_{g_1}^{-\frac{1}{p}}\gamma_{g_1}\theta_{g_1}= \ldots = c_{g_{l}}^{-\frac{1}{p}}\gamma_{g_{l}} \theta_{g_{l}} \\
\text{c)} & k=k+1
\end{array}
\]

It is easy to see that this algorithm must terminate as it can go for at most $|\gG|$ rounds. Further, by construction, we have $\theta_{g} \geq \omega \forall g \in \gG$ since we start at these values and only scale up any of the $\theta_g$.

At the $j$-th iteration of the loop,  we have that
\[c_{g_1}^{-\frac{1}{p}} \gamma_{g_1}\theta_{g_1}=\ldots=c_{g_j}^{-\frac{1}{p}} \gamma_{g_j}\theta_{g_j}\]
In the $j$-th iteration, we scale all $\theta_{g_j}$ ($m=1,\ldots,j+1$) up by the same factor $t$ by the same amount so that we achieve the above for all groups upto $g_{j+1}$, or we hit the constraint $\sum_g \theta_g\gamma_g=1$. If the former happens, we begin the next iteration with the same invariant. If the latter happens, we have obtained a $\theta$ that satisfies $\sum_g \theta_g \gamma_g=1$ and $\theta_{g_i} > \omega $ for $i < k$ and $g_i \in \argmin_{g^\prime} c_{g^\prime}^{-\frac{1}{p}} \gamma_g \theta_g$ for $i < k$. 

If we reach the iteration where $k=|\gG|$, we can simply scale up all the $\theta_g$ by the same amount until $\sum_g \theta_g\gamma_g=1$ is satisfied and we would have that all $c_g^{-\frac{1}{p}} \gamma_g\theta_g$ is equal for all $g \in \gG$ and all $\theta_g > \omega$. 

\end{proof}


\section{Proof of Theorems \ref{thm:group_detect_guanrantee_app} and \ref{thm:standardized_likelihood_guanrantee_app}}
\label{sec:group_detection_proof}
\subsection{Proof for Theorem \ref{thm:group_detect_guanrantee_app}}

\begin{proof}
By Bayes' rule, we have:
\begin{align}
P(g=1|F(l)>q) &= P(F(l)>q|g=1) * \frac{P(g=1)}{P(F(l)>q)}
\label{eq:detection_precision_0}
\end{align}

\underline{Expression for $P(g=1)$ and $P(F(l)>q)$}

Notice that $P(g=1)=1-\gamma_0$, and $P(F(l)>q)=P(U>q)=1-q$ for $U \sim Unif(0, 1)$. Here we used the fact that $F(l) \sim Unif(0, 1)$ when $F$ is the \gls{CDF} for the distribution of the random variable $l$ \citep{blitzstein2015introduction}. 

\underline{Expression for $P(F(l)>q|g=1)$}

Notice $P(F(l)>q|g=1)=P(F(l_1) > q)$, where $l_1=L(\tilde{y}_1, f_y(\rvx_1)), (\tilde{y}_1, \rvx_1)\sim P_1$ is the random variable for the minority-group loss.
Also notice that $F(l_1)=P(l' < l_1)$ where $l'=L(y', f_y(\rvx'))$ for $(y', \rvx', g')\sim P$ is the non-group-specific loss.  Then a conditional decomposition of $P(F(l_1)>q)$ reveals :
\begin{align*}
P(F(l_1)>q) &= P[P(l' < l_1)>q]\\
&= P \big[P(l' < l_1)>q | g'=0 \big]P(g'=0) + 
P\big[P(l' < l_1)>q| g'=1 \big]P(g'=1) \\
&= P \big( F_0(l_1)>q \big)\gamma_0 + (1-q)(1-\gamma_0),
\end{align*}
where the last equality follows since $P(g'=0)=\gamma_0, P(g'=1)=1-\gamma_0$ and $P\big[P(l' < l_1)>q \big]=P\big[F_1(l_1)>q \big]=P(U > q)=1-q$ for $U\sim Unif(0, 1)$. Here we again used the fact that $F_1(l_1) \sim Unif(0, 1)$ when $F_1$ is the \gls{CDF} for the distribution of $l_1$ \citep{blitzstein2015introduction}. 

\underline{Lower bound for $P \big( F_0(l_1)>q \big)$}

Lower bound $P \big( F_0(l_1)>q \big)$ by deriving a Cantelli-type inequality using the second moment method \citep{lyons2017probability}. Specifically, for a random variable $R$ with $E(R) \geq 0$, we have:
\begin{align*}
E(R)^2 \leq E(R * 1_{R>0})^2 &\leq E(R^2) P(R>0) 
\quad \mbox{which implies} \quad
\frac{P(R>0)}{1-P(R>0)} \geq \frac{E(R)^2}{Var(R)}.
\end{align*}
Setting $R=F_0(l_1) - q$, we have:
\begin{align*}
\frac{P \big( F_0(l_1)>q \big)}{1-P(F_0(l_1)>q)} &\geq \frac{(E(F_0(l_1))-q)^2}{Var(F_0(l_1))} = z^2, 
\quad \mbox{which implies} \quad 
P(F_0(l_1)>q) \geq \frac{z^2}{1 + z^2}
\end{align*}

\underline{Derive the final bound}

Finally, using the above three facts, we can express \Cref{eq:detection_precision_0} as:
\begin{align*}
P(g=1|F(l)>q) 
&= P(F(l)>q|g=1) * \frac{P(g=1)}{P(F(l_1)>q)} 
= P(F(l)>q|g=1) * \frac{1-\gamma_0}{1-q} 
\end{align*}
which further leads to:
\begin{alignat*}{2}
P(g=1|F(l)>q) 
&=      \Big( (1-q)(1-\gamma_0) + P \big( F_0(l_1)>q \big)  \gamma_0  \Big) * \frac{1-\gamma_0}{1-q} 
&\geq   \Big( (1-q)(1-\gamma_0) + \frac{z^2}{1 + z^2} * \gamma_0 \Big) * \frac{1-\gamma_0}{1-q},
\end{alignat*}
yielding the final bound $P(g=1|F(l)>q) \geq (1-\gamma_0)^2 + \gamma_0 * \frac{1-\gamma_0}{1-q} *  \frac{z^2}{1 + z^2}$ as in \Cref{eq:rank_estimator_lower_bound_app}.
\end{proof}

\subsection{Proof for Theorem \ref{thm:standardized_likelihood_guanrantee_app}}

\begin{proof}
For $F_0(l_1)$, recall $F_0$ is the \gls{CDF} of $l_0=L(\tilde{y}_0, f_y(\rvx_0))$ where $(\tilde{y}_0, \rvx_0) \sim P_0$ and  $l_1=L(\tilde{y}_1, f_y(\rvx_1))$ is the random variable of minority-group loss with  $(\tilde{y}_1, \rvx_1) \sim P_1$.

To derive lower bound for $z=\frac{E(F_0(l_1))-q}{Var(F_0(l_1))}$, first derive bounds on $E(F_0(l_1))$ and $Var(F_0(l_1))$:

\underline{Lower bound for $E(F_0(l_1))$}

Recall $P(F(l)|g=1) = P(F(l_1))$, where $l_1=L(\tilde{y}_1, f_y(\rvx_1))$ is the random variable for minority-group error with $(\tilde{y}_1, f_y(\rvx_1)) \sim P_1$. Then:
\begin{align*}
    E(F_0(l_1))  = E(F_0(l) | g = 1) = P(l_0 \leq l_1) = P(l_1 - l_0 \geq 0).
\end{align*}
By the second moment method inequality \citep{lyons2017probability}, we have:
\begin{align}
P(l_1 - l_0 \geq 0) 
&\geq \frac{E[l_1 - l_0]^2}{E[(l_1 - l_0)^2]} 
= \frac{E[l_1 - l_0]^2}{Var(l_1 - l_0) + E[l_1 - l_0]^2}
\nonumber\\
&\geq \frac{E[l_1 - l_0]^2}{Var(l_1) + Var(l_0) + E[l_1 - l_0]^2}
\nonumber\\
&= \frac{d^2}{(\sigma_0^2 + \sigma_1^2) + d^2}.
\label{eq:mean_loss_diff_bound}
\end{align}

\underline{Upper bound for $Var(F_0(l_1))$}

Recall that $Var(F(l))=E\Big[(F(l)-E(F(l)))^2\Big]$. Using the iterative expectation formula, we split the variance computation for $F_0(l_1)$ into two regions of $l_1$ depending on whether the minority-group loss of $l_1 \geq \mu_1 - \lambda \sigma_1$: 
\begin{align*}
    Var(F_0(l_1)) 
    = 
    & Var(F_0(l_1) | l_1 \leq \mu_1 - \lambda \sigma_1)
    P(l_1 \leq \mu_1 - \lambda \sigma_1)
    + 
    \nonumber  \\
    & Var(F_0(l_1) | l_1 > \mu_1 - \lambda \sigma_1)
    P(l_1 > \mu_1 - \lambda \sigma_1),
\end{align*}
which holds for any positive multiplier $\lambda > 0$. 

Notice that in the above,  $P(l_1 < \mu_1 - \lambda \sigma_1)$ in the second line describes the tail probability of the loss distribution of $l_1$, for suitably large $\lambda$, $P(l_1 < \mu_1 - \lambda \sigma_1)$ should be small. On the other hand, $Var(F_0(l_1) | l_1 \geq \mu_1 - \lambda \sigma_1)$ in the second line describes the variance of $F_0(l_1)$ in the region where $l_1$ is large. When the distribution of $l_0$ and $l_1$ is well separated (i.e., $\mu_1-\mu_0 = d > 0$), for suitable value of $\lambda$, we expect the value of $F_0(l_1=F_0( \mu_1 - \lambda \sigma_1))$ to be high and close to 1, and as a result the $Var(F_0(l_1)|l_1 \geq \mu_1 - \lambda \sigma_1)$ will be low since $F_0(l_1)$ is bounded within a small range $[F_0( \mu_1 - \lambda \sigma_1), 1]$. Consequently, to obtain a tight upper bound of $Var(F(l))$, it is sufficient to identify a suitable value of $\lambda$ such that both $Var(F_0(l_1) | l_1 \geq \mu_1 - \lambda \sigma_1)$ and $P(l_1 < \mu_1 - \lambda \sigma_1)$ are low.

To identify a suitable value of $\lambda$, first derive an upper bound of $Var(F_0(l_1))$ in terms of $\lambda$. Notice below two facts:
\begin{itemize}
\item By Cantelli's inequality:
$$P(l_1 \leq \mu_1 - \lambda \sigma_1) \leq \frac{1}{\lambda^2 + 1}.$$ 
\item By the inequality for variance of the bounded variables:
$$Var(F_0(l_1)|l_1 \leq \mu_1 - \lambda \sigma_1) \leq \frac{1}{4} F_0(\mu_1 - \lambda \sigma_1)^2,$$
where the first inequality follows by the fact that for a random variable $R$ bounded between $[a, b]$ (in this case between $[F(\mu_1 - \lambda \sigma_1), 1]$), its variance is bounded by $Var(R) \leq (b-E(R))(E(R)-a) \leq \frac{1}{4}(b-a)^2$.

\item By Markov's inequality, $F_0(l_1) = P(l_1 \leq l_1) \geq \mu_0/l_1$, which implies:
$$Var(F_0(l_1)|l_1 \geq \mu_1 - \lambda \sigma_1) 
\leq 
\frac{1}{4} (1-F_0(\mu_1 - \lambda \sigma_1))^2
\leq 
\frac{1}{4} (1-\frac{\mu_0}{\mu_1 - \lambda \sigma_1})^2,$$
where the first inequality also follows by the variance inequality of the bounded variables.
\end{itemize}
Using the above two facts, we can bound $Var(F_0(l_1))$ as:

\begin{align}
Var(F_0(l_1)) 
&= 
 Var(F_0(l_1) | l_1 \leq \mu_1 - \lambda \sigma_1) * 
P(l_1 \leq \mu_1 - \lambda \sigma_1) + 
 Var(F_0(l_1) | l_1 > \mu_1 - \lambda \sigma_1) * 
P(l_1 \geq \mu_1 - \lambda \sigma_1) 
\nonumber\\
&\leq 
Var(F_0(l_1) | l_1 \leq \mu_1 - \lambda \sigma_1) * 
\frac{1}{\lambda^2 + 1} + 
Var(F_0(l_1) | l_1 > \mu_1 - \lambda \sigma_1) * 
\frac{\lambda^2}{\lambda^2 + 1} 
\nonumber\\
&\leq 
\frac{1}{4} F_0(\mu_1 - \lambda \sigma_1)^2 * 
\frac{1}{\lambda^2 + 1} + 
\frac{1}{4} (1-\frac{\mu_0}{\mu_1 - \lambda \sigma_1})^2 * 
\frac{\lambda^2}{\lambda^2 + 1}.
\label{eq:var_F0_iterated_expectation}
\end{align}
where the first inequality holds since the first conditional variance term $Var(F_0(l_1) | l_1 \leq \mu_1 - \lambda \sigma_1)$ (i.e., variance in the majority bulk) is expected to be much larger than the second term $Var(F_0(l_1) | l_1 > \mu_1 - \lambda \sigma_1)$ (i.e., variance in the far tail), therefore assigning the highest possible probability weight $P(l_1 \leq \mu_1 - \lambda \sigma_1)$ to the larger variance term $Var(F_0(l_1) | l_1 \leq \mu_1 - \lambda \sigma_1)$ leads to an upper bound for $Var(F_0(l_1))$.

Further simplifying \Cref{eq:var_F0_iterated_expectation}, we arrive at:
\begin{align*}
Var(F_0(l_1)) 
&\leq 
\frac{1}{4} F_0(\mu_1 - \lambda \sigma_1)^2 * 
\frac{1}{\lambda^2 + 1} + 
\frac{1}{4} (1-\frac{\mu_0}{\mu_1 - \lambda \sigma_1})^2 * 
\frac{\lambda^2}{\lambda^2 + 1}\\
&= \frac{1}{4} * \frac{1}{\lambda^2 + 1} * 
\Big(
F_0(\mu_1 - \lambda \sigma_1)^2 + 
\frac{\lambda^2}{4} (1-\frac{\mu_0}{\mu_1 - \lambda \sigma_1})^2 
\Big) \\
&= \frac{1}{4} * \frac{1}{\lambda^2 + 1} * 
\Big(
F_0(\mu_1 - \lambda \sigma_1)^2 + 
\frac{\lambda^2}{4} (\frac{d - \lambda \sigma_1}{\mu_1 - \lambda \sigma_1})^2 
\Big)
\end{align*}
where recall $d=\mu_1 - \mu_0$ is the expected error gap between the majority and the minority groups. Setting $\lambda = \frac{d}{\sigma_1}$, we have:
\begin{align}
Var(F_0(l_1)) 
&\leq 
\frac{1}{4} * F_0(\mu_0)^2 * \frac{1}{\lambda^2 + 1}
=
\frac{F_0(\mu_0)^2}{4} * \frac{\sigma_1^2}{d^2 + \sigma_1^2}.
\label{eq:var_loss_diff_bound}
\end{align}

\underline{Lower bound for $z$}

Finally, plugging the bounds for $E(F_0(l_1))$ and $Var(F_0(l_1))$ from \Cref{eq:mean_loss_diff_bound} and \ref{eq:var_loss_diff_bound} into $z=\frac{E[F_0(l_1)] - q}{\sqrt{Var[F_0(l_1)]}}$ yields the final lower bound in \Cref{eq:standardized_likelihood_lower_bound_app}, i.e.,
\begin{align*}
z=\frac{E[F_0(l_1)] - q}{\sqrt{Var[F_0(l_1)]}} 
&\geq 
(\frac{d^2}{(\sigma_1^2 + \sigma_2^2) + d^2} - q) \Big/
\sqrt{\frac{F_0(\mu_0)^2}{4} * \frac{\sigma_1^2}{d^2 + \sigma_1^2}}
\\
&=
\frac{2}{F_0(\mu_0)}
* \sqrt{\frac{\sigma_1^2 + d^2}{\sigma_1^2}} * (\frac{d^2}{(\sigma_1^2 + \sigma_2^2) + d^2} - q).
\end{align*}

\end{proof}

\clearpage
\end{document}